\documentclass[11pt]{article}

\usepackage[preprint]{acl}
\usepackage{enumitem}

\usepackage{times}
\usepackage{latexsym}

\usepackage[T1]{fontenc}

\usepackage[utf8]{inputenc}

\usepackage{microtype}

\usepackage{inconsolata}

\newcommand{\datasetname}{CivilTime-Bench}
\usepackage{graphicx}

\usepackage{pifont}

\newcommand{\cmark}{\ding{51}}
\newcommand{\xmark}{\ding{55}}

\usepackage{multirow}

\usepackage{algorithm}
\usepackage{algpseudocode}
\usepackage{amsmath}
\usepackage{cleveref}
\usepackage{amssymb}
\usepackage{placeins}
\usepackage{dsfont}
\usepackage{graphicx}

\title{When 1+1 $\neq$ 2: Benchmarking Civil-Time Reasoning in Large Language Models}

\author{
\textbf{Melika Shirian}$^{1,*}$ \quad
\textbf{Kianoosh Vadaei}$^{1,*}$ \quad
\textbf{Arshia Hemmat}$^{2,*}$ \\
\textbf{Mohammad Hossein Shaker Ardakani}$^{1,*}$ \quad
\textbf{Ali Mamanpoosh}$^{1,\dagger}$ \\
\textbf{Matin Moqadas}$^{1,\dagger}$ \quad
\textbf{Amirmasoud Rismanchian}$^{3,\dagger}$ \quad
\textbf{Afsaneh Fatemi}$^{1,\ddagger}$ \\[4pt]
$^{1}$University of Isfahan \quad
$^{2}$University of Oxford \quad
$^{3}$University of Tehran
}

\begin{document}
\maketitle

\begingroup
\renewcommand{\thefootnote}{}\footnotemark
\footnotetext{
$^{*}$ Equal contribution (co-first authors).
$^{\dagger}$ Equal contribution within the second group.
$^{\ddagger}$ Third contribution group.
}
\endgroup

\begin{abstract}
Calendar arithmetic is non-linear. When 1:59 AM + 1 minute occurs during daylight saving spring-forward, the result is 3:00 AM, not 2:00 AM, as the hour is skipped. We introduce \datasetname{}, a benchmark for civil-time reasoning covering daylight saving transitions, non-hour offsets, date-line crossings, and ISO week boundaries. With 60,011 training and 10,500 evaluation examples under strict output-contract evaluation, we benchmark seven LLMs (six instruction-tuned models and Llama-3.1-8B base) and find: zero-shot exact match varies substantially across models, contract-aware prompting often improves format compliance but yields inconsistent accuracy gains, few-shot in-context learning provides model-dependent improvements, and fine-tuning achieves 74.6--86.6\% EM on synthetic evaluation but only 36.6--45.2\% on HumanTime-500 for models with valid results on both splits, highlighting substantial limitations in generalization. We release the benchmark and code.\footnote{The benchmark and code are available at \url{https://github.com/CivilTime-Bench/CivilTime-Bench}.}
\end{abstract}

\section{Introduction}
\label{sec:intro}

Calendar arithmetic pervades everyday computing from scheduling systems and travel bookings to log analysis and incident management. Yet even simple civil-time operations (adding 90 minutes at a DST boundary, converting across the international date line) expose fragile assumptions in deployed systems. Large language models, increasingly positioned as general-purpose assistants, face acute challenges here: users expect reliable temporal reasoning, yet production failures with DST transitions, timezone conversions, and date-line arithmetic remain common. Current benchmarks for temporal reasoning (TRAM, Test-of-Time, TimeDial) focus on event ordering and temporal logic, not exact calendar arithmetic under civil-time rules.

\begin{figure}[t]
  \centering
  \includegraphics[width=\linewidth]{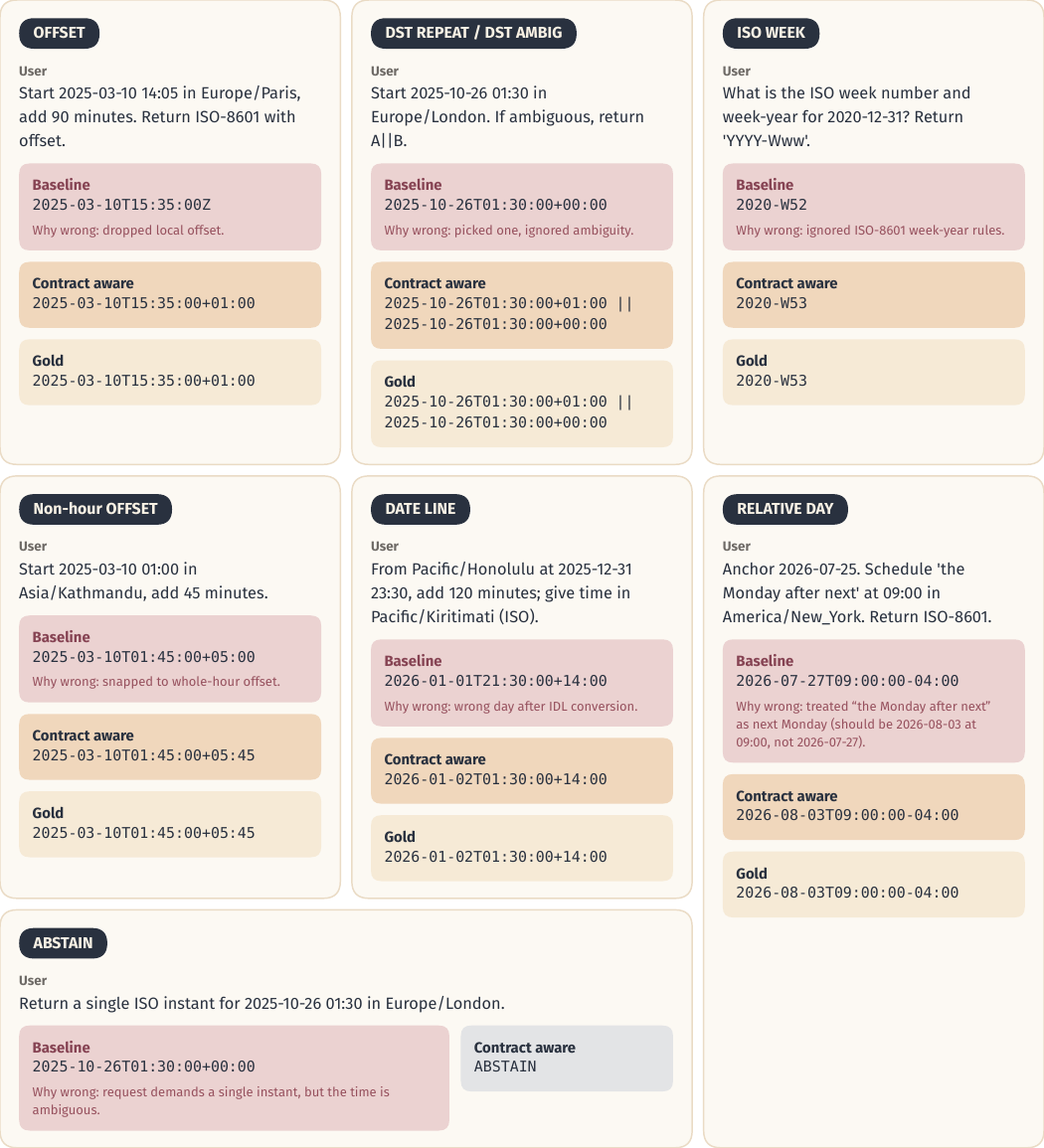}
  \caption{\textbf{Contract-aware prompting can improve output formatting.} Baseline (rose) vs.\ contract-aware (peach) outputs across civil-time categories. Contract restatement can improve format compliance but does not consistently improve exact-match accuracy, showing formatting $\neq$ reasoning.}
  \label{fig:demo}
\end{figure}

\paragraph{Core Challenge.}
Civil time violates basic arithmetic. When 1:59 AM + 1 minute occurs during daylight saving spring-forward, the result is 3:00 AM (the 2:00 AM hour is skipped). During fall-back, repeated wall-clock times can correspond to two distinct instants. Non-hour UTC offsets (e.g., +05:45 in Kathmandu), international date-line crossings, and ISO week boundaries further complicate calendar operations. We study whether current LLMs can handle such non-linear temporal reasoning.

\paragraph{The gap.}
Existing temporal benchmarks stress event ordering (TORQUE), temporal logic composition (Test-of-Time), and commonsense reasoning (TimeDial), but rarely stress DST pathologies, non-hour UTC offsets (±00:30, ±00:45), international date-line transitions, and ISO week/year conventions. Evaluation protocols often permit prose answers or flexible formats, mixing formatting errors with reasoning failures. This conflation obscures the true challenge: are failures due to output serialization or genuine calendrical misconceptions?

\paragraph{Our solution.}
We introduce \datasetname{}, a benchmark enforcing strict, machine-checkable output contracts with explicit offsets for timestamps to separate formatting from reasoning. Each item has a canonical answer in one of five forms: \texttt{ABSTAIN} (no valid instant exists), a date or ISO-week label (e.g., \texttt{YYYY-MM-DD} or \texttt{YYYY-Www}), \texttt{YYYY-MM-DDThh:mm:ss±hh:mm} (single timestamp with offset), \texttt{A||B} (ambiguity pair for DST fall-back), or \texttt{A..B} (half-open interval). This contract enables precise diagnostics: we separately measure raw-output contract compliance (\textsc{ContractOK}), extracted-answer format validity (\textsc{FormatOK}), exact match (EM), and abstention diagnostics (ARA/OAR/F1).

\paragraph{Key findings.}
Evaluating seven LLMs, including one base model, we find:

\begin{enumerate}

\item \textbf{Zero-shot performance varies substantially:} On HumanTime-500, open models with valid zero-shot runs achieve 10.2--28.0\% EM, while GPT-5-nano reaches 73.2\% and Gemini-2.5-Flash-Lite 57.8\%. LLaMA's zero-shot run exhibits a generation failure.

\item \textbf{Contract-aware prompting improves formatting inconsistently:} Output contracts improve FormatOK for several models (e.g., Phi-4: 74.6\%$\to$97.8\% on synthetic evaluation) but yield model-dependent EM changes (Phi-4: 40.8\%$\to$35.8\%), revealing formatting $\neq$ reasoning.

\item \textbf{Type-matched ICL bridges the gap partially:} With 2/4/8-shot exemplars, EM improves substantially for several open models (e.g., Qwen: 10.2\%$\to$51.8\% on HumanTime-500 at 8-shot), but gains vary across models and shot counts, with incomplete prediction coverage for some API-model conditions.

\item \textbf{Fine-tuning substantially improves synthetic performance:} QLoRA achieves 74.6--86.6\% EM on SyntheticTimeBench, but only 36.6--45.2\% on HumanTime-500 for models with valid results on both splits, indicating substantial limitations in transfer.

\item \textbf{Generalization to naturalistic inputs remains hard:} Fine-tuned models with valid results on both splits show 36.2--47.0 percentage point EM drops from synthetic evaluation to HumanTime-500, indicating that strong synthetic performance does not guarantee robust generalization.
\end{enumerate}

\paragraph{Contributions.}
We contribute five components:

\begin{enumerate}

\item \textbf{A comprehensive benchmark:} 60,011 train / 10,500 eval examples covering 11 task types spanning DST, timezones, non-hour offsets, date-line crossings, ISO week, business-day, and interval operations under a strict one-line output contract.

\item \textbf{Baseline evaluation on prior datasets:} Results on TRAM, Test-of-Time, and TimeDial positioning civil-time reasoning within the broader temporal landscape.

\item \textbf{Unified evaluation protocol:} EM, Macro-EM, FormatOK, ContractOK, ARA/OAR/F1, and MedianTE, with paired McNemar tests, disentangling formatting from calendrical reasoning.

\item \textbf{Three prompting regimes with significance testing:} Baseline zero-shot, contract-aware zero-shot, and type-matched ICL (k$\in\{$2,4,8$\}$) with fixed exemplar banks and McNemar analysis.

\item \textbf{Parameter-efficient fine-tuning analysis:} QLoRA results on four open-source models (7B--14B) achieving 74.6--86.6\% EM on synthetic evaluation, together with an assessment of their limited transfer to HumanTime-500.
\end{enumerate}

The benchmark and scorer are released publicly.

\section{Related works}
\label{sec:related}

\begin{table}[t]
  \centering
  \renewcommand{\arraystretch}{1.08}

  \resizebox{\columnwidth}{!}{%
    \begin{tabular}{@{}lcclcc@{}}
      \hline
      \textbf{Benchmark}
      & \textbf{Eval. Size}
      & \textbf{Train Size}
      & \textbf{Task}
      & \textbf{DST/TZ}
      & \textbf{Format} \\
      \hline

      TRAM
      & 10 sets
      & --
      & Ordering, logic
      & \xmark
      & MC/text \\

      Test-of-Time
      & 1.4k
      & --
      & Temporal logic
      & \xmark
      & Synthetic \\

      TimeDial
      & 1.5k
      & --
      & Commonsense
      & \xmark
      & MC cloze \\

      TORQUE
      & 3.2k
      & --
      & Event ordering
      & \xmark
      & Span \\

      COMPLEXTempQA
      & 5.7k
      & --
      & Multi-hop QA
      & \xmark
      & Short text \\

      \hline

      \textbf{Ours}
      & \textbf{10,500}
      & \textbf{60,011}
      & \textbf{Calendar arith.}
      & \textbf{\cmark}
      & \textbf{ISO-8601} \\

      \hline
    \end{tabular}%
  }

  \caption{\textbf{Comparison of temporal reasoning benchmarks.}
  Evaluation sizes are reported separately from training sizes;
  TRAM is listed by the number of component datasets, and --
  denotes training sizes not reported here.
  Prior work focuses on ordering, logic, and commonsense, whereas our benchmark
  evaluates exact calendar arithmetic under civil-time rules, including DST
  skips, repeated and ambiguous times, non-hour offsets, the international date
  line, and ISO week-year calculations.}
  \label{tab:benchmark_comparison}
\end{table}

Temporal reasoning benchmarks span event ordering (TORQUE~\citep{ning2020torque}), temporal logic (Test-of-Time~\citep{fatemi2024test}), and commonsense (TimeDial~\citep{qin2021timedial}), but rarely stress DST transitions, timezone conversions, or date-line arithmetic.

TRAM~\citep{wang2024tram} aggregates ten diverse datasets (scope: broad LLM temporal reasoning; probe: multi-choice/aggregation) but lacks explicit coverage of DST, non-hour offsets, or date-line edge cases. Test-of-Time introduces synthetic temporal logic and crowdsourced time arithmetic (scope: logic+arithmetic; probe: synthetic/crowd) but omits timezone and DST phenomena. TimeDial targets everyday temporal commonsense via cloze (scope: commonsense; probe: MC cloze) without calendrical arithmetic. TORQUE studies before/after ordering over news (scope: reading comprehension; probe: cue-sensitive ordering) but does not address civil-time edge cases. COMPLEXTempQA scales multi-hop temporal QA (scope: complex knowledge; probe: large-scale retrieval) yet lacks non-hour offsets and date-line arithmetic.

Classic temporal information extraction (TempEval-1/2/3, QA-TempEval)~\citep{uzzaman2013semeval} provides foundational TimeML schemas for event/timex extraction but does not stress calendar arithmetic. Recent multimodal work ("Lost in Time")~\citep{saxena2025lost} examines visual clock/calendar reading for MLLMs but focuses on perception rather than textual calendrical reasoning.

\textbf{Evaluation contracts.} To separate formatting mistakes from semantic failures, we adopt a strict one-line output contract (ISO-8601 timestamps with explicit offsets; ambiguity pair \texttt{A||B} for DST repeats; \texttt{ABSTAIN} when no valid instant exists). This design, inspired by prior work emphasizing output contracts, disambiguates evaluation from genuine civil-time ambiguity and improves comparability across paraphrases and model variants.

\paragraph{Real-World Critical Infrastructure.}
Beyond NLP research, robust civil-time reasoning is a prerequisite for decision-support agents in critical infrastructure. In modern power grids and industrial IoT, while hardware relies on precision synchronization (IEEE~1588/PTP) \citep{liu2025smartgrid,guo2017gps1588,balakrishnan2023iiotptp}, the \emph{human-facing} layer involves interpreting heterogeneous logs across time zones. Operators using LLM assistants to triage grid disturbances or audit financial transactions \citep{lombardi2016financialtime} risk catastrophic errors if the model hallucinates during a DST transition or miscalculates a non-hour offset. Similarly, in digital forensics, reconstructing timelines from multi-source logs requires precise handling of local clock rules to detect drift or tampering \citep{haber1991timestamp,vanini2024clockcorrect}.

\section{CivilTime-Bench Benchmark}
\label{sec:dataset}

\subsection{Output Contract}
\label{sec:output-contract}

Our benchmark enforces strict, machine-checkable output to separate formatting errors from calendrical reasoning. Each item has a canonical answer in one of five forms:

\begin{enumerate}[itemsep=2pt, topsep=3pt, leftmargin=1.3em]
\item \texttt{ABSTAIN} --- No valid instant exists under the stated policy (e.g., impossible wall-time during DST skip).
\item \texttt{YYYY-MM-DD} or \texttt{YYYY-Www} --- Date or ISO week label for date-based and week-based tasks.
\item \texttt{YYYY-MM-DDThh:mm:ss±hh:mm} or \texttt{...Z} --- Single civil-time instant with explicit offset.
\item \texttt{A||B} --- Ambiguity pair (two valid instants; canonical gold orders them chronologically). Used for DST fall-back when a wall-time repeats.
\item \texttt{A..B} --- Interval notation: half-open span $[A,B)$ with end excluded. Used for time-range queries.
\end{enumerate}

The documented contract reserves \texttt{||} for ambiguity and \texttt{..} for intervals.
In the released data, however, synthetic INTERVAL gold is stored as \texttt{A||B} (all 334 10k items and all 34 API 1k items), whereas HumanTime-500 interval labels use \texttt{A..B}.
Before exact-match scoring, the evaluator canonicalises \texttt{..}~$\to$~\texttt{||}, so the two serialisations are equivalent and no partial credit is awarded.
Raw-output compliance (ContractOK) is measured separately: the v2 evaluator accepts a canonical one-line answer or a JSON object containing a canonical answer value, without extracting answers from surrounding text.
Extracted and normalised answers are validated separately before exact-match scoring.
Further gold-serialisation details are in Appendix~A.

\subsection{Dataset Composition}
\label{sec:dataset-composition}

We release two complementary datasets:

\begin{itemize}
\item \textbf{SyntheticTimeBench:} 60,011 train / 10,000 evaluation examples programmatically generated via seeded randomness, ensuring systematic coverage of rare civil-time edge cases. Generated over 2024--2026 IANA timezone rules via \texttt{pytz}. Training and evaluation examples use the same generators, with disjoint splits.
\item \textbf{HumanTime-500:} 500 LLM-authored, human-validated evaluation items designed to appear realistic (varied tone, length, context) for transfer testing. \emph{Evaluation-only} to assess generalization beyond synthetic templates.
\end{itemize}

\textbf{Coverage:} Eight civil-time stressors distributed across 11 task types (see Appendix~\ref{app:ds:type-counts}):
DST skip (spring-forward gaps), DST repeat (fall-back duplicates), DST ambiguity (explicit ambiguity policies), non-hour offsets (e.g., ±00:30, ±00:45), cross-timezone conversion, international date-line transitions, ISO week/year computation, business-day composition, and interval queries.

\subsection{Evaluation Metrics}
\label{sec:eval-metrics}

\paragraph{Correctness.}
We report \textbf{Exact Match (EM)} as the primary metric: strict string equality between the normalised model prediction and the canonical gold answer. No temporal tolerance or partial credit is used for EM.

\paragraph{Format \& Abstention.}
\textbf{ContractOK} measures raw-output compliance without answer extraction, accepting either a canonical one-line answer or a JSON object containing a canonical answer value, as implemented in the v2 evaluator. \textbf{FormatOK} measures whether the extracted and normalised prediction matches one of the five canonical forms: \texttt{ABSTAIN}, date or ISO week label (\texttt{YYYY-MM-DD} or \texttt{YYYY-Www}), timestamp+offset (\texttt{YYYY-MM-DDTHH:MM:SS$\pm$HH:MM}), ambiguity (\texttt{A\texttt{||}B}), and interval (\texttt{A..B}). Abstention correctness is evaluated separately.

\paragraph{Temporal Error.}
We report Median Temporal Error (MedTE) in seconds only over predictions for which temporal error is defined, together with temporal-error coverage (TEcov). A zero MedTE does not imply error-free performance over the full evaluation set.

\paragraph{Normalisation.}
Raw model completions often contain correct temporal content in non-canonical form (e.g., extra prose, missing seconds, non-standard offset). We apply a deterministic, label-agnostic normalisation that canonicalises surface forms but performs no calendrical computation. This protocol (raw $\to$ extracted/normalised $\to$ scored) reduces noise from serialisation artifacts while maintaining evaluation rigour.

\section{Experiments}
\label{sec:experiments}

\subsection{Models}
\label{sec:models}

We evaluate seven LLMs spanning open-source and API-based families: six instruction-tuned models and one base model.

\begin{itemize}
\item \textbf{Open-source:} Gemma-2-9B-IT~\citep{gemma2},
  Mistral-7B-Instruct-v0.3~\citep{jiang2023mistral7b},
  Qwen-2.5-7B-Instruct~\citep{qwen25},
  Llama-3.1-8B (base)~\citep{grattafiori2024llama3herdmodels},
  Phi-4~\citep{abdin2024phi4}.
\item \textbf{API-based:} Gemini~2.5~Flash~Lite~\citep{gemini25}, GPT-5-nano~\cite{openai_gpt5_nano_2026}.
\end{itemize}

All models are evaluated under the same scoring protocol, using temperature $T=0$ and deterministic decoding settings where supported. Prompting runs use $\texttt{max\_new\_tokens}=128$. Open-source models are evaluated on the full 10,000-item SyntheticTimeBench evaluation split, while API-based models use a 1,000-item subset. HumanTime-500 contains 500 evaluation items for all models.

\subsection{Experimental Conditions}
\label{sec:experimental-conditions}

We evaluate six experimental conditions:

\begin{table}[t]
  \centering
  \small
  \setlength{\tabcolsep}{4pt}
  \renewcommand{\arraystretch}{1.08}
  \begin{tabular}{@{}lcc@{}}
    \hline
    \textbf{Condition}
    & \textbf{Contract}
    & \textbf{Fine-tuned} \\
    \hline

    Baseline Zero-Shot
    & \xmark
    & \xmark \\

    Contract-Aware Zero-Shot
    & \cmark
    & \xmark \\

    ICL-2-shot
    & \cmark
    & \xmark \\

    ICL-4-shot
    & \cmark
    & \xmark \\

    ICL-8-shot
    & \cmark
    & \xmark \\

    QLoRA Fine-Tuning
    & \cmark
    & \cmark \\

    \hline
  \end{tabular}

  \caption{\textbf{Experimental conditions.}
  Contract indicates explicit output-contract instructions in the prompt. All runs use temperature $T=0$ and deterministic decoding settings where supported.}
  \label{tab:experimental-conditions}
\end{table}

\textbf{Zero-shot:} (i) \emph{Baseline} with minimal instruction and no contract restatement; (ii) \emph{Contract-aware} prepending a brief system preamble that restates the evaluation contract, civil-time conventions, and output requirements, but includes no exemplars.

\textbf{In-context learning:} Type-matched fixed exemplar banks selected from the training split (no dev/test leakage). For each category $t$, we construct a bank $B_t = \{e_{t,1}, \ldots, e_{t,8}\}$ with fixed seed and order. For a $k$-shot prompt, every test item of type $t$ receives the same first-$k$ exemplars. Exemplars are clean input--output pairs with no chain-of-thought. The affected LLaMA and Phi-4 conditions were rerun with distinct 2/4/8-shot prompts after correcting a prompt-construction error.

\textbf{Fine-tuning:} Parameter-efficient QLoRA (4-bit NF4 quantization, LoRA rank 32) was applied to five open-source models. Training uses 60,011 examples with completion-only loss (user prompt tokens masked). Hyperparameters were held constant across models; full configuration is provided in Appendix~\ref{app:sft-details}. Table~\ref{tab:finetune_results} reports the four models with valid synthetic evaluation results; unusable runs are excluded.

\subsection{Statistical Testing}
\label{sec:significance}

We report paired McNemar tests on the corrected per-example EM outcomes and 95\% Wilson confidence intervals for overall and per-task-type EM. McNemar tests use the continuity-corrected chi-squared statistic, with a two-sided exact binomial test when the number of discordant pairs is at most 50. Since we decode with temperature $T=0$, we conduct a single run per model and condition; confidence intervals reflect uncertainty across evaluation items rather than variation across runs. Missing predictions are counted as incorrect against the full gold split, and incomplete prediction coverage is reported alongside the affected results. Significance markers, where reported, indicate unadjusted McNemar test results: * ($p<0.05$), ** ($p<0.01$), *** ($p<0.001$).

\section{Results \& Analysis}
\label{sec:results}

\begin{figure*}[t!]
\centering
\includegraphics[width=0.94\textwidth]{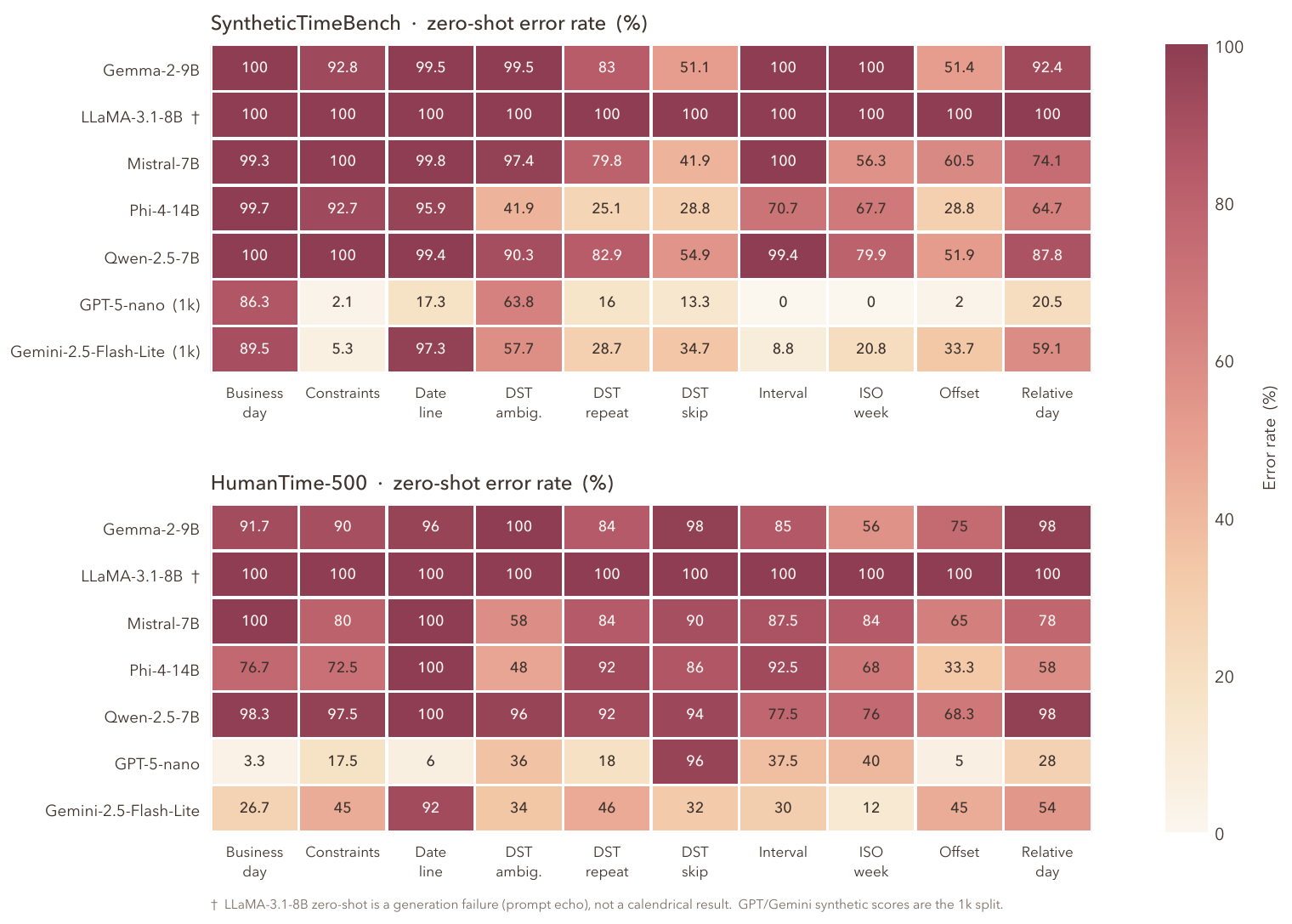}
\caption{\textbf{Baseline zero-shot error rates by task type and model.}
The heatmaps show per-type error rates ($100-\mathrm{EM}$, in percent) on SyntheticTimeBench (top) and HumanTime-500 (bottom), using the corrected evaluation results. Darker colors indicate higher error rates. Both panels display the same 10 task types; \textsc{WEEKSTART} is omitted from this visualization for comparability but remains included in the evaluation wherever available. Synthetic results use 10,000 items for open-source models and 1,000 for API models. LLaMA's zero-shot failure reflects prompt echoing rather than an isolated calendrical reasoning failure.}
\label{fig:error_heatmap}
\end{figure*}

\subsection{Prior Benchmarks as Context}
\label{sec:prior-bench}

To position civil-time reasoning within the broader temporal landscape, we run our models on representative prior datasets (TRAM, TimeDial, Test-of-Time), keeping their native formats. Results in Table~\ref{tab:temporal_bench_updated} show substantial variation across models and task formats, highlighting persistent difficulties with temporal reasoning and multi-step arithmetic and motivating our focus on civil-time calendar operations.

\begin{table}[t]
  \centering
  \renewcommand{\arraystretch}{1.08}
  \resizebox{\columnwidth}{!}{%
    \begin{tabular}{@{}lcccc@{}}
      \hline
      \textbf{Model}
      & \textbf{TRAM-M}
      & \textbf{TRAM-S}
      & \textbf{TimeDial}
      & \textbf{ToT} \\
      \hline

      Gemini-2.5-Flash-Lite
      & 95.2\%
      & 67.0\%
      & 92.9\%
      & 66.9\% \\

      GPT-5 Nano
      & 89.4\%
      & 43.6\%
      & 85.9\%
      & 32.0\% \\

      Gemma-2-9B
      & 40.2\%
      & 64.9\%
      & 86.1\%
      & -- \\

      LLaMA-3.1-8B
      & 50.0\%
      & 61.6\%
      & 40.7\%
      & 5.5\% \\

      Phi-4-14B
      & 50.6\%
      & 29.3\%
      & 89.4\%
      & 27.4\% \\

      Mistral-0.3v-7B
      & 40.4\%
      & 5.7\%
      & 54.4\%
      & 9.0\% \\

      Qwen-2.5-7B
      & 48.3\%
      & 7.3\%
      & 81.6\%
      & 16.2\% \\

      \hline
    \end{tabular}%
  }

  \caption{\textbf{Performance on prior temporal reasoning benchmarks.}
  Performance varies substantially across models and task formats, highlighting persistent difficulties with temporal reasoning and multi-step arithmetic. TRAM-M and TRAM-S denote the multiple-choice and short-answer settings, respectively.}
  \label{tab:temporal_bench_updated}
\end{table}

\textbf{Reproducibility and scoring note.} Table~\ref{tab:temporal_bench_updated} reports results from a fresh rerun of our earlier benchmark experiments, since artifacts from prior runs were not retained. The open-source models achieve relatively poor performance on some ToT settings, partly because the benchmark couples multi-step temporal arithmetic with strict output-format constraints. When a model deviates from the required format (e.g., missing fields or malformed structure), the official evaluation script cannot parse the prediction and the instance is scored as incorrect; in extreme cases this prevents a valid aggregate score.

\subsection{RQ1: Zero-Shot Performance}
\label{sec:zeroshot}

Can these LLMs handle civil-time reasoning without exemplars?

\textbf{Baseline vs.\ Contract-Aware Prompting.} Table~\ref{tab:zeroshot_results} compares baseline zero-shot prompting (minimal instruction) with contract-aware zero-shot prompting (explicit output contract + civil-time conventions, no exemplars). We evaluate on both SyntheticTimeBench and HumanTime-500 to measure performance on synthetic cases and more naturalistic language.

\textbf{Format compliance often improves, but accuracy gains are model-dependent.} On SyntheticTimeBench, FormatOK increases from 74.6\% to 97.8\% for Phi-4-14B and from 51.3\% to 96.4\% for Gemma-2-9B, but decreases from 58.7\% to 50.0\% for Mistral-7B. These changes do not consistently translate into exact-match gains. Phi-4's EM decreases from 40.8\% to 35.8\%, while Qwen-2.5-7B drops from 18.4\% to 8.4\%. Conversely, Gemma improves from 17.8\% to 26.8\%. On HumanTime-500, Phi-4 improves from 28.0\% to 45.0\% EM, whereas Qwen gains only 3.0 percentage points (10.2\% to 13.2\%).

\begin{table*}[t]
  \centering
  \scriptsize
  \setlength{\tabcolsep}{3pt}
  \renewcommand{\arraystretch}{1.1}

  \begin{tabular}{@{}l*{10}{c}@{}}
    \hline
    \multirow{3}{*}{\textbf{Model}}
    & \multicolumn{5}{c}{\textbf{SyntheticTimeBench (10k / 1k)$^{\dagger}$}}
    & \multicolumn{5}{c}{\textbf{HumanTime-500}} \\
    \cline{2-6}
    \cline{7-11}

    & \multicolumn{2}{c}{\textbf{EM ($\uparrow$)}}
    & \multicolumn{2}{c}{\textbf{FormatOK ($\uparrow$)}}
    & \textbf{$\Delta$EM}
    & \multicolumn{2}{c}{\textbf{EM ($\uparrow$)}}
    & \multicolumn{2}{c}{\textbf{FormatOK ($\uparrow$)}}
    & \textbf{$\Delta$EM} \\
    \cline{2-3}
    \cline{4-5}
    \cline{6-6}
    \cline{7-8}
    \cline{9-10}
    \cline{11-11}

    & \textbf{ZS}
    & \textbf{CA}
    & \textbf{ZS}
    & \textbf{CA}
    & \textbf{CA$-$ZS}
    & \textbf{ZS}
    & \textbf{CA}
    & \textbf{ZS}
    & \textbf{CA}
    & \textbf{CA$-$ZS} \\
    \hline

    Gemma-2-9B
    & 17.8
    & \textbf{26.8}$^{***}$
    & 51.3
    & \textbf{96.4}
    & \textbf{+9.0}
    & 12.0
    & \textbf{30.2}$^{***}$
    & 56.8
    & \textbf{100.0}
    & \textbf{+18.2} \\

    Gemini-2.5-Flash-Lite$^{\dagger}$
    & \textbf{57.3}
    & 50.4$^{***}$
    & \textbf{95.0}
    & 86.4
    & $-$6.9
    & 57.8
    & \textbf{63.0}$^{*}$
    & 96.8
    & \textbf{99.4}
    & \textbf{+5.2} \\

    GPT-5-nano$^{\dagger}$
    & 76.9
    & \textbf{78.0}
    & \textbf{100.0}
    & 99.6
    & \textbf{+1.1}
    & \textbf{73.2}
    & 73.0
    & \textbf{95.6}
    & 95.0
    & $-$0.2 \\

    LLaMA-3.1-8B$^{\ddagger}$
    & 0.0
    & \textbf{31.8}$^{***}$
    & 0.0
    & \textbf{97.6}
    & \textbf{+31.8}
    & 0.0
    & \textbf{28.8}$^{***}$
    & 0.0
    & \textbf{99.6}
    & \textbf{+28.8} \\

    Mistral-7B
    & \textbf{19.0}
    & 16.2$^{***}$
    & \textbf{58.7}
    & 50.0
    & $-$2.8
    & 17.0
    & \textbf{20.8}$^{*}$
    & 56.2
    & \textbf{74.0}
    & \textbf{+3.8} \\

    Phi-4-14B
    & \textbf{40.8}
    & 35.8$^{***}$
    & 74.6
    & \textbf{97.8}
    & $-$5.0
    & 28.0
    & \textbf{45.0}$^{***}$
    & 54.0
    & \textbf{100.0}
    & \textbf{+17.0} \\

    Qwen-2.5-7B
    & \textbf{18.4}
    & 8.4$^{***}$
    & 45.2
    & \textbf{51.3}
    & $-$10.0
    & 10.2
    & \textbf{13.2}
    & 29.4
    & \textbf{64.2}
    & \textbf{+3.0} \\
    \hline
  \end{tabular}

  \caption{\textbf{Contracts improve formatting but not reasoning.}
  Values are percentages. Stars on CA EM are unadjusted paired McNemar tests
  versus that model's own ZS baseline:
  $^{*}p<0.05$, $^{**}p<0.01$, $^{***}p<0.001$.
  $^{\dagger}$GPT/Gemini synthetic scores use the 1{,}000-item subset;
  open models use 10{,}000 items. HumanTime-500 uses 500 items for all models.
  $^{\ddagger}$LLaMA zero-shot is a generation failure (prompt echo), not a
  calendrical result. FormatOK often rises under contract-aware prompting,
  but exact-match changes are model-dependent and can be negative.
  ZS and CA denote baseline zero-shot and contract-aware zero-shot,
  respectively.}
  \label{tab:zeroshot_results}
\end{table*}

\textbf{Frontier models exhibit stronger zero-shot performance.} GPT-5-nano reaches 76.9\% EM on the 1k synthetic subset and 73.2\% on HumanTime-500, with contract-aware prompting yielding 78.0\% and 73.0\%, respectively. Gemini-2.5-Flash-Lite decreases from 57.3\% to 50.4\% EM on the 1k synthetic subset but improves from 57.8\% to 63.0\% on HumanTime-500. Thus, civil-time reasoning difficulty varies substantially across models and prompting conditions.

\textbf{Task-specific failures.} Figure~\ref{fig:error_heatmap} presents baseline zero-shot error rates across the 10 task types shared by the visualized splits. Performance varies considerably by task and model. For example, GPT-5-nano achieves relatively low error rates on several task types but struggles with DST skips on HumanTime-500. The heatmaps also illustrate that task-level weaknesses differ across models and datasets, rather than following a single universal difficulty ordering. LLaMA-3.1-8B's uniformly poor zero-shot results reflect a generation failure involving extensive prompt echoing and should not be interpreted as a clean measure of calendrical reasoning capability.

\textbf{Key insight: Formatting and reasoning are distinct bottlenecks.} The differences between FormatOK and EM show that producing an extractable, canonical answer does not guarantee correct calendrical reasoning. Raw-output compliance is measured separately by ContractOK (Section~\ref{sec:eval-metrics}). Contract restatement alone cannot reliably resolve calendrical errors, motivating in-context learning and fine-tuning (Sections~\ref{sec:icl} and~\ref{sec:finetuning}).

\subsection{RQ2: In-Context Learning with Type-Matched Exemplars}
\label{sec:icl}

How far can exemplars take us? We construct fixed exemplar banks per task type and test $k \in \{2, 4, 8\}$ demonstrations. All prompts use the contract-aware system preamble from Section~\ref{sec:zeroshot}.

\textbf{In-context learning provides substantial but model-dependent gains.}
Table~\ref{tab:icl_results} shows that adding 2-shot ICL improves EM for several open models.
On HumanTime-500, Gemma-2-9B rises from 12.0\% zero-shot to 36.8\% at 2-shot and 41.4\% at 8-shot.
Qwen-2.5-7B improves from 10.2\% to 41.0\% and 51.8\%, respectively.
LLaMA-3.1-8B reaches 36.8\%, 39.6\%, and 42.2\% at 2/4/8-shot, and Phi-4-14B reaches 42.2\%, 41.0\%, and 46.4\%.
Additional exemplars do not always yield monotonic gains: Mistral changes only from 36.0\% at 4-shot to 35.8\% at 8-shot, while Phi-4 increases from 41.0\% to 46.4\%.

\begin{table}[t]
  \centering
  \scriptsize
  \setlength{\tabcolsep}{4pt}
  \renewcommand{\arraystretch}{1.1}

  \resizebox{\columnwidth}{!}{%
    \begin{tabular}{@{}lcccccc@{}}
      \hline
      \multirow{2}{*}{\textbf{Model}}
      & \multicolumn{2}{c}{\textbf{2-Shot}}
      & \multicolumn{2}{c}{\textbf{4-Shot}}
      & \multicolumn{2}{c}{\textbf{8-Shot}} \\
      \cline{2-3}
      \cline{4-5}
      \cline{6-7}

      & \textbf{EM}
      & \textbf{FO}
      & \textbf{EM}
      & \textbf{FO}
      & \textbf{EM}
      & \textbf{FO} \\
      \hline

      Gemini-2.5-Flash-Lite$^{\dagger}$
      & 71.2$^{***}$
      & 95.6
      & 66.6$^{***}$
      & 91.2
      & 63.6$^{*}$
      & 82.4 \\

      Gemma-2-9B
      & 36.8$^{***}$
      & 100.0
      & 38.8$^{***}$
      & 100.0
      & 41.4$^{***}$
      & 100.0 \\

      GPT-5-nano$^{\dagger}$
      & \textbf{76.6}
      & 91.8
      & \textbf{74.0}
      & 88.8
      & \textbf{67.8}$^{*}$
      & 80.4 \\

      LLaMA-3.1-8B
      & 36.8$^{***}$
      & 100.0
      & 39.6$^{***}$
      & 99.8
      & 42.2$^{***}$
      & 100.0 \\

      Mistral-7B
      & 33.8$^{***}$
      & 99.6
      & 36.0$^{***}$
      & 99.4
      & 35.8$^{***}$
      & 100.0 \\

      Phi-4-14B
      & 42.2$^{***}$
      & 100.0
      & 41.0$^{***}$
      & 100.0
      & 46.4$^{***}$
      & 95.8 \\

      Qwen-2.5-7B
      & 41.0$^{***}$
      & 96.2
      & 41.2$^{***}$
      & 92.8
      & 51.8$^{***}$
      & 93.8 \\
      \hline
    \end{tabular}%
  }

  \caption{\textbf{In-context learning helps, but gains are model-dependent.}
  Exact match (EM) and FormatOK (FO) on HumanTime-500 for
  $k$-shot ($k \in \{2,4,8\}$) type-matched ICL.
  Values are percentages.
  Stars on EM are unadjusted paired McNemar tests versus that model's
  own zero-shot baseline:
  $^{*}p<0.05$, $^{**}p<0.01$, $^{***}p<0.001$.
  $^{\dagger}$GPT/Gemini few-shot runs have incomplete coverage
  ($n{=}478/456/412$ for 2/4/8-shot); missing predictions are counted
  as incorrect against the full 500-item gold split.
  Several open models continue to improve at 8-shot, while GPT and Gemini
  decline as $k$ increases.}
  \label{tab:icl_results}
\end{table}

\textbf{Gains vary across model families and shot counts.}
GPT-5-nano achieves 76.6\% EM at 2-shot but decreases to 67.8\% at 8-shot, while Gemini-2.5-Flash-Lite changes from 71.2\% to 63.6\%.
The API few-shot evaluations have incomplete prediction coverage ($n{=}478/456/412$ for 2/4/8-shot), with missing predictions counted as incorrect against the full 500-item gold split.
Paired McNemar tests indicate significant 2-to-8-shot improvements for Qwen ($p<0.001$) and LLaMA ($p<0.01$), but not for Mistral.

\begin{figure}[t]
\centering
\includegraphics[width=0.90\columnwidth]{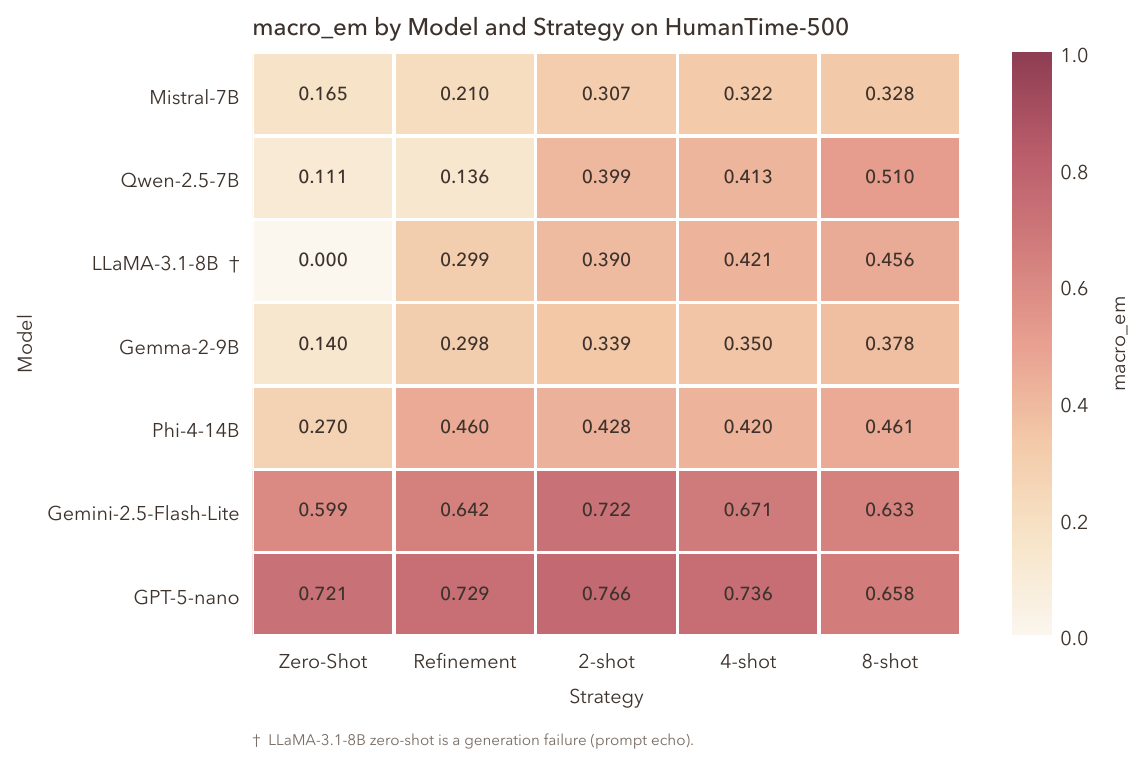}
\caption{\textbf{Macro-averaged exact match across prompting strategies on HumanTime-500.}
Results for the seven evaluated models under baseline zero-shot, contract-aware zero-shot (refinement), and 2/4/8-shot in-context learning. Each cell reports the unweighted mean of per-task-type EM scores (\texttt{macro\_em}), rather than the instance-weighted overall EM reported in Tables~\ref{tab:zeroshot_results} and~\ref{tab:icl_results}. Performance gains are model-dependent and are not consistently monotonic with additional exemplars. LLaMA's zero-shot result reflects a prompt-echo generation failure.}
\label{fig:strategy_macro_em}
\end{figure}

\textbf{Prompting strategies exhibit different effects across models.}
Figure~\ref{fig:strategy_macro_em} complements the instance-level EM results with macro-averaged performance across task types.
Qwen's macro-EM increases from 0.111 at zero-shot to 0.510 at 8-shot, whereas GPT-5-nano peaks at 0.766 with 2-shot and decreases to 0.658 at 8-shot.
Gemini similarly peaks at 2-shot (0.722), while Mistral's gains become smaller at higher shot counts.
These patterns indicate that prompting benefits depend on the model and task distribution, rather than increasing uniformly with the number of demonstrations.

\textbf{Format compliance does not ensure correctness.}
Although explicit instructions and type-matched exemplars can improve output structure, substantial gaps between FormatOK and EM remain in several settings.
LLaMA few-shot evaluations, for example, reach 36.8\% EM at 2-shot and 42.2\% at 8-shot, a large improvement over its failed zero-shot generation, while still leaving many civil-time problems unresolved.
These results suggest that examples help beyond surface formatting, but their benefits depend on the model and number of demonstrations.

\subsection{RQ3: Fine-Tuning for Calendrical Algebra}
\label{sec:finetuning}

Can models learn civil-time rules with parameter-efficient adaptation? We fine-tune five open-source models with QLoRA (4-bit, rank 32) on the full 60,011-example training set. Table~\ref{tab:finetune_results} reports four models with usable synthetic evaluation results, comparing zero-shot and fine-tuned performance on SyntheticTimeBench (10k) and HumanTime-500 where valid predictions are available.

\textbf{Fine-tuning substantially improves synthetic performance.} Table~\ref{tab:finetune_results} shows that fine-tuning achieves 74.6--86.6\% EM on SyntheticTimeBench. LLaMA-3.1-8B reaches 86.6\%, Mistral-7B 81.4\%, and Gemma-2-9B and Phi-4-14B each reach 74.6\%. For models with functioning zero-shot baselines, these represent substantial improvements. Paired McNemar tests indicate significant gains ($p<0.001$) on the synthetic split. LLaMA's improvement is interpreted separately because its zero-shot baseline suffered a generation failure.

\begin{table}[t]
  \centering
  \scriptsize
  \setlength{\tabcolsep}{3.5pt}
  \renewcommand{\arraystretch}{1.1}

  \begin{tabular}{@{}lcccc@{}}
    \hline
    \multirow{2}{*}{\textbf{Model}}
    & \multicolumn{2}{c}{\textbf{SyntheticTimeBench (10k)}}
    & \multicolumn{2}{c}{\textbf{HumanTime-500}} \\
    \cline{2-3}
    \cline{4-5}
    & \textbf{FT EM}
    & \textbf{$\Delta$EM}
    & \textbf{FT EM}
    & \textbf{$\Delta$EM} \\
    \hline

    Gemma-2-9B
    & 74.6$^{***}$
    & +56.8
    & ---
    & --- \\

    LLaMA-3.1-8B
    & \textbf{86.6}$^{***}$
    & +86.6
    & 39.6$^{***}$
    & +39.6 \\

    Mistral-7B
    & 81.4$^{***}$
    & +62.4
    & \textbf{45.2}$^{***}$
    & +28.2 \\

    Phi-4-14B
    & 74.6$^{***}$
    & +33.8
    & 36.6$^{**}$
    & +8.6 \\
    \hline
  \end{tabular}

  \caption{\textbf{Fine-tuning substantially improves exact match over
  zero-shot prompting.}
  Values are percentages.
  FT EM is fine-tuned exact match; $\Delta$EM is FT minus that model's
  Table~\ref{tab:zeroshot_results} zero-shot EM.
  Stars on FT EM are unadjusted paired McNemar tests versus that model's
  own zero-shot predictions:
  $^{*}p<0.05$, $^{**}p<0.01$, $^{***}p<0.001$.
  Gemma's HumanTime-500 fine-tuning dump is invalid and is omitted.
  Missing predictions are counted as incorrect against the full gold split
  (Gemma/Mistral synthetic coverage is 8{,}532/10{,}000; Mistral
  HumanTime-500 coverage is 432/500).
  Qwen is omitted: its synthetic fine-tuning IDs do not match the gold split.}
  \label{tab:finetune_results}
\end{table}

\textbf{Natural language variation remains challenging after fine-tuning.} On HumanTime-500, Mistral reaches 45.2\% EM, LLaMA 39.6\%, and Phi-4 36.6\%. Their corresponding synthetic-to-naturalistic gaps are 36.2, 47.0, and 38.0 percentage points, respectively. These gaps indicate that strong performance on synthetic examples generated from the training distribution does not guarantee generalization to more naturalistic inputs. HumanTime-500 remains a limited, LLM-authored and human-validated evaluation set, so broader generalization claims require caution.

\textbf{Coverage and temporal-error diagnostics.} Missing predictions are counted as incorrect against the full gold split. Gemma and Mistral have 8,532/10,000 predictions in their synthetic fine-tuning evaluations, and Mistral has 432/500 on HumanTime-500. Gemma's HumanTime-500 fine-tuning dump is unusable and excluded. Qwen's HumanTime-500 fine-tuning result is reported separately with incomplete coverage; no valid matched synthetic evaluation is available. Median Temporal Error is computed only over eligible, parsable predictions: on HumanTime-500, Mistral has a MedTE of 1,800 seconds and Phi-4 of 600 seconds. A zero median for other runs does not imply error-free performance overall. Despite substantial synthetic gains, these results highlight the need for improved transfer beyond synthetic templates.

\subsection{RQ4: Error Analysis}
\label{sec:error-analysis}

Where do models still fail? We categorize common civil-time failure modes:

\begin{enumerate}[itemsep=2pt, leftmargin=1.3em]
\item \textbf{Non-hour offset errors:} Models incorrectly round or apply fractional UTC offsets (e.g., +05:45 in Asia/Kathmandu), yielding wrong local times.
\item \textbf{DST branch selection:} When fall-back repeats a wall-time, models select the wrong occurrence (e.g., \texttt{first} vs.\ \texttt{second}), violating the stated policy.
\item \textbf{Date-line direction errors:} UTC $\leftrightarrow$ local conversion errors across the international date line (e.g., UTC$-$10 vs.\ UTC+14), producing incorrect calendar dates.
\item \textbf{Business-day composition:} Incorrect weekend skipping or business-day arithmetic.
\item \textbf{ISO week anomalies:} Edge cases where the ISO week-year differs from the calendar year (e.g., 2021-01-01 belongs to 2020-W53).
\end{enumerate}

\section{Discussion}
\label{sec:discussion}

\paragraph{Three-stage intervention hierarchy.}
Our results reveal three complementary interventions: (i)~explicit contracts improve formatting but not consistently reasoning accuracy; (ii)~type-matched exemplars improve exact match, with model-dependent gains; (iii)~fine-tuning substantially improves synthetic performance, with limited transfer to naturalistic inputs. These findings suggest a three-stage approach:

\begin{enumerate}[itemsep=2pt, leftmargin=1.3em]
\item \textbf{Stage 1 (Formatting):}
Restate the output contract.
On SyntheticTimeBench, Gemma's FormatOK rises from 51.3\% to 96.4\% and EM from 17.8\% to 26.8\%, while other models lose EM despite better formatting: serialization and correctness are distinct.
\item \textbf{Stage 2 (Reasoning via exemplars):}
Provide type-matched few-shot exemplars.
On HumanTime-500, Qwen improves by 41.6 points and Mistral by 18.8 points from zero-shot to 8-shot ICL; additional exemplars do not help all models.
\item \textbf{Stage 3 (Adaptation via fine-tuning):}
Fine-tune on civil-time examples.
QLoRA reaches 74.6--86.6\% EM on synthetic evaluation but only 36.6--45.2\% on HumanTime-500 for models with valid results on both splits.
\end{enumerate}

\paragraph{Implications.}
\textbf{Zero-shot failures are not necessarily architectural limitations.}
Fine-tuning yields large synthetic EM gains, but the remaining synthetic-to-naturalistic gap shows that template exposure is not enough for robust generalization.

\textbf{Formatting alone is insufficient.}
Contract-aware prompting often improves FormatOK with inconsistent EM effects.
Temporal benchmarks should report raw-output contract compliance, extracted-answer format validity, and exact match separately, rather than conflating formatting with reasoning.

\textbf{Type-matched exemplars are effective but model-dependent.}
Qwen benefits strongly from 8-shot ICL, whereas Mistral saturates beyond 4-shot, so extra demonstrations should not be assumed to help uniformly.


\section*{Limitations}
\label{sec:Limitations}

Our study prioritizes clarity and reproducibility. The benchmark targets calendrical operations under Gregorian assumptions and curated time zones/years, excluding irregular schedules and region/market-specific \emph{floating holidays}. Synthetic training and evaluation share generators, limiting generalization beyond these templates. HumanTime-500 provides naturalistic evaluation with 500 LLM-authored, human-validated items, limiting representativeness and per-category precision. Results cover seven models (six instruction-tuned and one base) in single runs, with model-dependent outcomes and incomplete prediction coverage. The synthetic-to-naturalistic performance gap limits claims of robust calendrical reasoning. Ground truth relies on pinned \texttt{tzdata} and libraries, while limited tool-calling support may reduce deployment transferability. These choices support reproducibility but constrain generality.

We aim to integrate external APIs for region- and market-specific \emph{floating holidays} (absent from the current release), expand time-zone and historical-rule coverage, add non-Gregorian variants, enlarge naturalistic evaluation with human-authored and multilingual paraphrases, evaluate larger reasoning-centric models under matched compute, and test interactive flows alongside cost/latency trade-offs. These steps aim to improve robustness and real-world applicability.

\section{Conclusion}
We introduced CivilTime-Bench to evaluate language models on civil-time reasoning.
Experiments across seven models show large gaps between format compliance and exact calendrical correctness.
Fine-tuning raises synthetic EM to 74.6--86.6\%, but HumanTime-500 remains at 36.6--45.2\% for models with valid results on both splits.
Reliable civil-time reasoning still requires evaluation beyond formatting, and synthetic gains do not guarantee transfer.

\bibliography{custom}

@misc{wang2024tram,
      title={TRAM: Benchmarking Temporal Reasoning for Large Language Models}, 
      author={Yuqing Wang and Yun Zhao},
      year={2024},
      eprint={2310.00835},
      archivePrefix={arXiv},
      primaryClass={cs.CL},
      url={https://arxiv.org/abs/2310.00835}, 
}

@misc{fatemi2024test,
      title={Test of Time: A Benchmark for Evaluating LLMs on Temporal Reasoning}, 
      author={Bahare Fatemi and Mehran Kazemi and Anton Tsitsulin and Karishma Malkan and Jinyeong Yim and John Palowitch and Sungyong Seo and Jonathan Halcrow and Bryan Perozzi},
      year={2024},
      eprint={2406.09170},
      archivePrefix={arXiv},
      primaryClass={cs.CL},
      url={https://arxiv.org/abs/2406.09170}, 
}

@misc{qin2021timedial,
      title={TIMEDIAL: Temporal Commonsense Reasoning in Dialog}, 
      author={Lianhui Qin and Aditya Gupta and Shyam Upadhyay and Luheng He and Yejin Choi and Manaal Faruqui},
      year={2021},
      eprint={2106.04571},
      archivePrefix={arXiv},
      primaryClass={cs.CL},
      url={https://arxiv.org/abs/2106.04571}, 
}

@misc{ning2020torque,
      title={TORQUE: A Reading Comprehension Dataset of Temporal Ordering Questions}, 
      author={Qiang Ning and Hao Wu and Rujun Han and Nanyun Peng and Matt Gardner and Dan Roth},
      year={2020},
      eprint={2005.00242},
      archivePrefix={arXiv},
      primaryClass={cs.CL},
      url={https://arxiv.org/abs/2005.00242}, 
}

@misc{uzzaman2013semeval,
      title={TempEval-3: Evaluating Events, Time Expressions, and Temporal Relations}, 
      author={Naushad UzZaman and Hector Llorens and James Allen and Leon Derczynski and Marc Verhagen and James Pustejovsky},
      year={2014},
      eprint={1206.5333},
      archivePrefix={arXiv},
      primaryClass={cs.CL},
      url={https://arxiv.org/abs/1206.5333}, 
}

@misc{saxena2025lost,
      title={Lost in Time: Clock and Calendar Understanding Challenges in Multimodal LLMs}, 
      author={Rohit Saxena and Aryo Pradipta Gema and Pasquale Minervini},
      year={2025},
      eprint={2502.05092},
      archivePrefix={arXiv},
      primaryClass={cs.CV},
      url={https://arxiv.org/abs/2502.05092}, 
}

@misc{gemma2,
      title={Gemma 2: Improving Open Language Models at a Practical Size}, 
      author={Gemma Team and Morgane Riviere and Shreya Pathak and Pier Giuseppe Sessa and Cassidy Hardin and Surya Bhupatiraju and Léonard Hussenot and Thomas Mesnard and Bobak Shahriari and Alexandre Ramé and Johan Ferret and Peter Liu and Pouya Tafti and Abe Friesen and Michelle Casbon and Sabela Ramos and Ravin Kumar and Charline Le Lan and Sammy Jerome and Anton Tsitsulin and Nino Vieillard and Piotr Stanczyk and Sertan Girgin and Nikola Momchev and Matt Hoffman and Shantanu Thakoor and Jean-Bastien Grill and Behnam Neyshabur and Olivier Bachem and Alanna Walton and Aliaksei Severyn and Alicia Parrish and Aliya Ahmad and Allen Hutchison and Alvin Abdagic and Amanda Carl and Amy Shen and Andy Brock and Andy Coenen and Anthony Laforge and Antonia Paterson and Ben Bastian and Bilal Piot and Bo Wu and Brandon Royal and Charlie Chen and Chintu Kumar and Chris Perry and Chris Welty and Christopher A. Choquette-Choo and Danila Sinopalnikov and David Weinberger and Dimple Vijaykumar and Dominika Rogozińska and Dustin Herbison and Elisa Bandy and Emma Wang and Eric Noland and Erica Moreira and Evan Senter and Evgenii Eltyshev and Francesco Visin and Gabriel Rasskin and Gary Wei and Glenn Cameron and Gus Martins and Hadi Hashemi and Hanna Klimczak-Plucińska and Harleen Batra and Harsh Dhand and Ivan Nardini and Jacinda Mein and Jack Zhou and James Svensson and Jeff Stanway and Jetha Chan and Jin Peng Zhou and Joana Carrasqueira and Joana Iljazi and Jocelyn Becker and Joe Fernandez and Joost van Amersfoort and Josh Gordon and Josh Lipschultz and Josh Newlan and Ju-yeong Ji and Kareem Mohamed and Kartikeya Badola and Kat Black and Katie Millican and Keelin McDonell and Kelvin Nguyen and Kiranbir Sodhia and Kish Greene and Lars Lowe Sjoesund and Lauren Usui and Laurent Sifre and Lena Heuermann and Leticia Lago and Lilly McNealus and Livio Baldini Soares and Logan Kilpatrick and Lucas Dixon and Luciano Martins and Machel Reid and Manvinder Singh and Mark Iverson and Martin Görner and Mat Velloso and Mateo Wirth and Matt Davidow and Matt Miller and Matthew Rahtz and Matthew Watson and Meg Risdal and Mehran Kazemi and Michael Moynihan and Ming Zhang and Minsuk Kahng and Minwoo Park and Mofi Rahman and Mohit Khatwani and Natalie Dao and Nenshad Bardoliwalla and Nesh Devanathan and Neta Dumai and Nilay Chauhan and Oscar Wahltinez and Pankil Botarda and Parker Barnes and Paul Barham and Paul Michel and Pengchong Jin and Petko Georgiev and Phil Culliton and Pradeep Kuppala and Ramona Comanescu and Ramona Merhej and Reena Jana and Reza Ardeshir Rokni and Rishabh Agarwal and Ryan Mullins and Samaneh Saadat and Sara Mc Carthy and Sarah Cogan and Sarah Perrin and Sébastien M. R. Arnold and Sebastian Krause and Shengyang Dai and Shruti Garg and Shruti Sheth and Sue Ronstrom and Susan Chan and Timothy Jordan and Ting Yu and Tom Eccles and Tom Hennigan and Tomas Kocisky and Tulsee Doshi and Vihan Jain and Vikas Yadav and Vilobh Meshram and Vishal Dharmadhikari and Warren Barkley and Wei Wei and Wenming Ye and Woohyun Han and Woosuk Kwon and Xiang Xu and Zhe Shen and Zhitao Gong and Zichuan Wei and Victor Cotruta and Phoebe Kirk and Anand Rao and Minh Giang and Ludovic Peran and Tris Warkentin and Eli Collins and Joelle Barral and Zoubin Ghahramani and Raia Hadsell and D. Sculley and Jeanine Banks and Anca Dragan and Slav Petrov and Oriol Vinyals and Jeff Dean and Demis Hassabis and Koray Kavukcuoglu and Clement Farabet and Elena Buchatskaya and Sebastian Borgeaud and Noah Fiedel and Armand Joulin and Kathleen Kenealy and Robert Dadashi and Alek Andreev},
      year={2024},
      eprint={2408.00118},
      archivePrefix={arXiv},
      primaryClass={cs.CL},
      url={https://arxiv.org/abs/2408.00118}, 
}

@misc{abdin2024phi4,
      title={Phi-4 Technical Report}, 
      author={Marah Abdin and Jyoti Aneja and Harkirat Behl and Sébastien Bubeck and Ronen Eldan and Suriya Gunasekar and Michael Harrison and Russell J. Hewett and Mojan Javaheripi and Piero Kauffmann and James R. Lee and Yin Tat Lee and Yuanzhi Li and Weishung Liu and Caio C. T. Mendes and Anh Nguyen and Eric Price and Gustavo de Rosa and Olli Saarikivi and Adil Salim and Shital Shah and Xin Wang and Rachel Ward and Yue Wu and Dingli Yu and Cyril Zhang and Yi Zhang},
      year={2024},
      eprint={2412.08905},
      archivePrefix={arXiv},
      primaryClass={cs.CL},
      url={https://arxiv.org/abs/2412.08905}, 
}

@misc{jiang2023mistral7b,
      title={Mistral 7B}, 
      author={Albert Q. Jiang and Alexandre Sablayrolles and Arthur Mensch and Chris Bamford and Devendra Singh Chaplot and Diego de las Casas and Florian Bressand and Gianna Lengyel and Guillaume Lample and Lucile Saulnier and Lélio Renard Lavaud and Marie-Anne Lachaux and Pierre Stock and Teven Le Scao and Thibaut Lavril and Thomas Wang and Timothée Lacroix and William El Sayed},
      year={2023},
      eprint={2310.06825},
      archivePrefix={arXiv},
      primaryClass={cs.CL},
      url={https://arxiv.org/abs/2310.06825}, 
}

@misc{qwen25,
      title={Qwen2.5 Technical Report}, 
      author={Qwen and An Yang and Baosong Yang and Beichen Zhang and Binyuan Hui and Bo Zheng and Bowen Yu and Chengyuan Li and Dayiheng Liu and Fei Huang and Haoran Wei and Huan Lin and Jian Yang and Jianhong Tu and Jianwei Zhang and Jianxin Yang and Jiaxi Yang and Jingren Zhou and Junyang Lin and Kai Dang and Keming Lu and Keqin Bao and Kexin Yang and Le Yu and Mei Li and Mingfeng Xue and Pei Zhang and Qin Zhu and Rui Men and Runji Lin and Tianhao Li and Tianyi Tang and Tingyu Xia and Xingzhang Ren and Xuancheng Ren and Yang Fan and Yang Su and Yichang Zhang and Yu Wan and Yuqiong Liu and Zeyu Cui and Zhenru Zhang and Zihan Qiu},
      year={2025},
      eprint={2412.15115},
      archivePrefix={arXiv},
      primaryClass={cs.CL},
      url={https://arxiv.org/abs/2412.15115}, 
}

@misc{gemini25,
      title={Gemini 2.5: Pushing the Frontier with Advanced Reasoning, Multimodality, Long Context, and Next Generation Agentic Capabilities}, 
      author={Gheorghe Comanici and Eric Bieber and Mike Schaekermann and Ice Pasupat and Noveen Sachdeva and Inderjit Dhillon and Marcel Blistein and Ori Ram and Dan Zhang and Evan Rosen and Luke Marris and Sam Petulla and Colin Gaffney and Asaf Aharoni and Nathan Lintz and Tiago Cardal Pais and Henrik Jacobsson and Idan Szpektor and Nan-Jiang Jiang and Krishna Haridasan and Ahmed Omran and Nikunj Saunshi and Dara Bahri and Gaurav Mishra and Eric Chu and Toby Boyd and Brad Hekman and Aaron Parisi and Chaoyi Zhang and Kornraphop Kawintiranon and Tania Bedrax-Weiss and Oliver Wang and Ya Xu and Ollie Purkiss and Uri Mendlovic and Ilaï Deutel and Nam Nguyen and Adam Langley and Flip Korn and Lucia Rossazza and Alexandre Ramé and Sagar Waghmare and Helen Miller and Nathan Byrd and Ashrith Sheshan and Raia Hadsell and Sangnie Bhardwaj and Pawel Janus and Tero Rissa and Dan Horgan and Alvin Abdagic and Lior Belenki and James Allingham and Anima Singh and Theo Guidroz and Srivatsan Srinivasan and Herman Schmit and Kristen Chiafullo and Andre Elisseeff and Nilpa Jha and Prateek Kolhar and Leonard Berrada and Frank Ding and Xiance Si and Shrestha Basu Mallick and Franz Och and Sofia Erell and Eric Ni and Tejasi Latkar and Sherry Yang and Petar Sirkovic and Ziqiang Feng and Robert Leland and Rachel Hornung and Gang Wu and Charles Blundell and Hamidreza Alvari and Po-Sen Huang and Cathy Yip and Sanja Deur and Li Liu and Gabriela Surita and Pablo Duque and Dima Damen and Johnson Jia and Arthur Guez and Markus Mircea and Animesh Sinha and Alberto Magni and Paweł Stradomski and Tal Marian and Vlado Galić and Wenhu Chen and Hisham Husain and Achintya Singhal and Dominik Grewe and François-Xavier Aubet and Shuang Song and Lorenzo Blanco and Leland Rechis and Lewis Ho and Rich Munoz and Kelvin Zheng and Jessica Hamrick and Kevin Mather and Hagai Taitelbaum and Eliza Rutherford and Yun Lei and Kuangyuan Chen and Anand Shukla and Erica Moreira and Eric Doi and Berivan Isik and Nir Shabat and Dominika Rogozińska and Kashyap Kolipaka and Jason Chang and Eugen Vušak and Srinivasan Venkatachary and Shadi Noghabi and Tarun Bharti and Younghoon Jun and Aleksandr Zaks and Simon Green and Jeshwanth Challagundla and William Wong and Muqthar Mohammad and Dean Hirsch and Yong Cheng and Iftekhar Naim and Lev Proleev and Damien Vincent and Aayush Singh and Maxim Krikun and Dilip Krishnan and Zoubin Ghahramani and Aviel Atias and Rajeev Aggarwal and Christo Kirov and Dimitrios Vytiniotis and Christy Koh and Alexandra Chronopoulou and Pawan Dogra and Vlad-Doru Ion and Gladys Tyen and Jason Lee and Felix Weissenberger and Trevor Strohman and Ashwin Balakrishna and Jack Rae and Marko Velic and Raoul de Liedekerke and Oded Elyada and Wentao Yuan and Canoee Liu and Lior Shani and Sergey Kishchenko and Bea Alessio and Yandong Li and Richard Song and Sam Kwei and Orion Jankowski and Aneesh Pappu and Youhei Namiki and Yenai Ma and Nilesh Tripuraneni and Colin Cherry and Marissa Ikonomidis and Yu-Cheng Ling and Colin Ji and Beka Westberg and Auriel Wright and Da Yu and David Parkinson and Swaroop Ramaswamy and Jerome Connor and Soheil Hassas Yeganeh and Snchit Grover and George Kenwright and Lubo Litchev and Chris Apps and Alex Tomala and Felix Halim and Alex Castro-Ros and Zefei Li and Anudhyan Boral and Pauline Sho and Michal Yarom and Eric Malmi and David Klinghoffer and Rebecca Lin and Alan Ansell and Pradeep Kumar S and Shubin Zhao and Siqi Zuo and Adam Santoro and Heng-Tze Cheng and Solomon Demmessie and Yuchi Liu and Nicole Brichtova and Allie Culp and Nathaniel Braun and Dan Graur and Will Ng and Nikhil Mehta and Aaron Phillips and Patrik Sundberg and Varun Godbole and Fangyu Liu and Yash Katariya and David Rim and Mojtaba Seyedhosseini and Sean Ammirati and Jonas Valfridsson and Mahan Malihi and Timothy Knight and Andeep Toor and Thomas Lampe and Abe Ittycheriah and Lewis Chiang and Chak Yeung and Alexandre Fréchette and Jinmeng Rao and Huisheng Wang and Himanshu Srivastava and Richard Zhang and Rocky Rhodes and Ariel Brand and Dean Weesner and Ilya Figotin and Felix Gimeno and Rachana Fellinger and Pierre Marcenac and José Leal and Eyal Marcus and Victor Cotruta and Rodrigo Cabrera and Sheryl Luo and Dan Garrette and Vera Axelrod and Sorin Baltateanu and David Barker and Dongkai Chen and Horia Toma and Ben Ingram and Jason Riesa and Chinmay Kulkarni and Yujing Zhang and Hongbin Liu and Chao Wang and Martin Polacek and Will Wu and Kai Hui and Adrian N Reyes and Yi Su and Megan Barnes and Ishaan Malhi and Anfal Siddiqui and Qixuan Feng and Mihai Damaschin and Daniele Pighin and Andreas Steiner and Samuel Yang and Ramya Sree Boppana and Simeon Ivanov and Arun Kandoor and Aditya Shah and Asier Mujika and Da Huang and Christopher A. Choquette-Choo and Mohak Patel and Tianhe Yu and Toni Creswell and Jerry and Liu and Catarina Barros and Yasaman Razeghi and Aurko Roy and Phil Culliton and Binbin Xiong and Jiaqi Pan and Thomas Strohmann and Tolly Powell and Babi Seal and Doug DeCarlo and Pranav Shyam and Kaan Katircioglu and Xuezhi Wang and Cassidy Hardin and Immanuel Odisho and Josef Broder and Oscar Chang and Arun Nair and Artem Shtefan and Maura O'Brien and Manu Agarwal and Sahitya Potluri and Siddharth Goyal and Amit Jhindal and Saksham Thakur and Yury Stuken and James Lyon and Kristina Toutanova and Fangxiaoyu Feng and Austin Wu and Ben Horn and Alek Wang and Alex Cullum and Gabe Taubman and Disha Shrivastava and Chongyang Shi and Hamish Tomlinson and Roma Patel and Tao Tu and Ada Maksutaj Oflazer and Francesco Pongetti and Mingyao Yang and Adrien Ali Taïga and Vincent Perot and Nuo Wang Pierse and Feng Han and Yoel Drori and Iñaki Iturrate and Ayan Chakrabarti and Legg Yeung and Dave Dopson and Yi-ting Chen and Apoorv Kulshreshtha and Tongfei Guo and Philip Pham and Tal Schuster and Junquan Chen and Alex Polozov and Jinwei Xing and Huanjie Zhou and Praneeth Kacham and Doron Kukliansky and Antoine Miech and Sergey Yaroshenko and Ed Chi and Sholto Douglas and Hongliang Fei and Mathieu Blondel and Preethi Myla and Lior Madmoni and Xing Wu and Daniel Keysers and Kristian Kjems and Isabela Albuquerque and Lijun Yu and Joel D'sa and Michelle Plantan and Vlad Ionescu and Jaume Sanchez Elias and Abhirut Gupta and Manish Reddy Vuyyuru and Fred Alcober and Tong Zhou and Kaiyang Ji and Florian Hartmann and Subha Puttagunta and Hugo Song and Ehsan Amid and Anca Stefanoiu and Andrew Lee and Paul Pucciarelli and Emma Wang and Amit Raul and Slav Petrov and Isaac Tian and Valentin Anklin and Nana Nti and Victor Gomes and Max Schumacher and Grace Vesom and Alex Panagopoulos and Konstantinos Bousmalis and Daniel Andor and Josh Jacob and Yuan Zhang and Bill Rosgen and Matija Kecman and Matthew Tung and Alexandra Belias and Noah Goodman and Paul Covington and Brian Wieder and Nikita Saxena and Elnaz Davoodi and Muhuan Huang and Sharath Maddineni and Vincent Roulet and Folawiyo Campbell-Ajala and Pier Giuseppe Sessa and Xintian and Wu and Guangda Lai and Paul Collins and Alex Haig and Vytenis Sakenas and Xiaowei Xu and Marissa Giustina and Laurent El Shafey and Pichi Charoenpanit and Shefali Garg and Joshua Ainslie and Boone Severson and Montse Gonzalez Arenas and Shreya Pathak and Sujee Rajayogam and Jie Feng and Michiel Bakker and Sheng Li and Nevan Wichers and Jamie Rogers and Xinyang Geng and Yeqing Li and Rolf Jagerman and Chao Jia and Nadav Olmert and David Sharon and Matthew Mauger and Sandeep Mariserla and Hongxu Ma and Megha Mohabey and Kyuyeun Kim and Alek Andreev and Scott Pollom and Juliette Love and Vihan Jain and Priyanka Agrawal and Yannick Schroecker and Alisa Fortin and Manfred Warmuth and Ji Liu and Andrew Leach and Irina Blok and Ganesh Poomal Girirajan and Roee Aharoni and Benigno Uria and Andrei Sozanschi and Dan Goldberg and Lucian Ionita and Marco Tulio Ribeiro and Martin Zlocha and Vighnesh Birodkar and Sami Lachgar and Liangzhe Yuan and Himadri Choudhury and Matt Ginsberg and Fei Zheng and Gregory Dibb and Emily Graves and Swachhand Lokhande and Gabriel Rasskin and George-Cristian Muraru and Corbin Quick and Sandeep Tata and Pierre Sermanet and Aditya Chawla and Itay Karo and Yan Wang and Susan Zhang and Orgad Keller and Anca Dragan and Guolong Su and Ian Chou and Xi Liu and Yiqing Tao and Shruthi Prabhakara and Marc Wilson and Ruibo Liu and Shibo Wang and Georgie Evans and David Du and Alfonso Castaño and Gautam Prasad and Mona El Mahdy and Sebastian Gerlach and Machel Reid and Jarrod Kahn and Amir Zait and Thanumalayan Sankaranarayana Pillai and Thatcher Ulrich and Guanyu Wang and Jan Wassenberg and Efrat Farkash and Kiran Yalasangi and Congchao Wang and Maria Bauza and Simon Bucher and Ting Liu and Jun Yan and Gary Leung and Vikas Sindhwani and Parker Barnes and Avi Singh and Ivan Jurin and Jichuan Chang and Niket Kumar Bhumihar and Sivan Eiger and Gui Citovsky and Ben Withbroe and Zhang Li and Siyang Xue and Niccolò Dal Santo and Georgi Stoyanov and Yves Raimond and Steven Zheng and Yilin Gao and Vít Listík and Sławek Kwasiborski and Rachel Saputro and Adnan Ozturel and Ganesh Mallya and Kushal Majmundar and Ross West and Paul Caron and Jinliang Wei and Lluis Castrejon and Sharad Vikram and Deepak Ramachandran and Nikhil Dhawan and Jiho Park and Sara Smoot and George van den Driessche and Yochai Blau and Chase Malik and Wei Liang and Roy Hirsch and Cicero Nogueira dos Santos and Eugene Weinstein and Aäron van den Oord and Sid Lall and Nicholas FitzGerald and Zixuan Jiang and Xuan Yang and Dale Webster and Ali Elqursh and Aedan Pope and Georges Rotival and David Raposo and Wanzheng Zhu and Jeff Dean and Sami Alabed and Dustin Tran and Arushi Gupta and Zach Gleicher and Jessica Austin and Edouard Rosseel and Megh Umekar and Dipanjan Das and Yinghao Sun and Kai Chen and Karolis Misiunas and Xiang Zhou and Yixian Di and Alyssa Loo and Josh Newlan and Bo Li and Vinay Ramasesh and Ying Xu and Alex Chen and Sudeep Gandhe and Radu Soricut and Nikita Gupta and Shuguang Hu and Seliem El-Sayed and Xavier Garcia and Idan Brusilovsky and Pu-Chin Chen and Andrew Bolt and Lu Huang and Alex Gurney and Zhiying Zhang and Alexander Pritzel and Jarek Wilkiewicz and Bryan Seybold and Bhargav Kanagal Shamanna and Felix Fischer and Josef Dean and Karan Gill and Ross Mcilroy and Abhishek Bhowmick and Jeremy Selier and Antoine Yang and Derek Cheng and Vladimir Magay and Jie Tan and Dhriti Varma and Christian Walder and Tomas Kocisky and Ryo Nakashima and Paul Natsev and Mike Kwong and Ionel Gog and Chiyuan Zhang and Sander Dieleman and Thomas Jimma and Andrey Ryabtsev and Siddhartha Brahma and David Steiner and Dayou Du and Ante Žužul and Mislav Žanić and Mukund Raghavachari and Willi Gierke and Zeyu Zheng and Dessie Petrova and Yann Dauphin and Yuchuan Liu and Ido Kessler and Steven Hand and Chris Duvarney and Seokhwan Kim and Hyo Lee and Léonard Hussenot and Jeffrey Hui and Josh Smith and Deepali Jain and Jiawei Xia and Gaurav Singh Tomar and Keyvan Amiri and Du Phan and Fabian Fuchs and Tobias Weyand and Nenad Tomasev and Alexandra Cordell and Xin Liu and Jonathan Mallinson and Pankaj Joshi and Andy Crawford and Arun Suggala and Steve Chien and Nick Fernando and Mariella Sanchez-Vargas and Duncan Williams and Phil Crone and Xiyang Luo and Igor Karpov and Jyn Shan and Terry Thurk and Robin Strudel and Paul Voigtlaender and Piyush Patil and Tim Dozat and Ali Khodaei and Sahil Singla and Piotr Ambroszczyk and Qiyin Wu and Yifan Chang and Brian Roark and Chaitra Hegde and Tianli Ding and Angelos Filos and Zhongru Wu and André Susano Pinto and Shuang Liu and Saarthak Khanna and Aditya Pandey and Siobhan Mcloughlin and Qiujia Li and Sam Haves and Allan Zhou and Elena Buchatskaya and Isabel Leal and Peter de Boursac and Nami Akazawa and Nina Anderson and Terry Chen and Krishna Somandepalli and Chen Liang and Sheela Goenka and Stephanie Winkler and Alexander Grushetsky and Yifan Ding and Jamie Smith and Fan Ye and Jordi Pont-Tuset and Eric Li and Ruichao Li and Tomer Golany and Dawid Wegner and Tao Jiang and Omer Barak and Yuan Shangguan and Eszter Vértes and Renee Wong and Jörg Bornschein and Alex Tudor and Michele Bevilacqua and Tom Schaul and Ankit Singh Rawat and Yang Zhao and Kyriakos Axiotis and Lei Meng and Cory McLean and Jonathan Lai and Jennifer Beattie and Nate Kushman and Yaxin Liu and Blair Kutzman and Fiona Lang and Jingchen Ye and Praneeth Netrapalli and Pushkar Mishra and Myriam Khan and Megha Goel and Rob Willoughby and David Tian and Honglei Zhuang and JD Chen and Zak Tsai and Tasos Kementsietsidis and Arjun Khare and James Keeling and Keyang Xu and Nathan Waters and Florent Altché and Ashok Popat and Bhavishya Mittal and David Saxton and Dalia El Badawy and Michael Mathieu and Zheng Zheng and Hao Zhou and Nishant Ranka and Richard Shin and Qingnan Duan and Tim Salimans and Ioana Mihailescu and Uri Shaham and Ming-Wei Chang and Yannis Assael and Nishanth Dikkala and Martin Izzard and Vincent Cohen-Addad and Cat Graves and Vlad Feinberg and Grace Chung and DJ Strouse and Danny Karmon and Sahand Sharifzadeh and Zoe Ashwood and Khiem Pham and Jon Blanton and Alex Vasiloff and Jarred Barber and Mark Geller and Aurick Zhou and Fedir Zubach and Tzu-Kuo Huang and Lei Zhang and Himanshu Gupta and Matt Young and Julia Proskurnia and Ronny Votel and Valentin Gabeur and Gabriel Barcik and Aditya Tripathi and Hongkun Yu and Geng Yan and Beer Changpinyo and Filip Pavetić and Amy Coyle and Yasuhisa Fujii and Jorge Gonzalez Mendez and Tianhao Zhou and Harish Rajamani and Blake Hechtman and Eddie Cao and Da-Cheng Juan and Yi-Xuan Tan and Valentin Dalibard and Yilun Du and Natalie Clay and Kaisheng Yao and Wenhao Jia and Dimple Vijaykumar and Yuxiang Zhou and Xinyi Bai and Wei-Chih Hung and Steven Pecht and Georgi Todorov and Nikhil Khadke and Pramod Gupta and Preethi Lahoti and Arnaud Autef and Karthik Duddu and James Lee-Thorp and Alexander Bykovsky and Tautvydas Misiunas and Sebastian Flennerhag and Santhosh Thangaraj and Jed McGiffin and Zack Nado and Markus Kunesch and Andreas Noever and Amir Hertz and Marco Liang and Victor Stone and Evan Palmer and Samira Daruki and Arijit Pramanik and Siim Põder and Austin Kyker and Mina Khan and Evgeny Sluzhaev and Marvin Ritter and Avraham Ruderman and Wenlei Zhou and Chirag Nagpal and Kiran Vodrahalli and George Necula and Paul Barham and Ellie Pavlick and Jay Hartford and Izhak Shafran and Long Zhao and Maciej Mikuła and Tom Eccles and Hidetoshi Shimokawa and Kanav Garg and Luke Vilnis and Hanwen Chen and Ilia Shumailov and Kuang-Huei Lee and Abdelrahman Abdelhamed and Meiyan Xie and Vered Cohen and Ester Hlavnova and Dan Malkin and Chawin Sitawarin and James Lottes and Pauline Coquinot and Tianli Yu and Sandeep Kumar and Jingwei Zhang and Aroma Mahendru and Zafarali Ahmed and James Martens and Tao Chen and Aviel Boag and Daiyi Peng and Coline Devin and Arseniy Klimovskiy and Mary Phuong and Danny Vainstein and Jin Xie and Bhuvana Ramabhadran and Nathan Howard and Xinxin Yu and Gitartha Goswami and Jingyu Cui and Sam Shleifer and Mario Pinto and Chih-Kuan Yeh and Ming-Hsuan Yang and Sara Javanmardi and Dan Ethier and Chace Lee and Jordi Orbay and Suyog Kotecha and Carla Bromberg and Pete Shaw and James Thornton and Adi Gerzi Rosenthal and Shane Gu and Matt Thomas and Ian Gemp and Aditya Ayyar and Asahi Ushio and Aarush Selvan and Joel Wee and Chenxi Liu and Maryam Majzoubi and Weiren Yu and Jake Abernethy and Tyler Liechty and Renke Pan and Hoang Nguyen and Qiong and Hu and Sarah Perrin and Abhinav Arora and Emily Pitler and Weiyi Wang and Kaushik Shivakumar and Flavien Prost and Ben Limonchik and Jing Wang and Yi Gao and Timothee Cour and Shyamal Buch and Huan Gui and Maria Ivanova and Philipp Neubeck and Kelvin Chan and Lucy Kim and Huizhong Chen and Naman Goyal and Da-Woon Chung and Lu Liu and Yao Su and Anastasia Petrushkina and Jiajun Shen and Armand Joulin and Yuanzhong Xu and Stein Xudong Lin and Yana Kulizhskaya and Ciprian Chelba and Shobha Vasudevan and Eli Collins and Vasilisa Bashlovkina and Tony Lu and Doug Fritz and Jongbin Park and Yanqi Zhou and Chen Su and Richard Tanburn and Mikhail Sushkov and Mitchelle Rasquinha and Jinning Li and Jennifer Prendki and Yiming Li and Pallavi LV and Shriya Sharma and Hen Fitoussi and Hui Huang and Andrew Dai and Phuong Dao and Mike Burrows and Henry Prior and Danfeng Qin and Golan Pundak and Lars Lowe Sjoesund and Art Khurshudov and Zhenkai Zhu and Albert Webson and Elizabeth Kemp and Tat Tan and Saurabh Agrawal and Susie Sargsyan and Liqun Cheng and Jim Stephan and Tom Kwiatkowski and David Reid and Arunkumar Byravan and Assaf Hurwitz Michaely and Nicolas Heess and Luowei Zhou and Sonam Goenka and Viral Carpenter and Anselm Levskaya and Bo Wang and Reed Roberts and Rémi Leblond and Sharat Chikkerur and Stav Ginzburg and Max Chang and Robert Riachi and Chuqiao and Xu and Zalán Borsos and Michael Pliskin and Julia Pawar and Morgane Lustman and Hannah Kirkwood and Ankit Anand and Aditi Chaudhary and Norbert Kalb and Kieran Milan and Sean Augenstein and Anna Goldie and Laurel Prince and Karthik Raman and Yanhua Sun and Vivian Xia and Aaron Cohen and Zhouyuan Huo and Josh Camp and Seher Ellis and Lukas Zilka and David Vilar Torres and Lisa Patel and Sho Arora and Betty Chan and Jonas Adler and Kareem Ayoub and Jacky Liang and Fayaz Jamil and Jiepu Jiang and Simon Baumgartner and Haitian Sun and Yael Karov and Yaroslav Akulov and Hui Zheng and Irene Cai and Claudio Fantacci and James Rubin and Alex Rav Acha and Mengchao Wang and Nina D'Souza and Rohit Sathyanarayana and Shengyang Dai and Simon Rowe and Andrey Simanovsky and Omer Goldman and Yuheng Kuang and Xiaoyue Pan and Andrew Rosenberg and Tania Rojas-Esponda and Praneet Dutta and Amy Zeng and Irina Jurenka and Greg Farquhar and Yamini Bansal and Shariq Iqbal and Becca Roelofs and Ga-Young Joung and Parker Beak and Changwan Ryu and Ryan Poplin and Yan Wu and Jean-Baptiste Alayrac and Senaka Buthpitiya and Olaf Ronneberger and Caleb Habtegebriel and Wei Li and Paul Cavallaro and Aurora Wei and Guy Bensky and Timo Denk and Harish Ganapathy and Jeff Stanway and Pratik Joshi and Francesco Bertolini and Jessica Lo and Olivia Ma and Zachary Charles and Geta Sampemane and Himanshu Sahni and Xu Chen and Harry Askham and David Gaddy and Peter Young and Jiewen Tan and Matan Eyal and Arthur Bražinskas and Li Zhong and Zhichun Wu and Mark Epstein and Kai Bailey and Andrew Hard and Kamyu Lee and Sasha Goldshtein and Alex Ruiz and Mohammed Badawi and Matthias Lochbrunner and JK Kearns and Ashley Brown and Fabio Pardo and Theophane Weber and Haichuan Yang and Pan-Pan Jiang and Berkin Akin and Zhao Fu and Marcus Wainwright and Chi Zou and Meenu Gaba and Pierre-Antoine Manzagol and Wendy Kan and Yang Song and Karina Zainullina and Rui Lin and Jeongwoo Ko and Salil Deshmukh and Apoorv Jindal and James Svensson and Divya Tyam and Heri Zhao and Christine Kaeser-Chen and Scott Baird and Pooya Moradi and Jamie Hall and Qiuchen Guo and Vincent Tsang and Bowen Liang and Fernando Pereira and Suhas Ganesh and Ivan Korotkov and Jakub Adamek and Sridhar Thiagarajan and Vinh Tran and Charles Chen and Chris Tar and Sanil Jain and Ishita Dasgupta and Taylan Bilal and David Reitter and Kai Zhao and Giulia Vezzani and Yasmin Gehman and Pulkit Mehta and Lauren Beltrone and Xerxes Dotiwalla and Sergio Guadarrama and Zaheer Abbas and Stefani Karp and Petko Georgiev and Chun-Sung Ferng and Marc Brockschmidt and Liqian Peng and Christoph Hirnschall and Vikas Verma and Yingying Bi and Ying Xiao and Avigail Dabush and Kelvin Xu and Phil Wallis and Randall Parker and Qifei Wang and Yang Xu and Ilkin Safarli and Dinesh Tewari and Yin Zhang and Seungyeon Kim and Andrea Gesmundo and Mackenzie Thomas and Sergey Levi and Ahmed Chowdhury and Kanishka Rao and Peter Garst and Sam Conway-Rahman and Helen Ran and Kay McKinney and Zhisheng Xiao and Wenhao Yu and Rohan Agrawal and Axel Stjerngren and Catalin Ionescu and Jingjing Chen and Vivek Sharma and Justin Chiu and Fei Liu and Ken Franko and Clayton Sanford and Xingyu Cai and Paul Michel and Sanjay Ganapathy and Jane Labanowski and Zachary Garrett and Ben Vargas and Sean Sun and Bryan Gale and Thomas Buschmann and Guillaume Desjardins and Nimesh Ghelani and Palak Jain and Mudit Verma and Chulayuth Asawaroengchai and Julian Eisenschlos and Jitendra Harlalka and Hideto Kazawa and Don Metzler and Joshua Howland and Ying Jian and Jake Ades and Viral Shah and Tynan Gangwani and Seungji Lee and Roman Ring and Steven M. Hernandez and Dean Reich and Amer Sinha and Ashutosh Sathe and Joe Kovac and Ashleah Gill and Ajay Kannan and Andrea D'olimpio and Martin Sevenich and Jay Whang and Been Kim and Khe Chai Sim and Jilin Chen and Jiageng Zhang and Shuba Lall and Yossi Matias and Bill Jia and Abe Friesen and Sara Nasso and Ashish Thapliyal and Bryan Perozzi and Ting Yu and Anna Shekhawat and Safeen Huda and Peter Grabowski and Eric Wang and Ashwin Sreevatsa and Hilal Dib and Mehadi Hassen and Parker Schuh and Vedrana Milutinovic and Chris Welty and Michael Quinn and Ali Shah and Bangju Wang and Gabe Barth-Maron and Justin Frye and Natalie Axelsson and Tao Zhu and Yukun Ma and Irene Giannoumis and Hanie Sedghi and Chang Ye and Yi Luan and Kevin Aydin and Bilva Chandra and Vivek Sampathkumar and Ronny Huang and Victor Lavrenko and Ahmed Eleryan and Zhi Hong and Steven Hansen and Sara Mc Carthy and Bidisha Samanta and Domagoj Ćevid and Xin Wang and Fangtao Li and Michael Voznesensky and Matt Hoffman and Andreas Terzis and Vikash Sehwag and Gil Fidel and Luheng He and Mu Cai and Yanzhang He and Alex Feng and Martin Nikoltchev and Samrat Phatale and Jason Chase and Rory Lawton and Ming Zhang and Tom Ouyang and Manuel Tragut and Mehdi Hafezi Manshadi and Arjun Narayanan and Jiaming Shen and Xu Gao and Tolga Bolukbasi and Nick Roy and Xin Li and Daniel Golovin and Liviu Panait and Zhen Qin and Guangxing Han and Thomas Anthony and Sneha Kudugunta and Viorica Patraucean and Aniket Ray and Xinyun Chen and Xiaochen Yang and Tanuj Bhatia and Pranav Talluri and Alex Morris and Andrija Ražnatović and Bethanie Brownfield and James An and Sheng Peng and Patrick Kane and Ce Zheng and Nico Duduta and Joshua Kessinger and James Noraky and Siqi Liu and Keran Rong and Petar Veličković and Keith Rush and Alex Goldin and Fanny Wei and Shiva Mohan Reddy Garlapati and Caroline Pantofaru and Okwan Kwon and Jianmo Ni and Eric Noland and Julia Di Trapani and Françoise Beaufays and Abhijit Guha Roy and Yinlam Chow and Aybuke Turker and Geoffrey Cideron and Lantao Mei and Jon Clark and Qingyun Dou and Matko Bošnjak and Ralph Leith and Yuqing Du and Amir Yazdanbakhsh and Milad Nasr and Chester Kwak and Suraj Satishkumar Sheth and Alex Kaskasoli and Ankesh Anand and Balaji Lakshminarayanan and Sammy Jerome and David Bieber and Chun-Te Chu and Alexandre Senges and Tianxiao Shen and Mukund Sridhar and Ndaba Ndebele and Benjamin Beyret and Shakir Mohamed and Mia Chen and Markus Freitag and Jiaxian Guo and Luyang Liu and Paul Roit and Heng Chen and Shen Yan and Tom Stone and JD Co-Reyes and Jeremy Cole and Salvatore Scellato and Shekoofeh Azizi and Hadi Hashemi and Alicia Jin and Anand Iyer and Marcella Valentine and András György and Arun Ahuja and Daniel Hernandez Diaz and Chen-Yu Lee and Nathan Clement and Weize Kong and Drew Garmon and Ishaan Watts and Kush Bhatia and Khyatti Gupta and Matt Miecnikowski and Hugo Vallet and Ankur Taly and Edward Loper and Saket Joshi and James Atwood and Jo Chick and Mark Collier and Fotis Iliopoulos and Ryan Trostle and Beliz Gunel and Ramiro Leal-Cavazos and Arnar Mar Hrafnkelsson and Michael Guzman and Xiaoen Ju and Andy Forbes and Jesse Emond and Kushal Chauhan and Ben Caine and Li Xiao and Wenjun Zeng and Alexandre Moufarek and Daniel Murphy and Maya Meng and Nitish Gupta and Felix Riedel and Anil Das and Elijah Lawal and Shashi Narayan and Tiberiu Sosea and James Swirhun and Linda Friso and Behnam Neyshabur and Jing Lu and Sertan Girgin and Michael Wunder and Edouard Yvinec and Aroonalok Pyne and Victor Carbune and Shruti Rijhwani and Yang Guo and Tulsee Doshi and Anton Briukhov and Max Bain and Ayal Hitron and Xuanhui Wang and Ashish Gupta and Ke Chen and Cosmo Du and Weiyang Zhang and Dhruv Shah and Arjun Akula and Max Dylla and Ashyana Kachra and Weicheng Kuo and Tingting Zou and Lily Wang and Luyao Xu and Jifan Zhu and Justin Snyder and Sachit Menon and Orhan Firat and Igor Mordatch and Yuan Yuan and Natalia Ponomareva and Rory Blevins and Lawrence Moore and Weijun Wang and Phil Chen and Martin Scholz and Artur Dwornik and Jason Lin and Sicheng Li and Diego Antognini and Te I and Xiaodan Song and Matt Miller and Uday Kalra and Adam Raveret and Oscar Akerlund and Felix Wu and Andrew Nystrom and Namrata Godbole and Tianqi Liu and Hannah DeBalsi and Jewel Zhao and Buhuang Liu and Avi Caciularu and Lauren Lax and Urvashi Khandelwal and Victoria Langston and Eric Bailey and Silvio Lattanzi and Yufei Wang and Neel Kovelamudi and Sneha Mondal and Guru Guruganesh and Nan Hua and Ofir Roval and Paweł Wesołowski and Rishikesh Ingale and Jonathan Halcrow and Tim Sohn and Christof Angermueller and Bahram Raad and Eli Stickgold and Eva Lu and Alec Kosik and Jing Xie and Timothy Lillicrap and Austin Huang and Lydia Lihui Zhang and Dominik Paulus and Clement Farabet and Alex Wertheim and Bing Wang and Rishabh Joshi and Chu-ling Ko and Yonghui Wu and Shubham Agrawal and Lily Lin and XiangHai Sheng and Peter Sung and Tyler Breland-King and Christina Butterfield and Swapnil Gawde and Sumeet Singh and Qiao Zhang and Raj Apte and Shilpa Shetty and Adrian Hutter and Tao Li and Elizabeth Salesky and Federico Lebron and Jonni Kanerva and Michela Paganini and Arthur Nguyen and Rohith Vallu and Jan-Thorsten Peter and Sarmishta Velury and David Kao and Jay Hoover and Anna Bortsova and Colton Bishop and Shoshana Jakobovits and Alessandro Agostini and Alekh Agarwal and Chang Liu and Charles Kwong and Sasan Tavakkol and Ioana Bica and Alex Greve and Anirudh GP and Jake Marcus and Le Hou and Tom Duerig and Rivka Moroshko and Dave Lacey and Andy Davis and Julien Amelot and Guohui Wang and Frank Kim and Theofilos Strinopoulos and Hui Wan and Charline Le Lan and Shankar Krishnan and Haotian Tang and Peter Humphreys and Junwen Bai and Idan Heimlich Shtacher and Diego Machado and Chenxi Pang and Ken Burke and Dangyi Liu and Renga Aravamudhan and Yue Song and Ed Hirst and Abhimanyu Singh and Brendan Jou and Liang Bai and Francesco Piccinno and Chuyuan Kelly Fu and Robin Alazard and Barak Meiri and Daniel Winter and Charlie Chen and Mingda Zhang and Jens Heitkaemper and John Lambert and Jinhyuk Lee and Alexander Frömmgen and Sergey Rogulenko and Pranav Nair and Paul Niemczyk and Anton Bulyenov and Bibo Xu and Hadar Shemtov and Morteza Zadimoghaddam and Serge Toropov and Mateo Wirth and Hanjun Dai and Sreenivas Gollapudi and Daniel Zheng and Alex Kurakin and Chansoo Lee and Kalesha Bullard and Nicolas Serrano and Ivana Balazevic and Yang Li and Johan Schalkwyk and Mark Murphy and Mingyang Zhang and Kevin Sequeira and Romina Datta and Nishant Agrawal and Charles Sutton and Nithya Attaluri and Mencher Chiang and Wael Farhan and Gregory Thornton and Kate Lin and Travis Choma and Hung Nguyen and Kingshuk Dasgupta and Dirk Robinson and Iulia Comşa and Michael Riley and Arjun Pillai and Basil Mustafa and Ben Golan and Amir Zandieh and Jean-Baptiste Lespiau and Billy Porter and David Ross and Sujeevan Rajayogam and Mohit Agarwal and Subhashini Venugopalan and Bobak Shahriari and Qiqi Yan and Hao Xu and Taylor Tobin and Pavel Dubov and Hongzhi Shi and Adrià Recasens and Anton Kovsharov and Sebastian Borgeaud and Lucio Dery and Shanthal Vasanth and Elena Gribovskaya and Linhai Qiu and Mahdis Mahdieh and Wojtek Skut and Elizabeth Nielsen and CJ Zheng and Adams Yu and Carrie Grimes Bostock and Shaleen Gupta and Aaron Archer and Chris Rawles and Elinor Davies and Alexey Svyatkovskiy and Tomy Tsai and Yoni Halpern and Christian Reisswig and Bartek Wydrowski and Bo Chang and Joan Puigcerver and Mor Hazan Taege and Jian Li and Eva Schnider and Xinjian Li and Dragos Dena and Yunhan Xu and Umesh Telang and Tianze Shi and Heiga Zen and Kyle Kastner and Yeongil Ko and Neesha Subramaniam and Aviral Kumar and Pete Blois and Zhuyun Dai and John Wieting and Yifeng Lu and Yoel Zeldes and Tian Xie and Anja Hauth and Alexandru Ţifrea and Yuqi Li and Sam El-Husseini and Dan Abolafia and Howard Zhou and Wen Ding and Sahra Ghalebikesabi and Carlos Guía and Andrii Maksai and Ágoston Weisz and Sercan Arik and Nick Sukhanov and Aga Świetlik and Xuhui Jia and Luo Yu and Weiyue Wang and Mark Brand and Dawn Bloxwich and Sean Kirmani and Zhe Chen and Alec Go and Pablo Sprechmann and Nithish Kannen and Alen Carin and Paramjit Sandhu and Isabel Edkins and Leslie Nooteboom and Jai Gupta and Loren Maggiore and Javad Azizi and Yael Pritch and Pengcheng Yin and Mansi Gupta and Danny Tarlow and Duncan Smith and Desi Ivanov and Mohammad Babaeizadeh and Ankita Goel and Satish Kambala and Grace Chu and Matej Kastelic and Michelle Liu and Hagen Soltau and Austin Stone and Shivani Agrawal and Min Kim and Kedar Soparkar and Srinivas Tadepalli and Oskar Bunyan and Rachel Soh and Arvind Kannan and DY Kim and Blake JianHang Chen and Afief Halumi and Sudeshna Roy and Yulong Wang and Olcan Sercinoglu and Gena Gibson and Sijal Bhatnagar and Motoki Sano and Daniel von Dincklage and Qingchun Ren and Blagoj Mitrevski and Mirek Olšák and Jennifer She and Carl Doersch and Jilei and Wang and Bingyuan Liu and Qijun Tan and Tamar Yakar and Tris Warkentin and Alex Ramirez and Carl Lebsack and Josh Dillon and Rajiv Mathews and Tom Cobley and Zelin Wu and Zhuoyuan Chen and Jon Simon and Swaroop Nath and Tara Sainath and Alexei Bendebury and Ryan Julian and Bharath Mankalale and Daria Ćurko and Paulo Zacchello and Adam R. Brown and Kiranbir Sodhia and Heidi Howard and Sergi Caelles and Abhinav Gupta and Gareth Evans and Anna Bulanova and Lesley Katzen and Roman Goldenberg and Anton Tsitsulin and Joe Stanton and Benoit Schillings and Vitaly Kovalev and Corey Fry and Rushin Shah and Kuo Lin and Shyam Upadhyay and Cheng Li and Soroush Radpour and Marcello Maggioni and Jing Xiong and Lukas Haas and Jenny Brennan and Aishwarya Kamath and Nikolay Savinov and Arsha Nagrani and Trevor Yacovone and Ryan Kappedal and Kostas Andriopoulos and Li Lao and YaGuang Li and Grigory Rozhdestvenskiy and Kazuma Hashimoto and Andrew Audibert and Sophia Austin and Daniel Rodriguez and Anian Ruoss and Garrett Honke and Deep Karkhanis and Xi Xiong and Qing Wei and James Huang and Zhaoqi Leng and Vittal Premachandran and Stan Bileschi and Georgios Evangelopoulos and Thomas Mensink and Jay Pavagadhi and Denis Teplyashin and Paul Chang and Linting Xue and Garrett Tanzer and Sally Goldman and Kaushal Patel and Shixin Li and Jeremy Wiesner and Ivy Zheng and Ian Stewart-Binks and Jie Han and Zhi Li and Liangchen Luo and Karel Lenc and Mario Lučić and Fuzhao Xue and Ryan Mullins and Alexey Guseynov and Chung-Ching Chang and Isaac Galatzer-Levy and Adam Zhang and Garrett Bingham and Grace Hu and Ale Hartman and Yue Ma and Jordan Griffith and Alex Irpan and Carey Radebaugh and Summer Yue and Lijie Fan and Victor Ungureanu and Christina Sorokin and Hannah Teufel and Peiran Li and Rohan Anil and Dimitris Paparas and Todd Wang and Chu-Cheng Lin and Hui Peng and Megan Shum and Goran Petrovic and Demetra Brady and Richard Nguyen and Klaus Macherey and Zhihao Li and Harman Singh and Madhavi Yenugula and Mariko Iinuma and Xinyi Chen and Kavya Kopparapu and Alexey Stern and Shachi Dave and Chandu Thekkath and Florence Perot and Anurag Kumar and Fangda Li and Yang Xiao and Matthew Bilotti and Mohammad Hossein Bateni and Isaac Noble and Lisa Lee and Amelio Vázquez-Reina and Julian Salazar and Xiaomeng Yang and Boyu Wang and Ela Gruzewska and Anand Rao and Sindhu Raghuram and Zheng Xu and Eyal Ben-David and Jieru Mei and Sid Dalmia and Zhaoyi Zhang and Yuchen Liu and Gagan Bansal and Helena Pankov and Steven Schwarcz and Andrea Burns and Christine Chan and Sumit Sanghai and Ricky Liang and Ethan Liang and Antoine He and Amy Stuart and Arun Narayanan and Yukun Zhu and Christian Frank and Bahar Fatemi and Amit Sabne and Oran Lang and Indro Bhattacharya and Shane Settle and Maria Wang and Brendan McMahan and Andrea Tacchetti and Livio Baldini Soares and Majid Hadian and Serkan Cabi and Timothy Chung and Nikita Putikhin and Gang Li and Jeremy Chen and Austin Tarango and Henryk Michalewski and Mehran Kazemi and Hussain Masoom and Hila Sheftel and Rakesh Shivanna and Archita Vadali and Ramona Comanescu and Doug Reid and Joss Moore and Arvind Neelakantan and Michaël Sander and Jonathan Herzig and Aviv Rosenberg and Mostafa Dehghani and JD Choi and Michael Fink and Reid Hayes and Eric Ge and Shitao Weng and Chia-Hua Ho and John Karro and Kalpesh Krishna and Lam Nguyen Thiet and Amy Skerry-Ryan and Daniel Eppens and Marco Andreetto and Navin Sarma and Silvano Bonacina and Burcu Karagol Ayan and Megha Nawhal and Zhihao Shan and Mike Dusenberry and Shantanu Thakoor and Sagar Gubbi and Duc Dung Nguyen and Reut Tsarfaty and Samuel Albanie and Jovana Mitrović and Meet Gandhi and Bo-Juen Chen and Alessandro Epasto and Georgi Stephanov and Ye Jin and Samuel Gehman and Aida Amini and Jack Weber and Feryal Behbahani and Shawn Xu and Miltos Allamanis and Xi Chen and Myle Ott and Claire Sha and Michal Jastrzebski and Hang Qi and David Greene and Xinyi Wu and Abodunrinwa Toki and Daniel Vlasic and Jane Shapiro and Ragha Kotikalapudi and Zhe Shen and Takaaki Saeki and Sirui Xie and Albin Cassirer and Shikhar Bharadwaj and Tatsuya Kiyono and Srinadh Bhojanapalli and Elan Rosenfeld and Sam Ritter and Jieming Mao and João Gabriel Oliveira and Zoltan Egyed and Bernd Bandemer and Emilio Parisotto and Keisuke Kinoshita and Juliette Pluto and Petros Maniatis and Steve Li and Yaohui Guo and Golnaz Ghiasi and Jean Tarbouriech and Srimon Chatterjee and Julie Jin and Katrina and Xu and Jennimaria Palomaki and Séb Arnold and Madhavi Sewak and Federico Piccinini and Mohit Sharma and Ben Albrecht and Sean Purser-haskell and Ashwin Vaswani and Chongyan Chen and Matheus Wisniewski and Qin Cao and John Aslanides and Nguyet Minh Phu and Maximilian Sieb and Lauren Agubuzu and Anne Zheng and Daniel Sohn and Marco Selvi and Anders Andreassen and Krishan Subudhi and Prem Eruvbetine and Oliver Woodman and Tomas Mery and Sebastian Krause and Xiaoqi Ren and Xiao Ma and Jincheng Luo and Dawn Chen and Wei Fan and Henry Griffiths and Christian Schuler and Alice Li and Shujian Zhang and Jean-Michel Sarr and Shixin Luo and Riccardo Patana and Matthew Watson and Dani Naboulsi and Michael Collins and Sailesh Sidhwani and Emiel Hoogeboom and Sharon Silver and Emily Caveness and Xiaokai Zhao and Mikel Rodriguez and Maxine Deines and Libin Bai and Patrick Griffin and Marco Tagliasacchi and Emily Xue and Spandana Raj Babbula and Bo Pang and Nan Ding and Gloria Shen and Elijah Peake and Remi Crocker and Shubha Srinivas Raghvendra and Danny Swisher and Woohyun Han and Richa Singh and Ling Wu and Vladimir Pchelin and Tsendsuren Munkhdalai and Dana Alon and Geoff Bacon and Efren Robles and Jannis Bulian and Melvin Johnson and George Powell and Felipe Tiengo Ferreira and Yaoyiran Li and Frederik Benzing and Mihajlo Velimirović and Hubert Soyer and William Kong and Tony and Nguyên and Zhen Yang and Jeremiah Liu and Joost van Amersfoort and Daniel Gillick and Baochen Sun and Nathalie Rauschmayr and Katie Zhang and Serena Zhan and Tao Zhou and Alexey Frolov and Chengrun Yang and Denis Vnukov and Louis Rouillard and Hongji Li and Amol Mandhane and Nova Fallen and Rajesh Venkataraman and Clara Huiyi Hu and Jennifer Brennan and Jenny Lee and Jerry Chang and Martin Sundermeyer and Zhufeng Pan and Rosemary Ke and Simon Tong and Alex Fabrikant and William Bono and Jindong Gu and Ryan Foley and Yiran Mao and Manolis Delakis and Dhruva Bhaswar and Roy Frostig and Nick Li and Avital Zipori and Cath Hope and Olga Kozlova and Swaroop Mishra and Josip Djolonga and Craig Schiff and Majd Al Merey and Eleftheria Briakou and Peter Morgan and Andy Wan and Avinatan Hassidim and RJ Skerry-Ryan and Kuntal Sengupta and Mary Jasarevic and Praveen Kallakuri and Paige Kunkle and Hannah Brennan and Tom Lieber and Hassan Mansoor and Julian Walker and Bing Zhang and Annie Xie and Goran Žužić and Adaeze Chukwuka and Alex Druinsky and Donghyun Cho and Rui Yao and Ferjad Naeem and Shiraz Butt and Eunyoung Kim and Zhipeng Jia and Mandy Jordan and Adam Lelkes and Mark Kurzeja and Sophie Wang and James Zhao and Andrew Over and Abhishek Chakladar and Marcel Prasetya and Neha Jha and Sriram Ganapathy and Yale Cong and Prakash Shroff and Carl Saroufim and Sobhan Miryoosefi and Mohamed Hammad and Tajwar Nasir and Weijuan Xi and Yang Gao and Young Maeng and Ben Hora and Chin-Yi Cheng and Parisa Haghani and Yoad Lewenberg and Caden Lu and Martin Matysiak and Naina Raisinghani and Huiyu Wang and Lexi Baugher and Rahul Sukthankar and Minh Giang and John Schultz and Noah Fiedel and Minmin Chen and Cheng-Chun Lee and Tapomay Dey and Hao Zheng and Shachi Paul and Celine Smith and Andy Ly and Yicheng Wang and Rishabh Bansal and Bartek Perz and Susanna Ricco and Stasha Blank and Vaishakh Keshava and Deepak Sharma and Marvin Chow and Kunal Lad and Komal Jalan and Simon Osindero and Craig Swanson and Jacob Scott and Anastasija Ilić and Xiaowei Li and Siddhartha Reddy Jonnalagadda and Afzal Shama Soudagar and Yan Xiong and Bat-Orgil Batsaikhan and Daniel Jarrett and Naveen Kumar and Maulik Shah and Matt Lawlor and Austin Waters and Mark Graham and Rhys May and Sabela Ramos and Sandra Lefdal and Zeynep Cankara and Nacho Cano and Brendan O'Donoghue and Jed Borovik and Frederick Liu and Jordan Grimstad and Mahmoud Alnahlawi and Katerina Tsihlas and Tom Hudson and Nikolai Grigorev and Yiling Jia and Terry Huang and Tobenna Peter Igwe and Sergei Lebedev and Xiaodan Tang and Igor Krivokon and Frankie Garcia and Melissa Tan and Eric Jia and Peter Stys and Shikhar Vashishth and Yu Liang and Balaji Venkatraman and Chenjie Gu and Anastasios Kementsietsidis and Chen Zhu and Junehyuk Jung and Yunfei Bai and Mohammad Javad Hosseini and Faruk Ahmed and Aditya Gupta and Xin Yuan and Shereen Ashraf and Shitij Nigam and Gautam Vasudevan and Pranjal Awasthi and Adi Mayrav Gilady and Zelda Mariet and Ramy Eskander and Haiguang Li and Hexiang Hu and Guillermo Garrido and Philippe Schlattner and George Zhang and Rohun Saxena and Petar Dević and Kritika Muralidharan and Ashwin Murthy and Yiqian Zhou and Min Choi and Arissa Wongpanich and Zhengdong Wang and Premal Shah and Yuntao Xu and Yiling Huang and Stephen Spencer and Alice Chen and James Cohan and Junjie Wang and Jonathan Tompson and Junru Wu and Ruba Haroun and Haiqiong Li and Blanca Huergo and Fan Yang and Tongxin Yin and James Wendt and Michael Bendersky and Rahma Chaabouni and Javier Snaider and Johan Ferret and Abhishek Jindal and Tara Thompson and Andrew Xue and Will Bishop and Shubham Milind Phal and Archit Sharma and Yunhsuan Sung and Prabakar Radhakrishnan and Mo Shomrat and Reeve Ingle and Roopali Vij and Justin Gilmer and Mihai Dorin Istin and Sam Sobell and Yang Lu and Emily Nottage and Dorsa Sadigh and Jeremiah Willcock and Tingnan Zhang and Steve Xu and Sasha Brown and Katherine Lee and Gary Wang and Yun Zhu and Yi Tay and Cheolmin Kim and Audrey Gutierrez and Abhanshu Sharma and Yongqin Xian and Sungyong Seo and Claire Cui and Elena Pochernina and Cip Baetu and Krzysztof Jastrzębski and Mimi Ly and Mohamed Elhawaty and Dan Suh and Eren Sezener and Pidong Wang and Nancy Yuen and George Tucker and Jiahao Cai and Zuguang Yang and Cindy Wang and Alex Muzio and Hai Qian and Jae Yoo and Derek Lockhart and Kevin R. McKee and Mandy Guo and Malika Mehrotra and Artur Mendonça and Sanket Vaibhav Mehta and Sherry Ben and Chetan Tekur and Jiaqi Mu and Muye Zhu and Victoria Krakovna and Hongrae Lee and AJ Maschinot and Sébastien Cevey and HyunJeong Choe and Aijun Bai and Hansa Srinivasan and Derek Gasaway and Nick Young and Patrick Siegler and Dan Holtmann-Rice and Vihari Piratla and Kate Baumli and Roey Yogev and Alex Hofer and Hado van Hasselt and Svetlana Grant and Yuri Chervonyi and David Silver and Andrew Hogue and Ayushi Agarwal and Kathie Wang and Preeti Singh and Four Flynn and Josh Lipschultz and Robert David and Lizzetth Bellot and Yao-Yuan Yang and Long Le and Filippo Graziano and Kate Olszewska and Kevin Hui and Akanksha Maurya and Nikos Parotsidis and Weijie Chen and Tayo Oguntebi and Joe Kelley and Anirudh Baddepudi and Johannes Mauerer and Gregory Shaw and Alex Siegman and Lin Yang and Shravya Shetty and Subhrajit Roy and Yunting Song and Wojciech Stokowiec and Ryan Burnell and Omkar Savant and Robert Busa-Fekete and Jin Miao and Samrat Ghosh and Liam MacDermed and Phillip Lippe and Mikhail Dektiarev and Zach Behrman and Fabian Mentzer and Kelvin Nguyen and Meng Wei and Siddharth Verma and Chris Knutsen and Sudeep Dasari and Zhipeng Yan and Petr Mitrichev and Xingyu Wang and Virat Shejwalkar and Jacob Austin and Srinivas Sunkara and Navneet Potti and Yan Virin and Christian Wright and Gaël Liu and Oriana Riva and Etienne Pot and Greg Kochanski and Quoc Le and Gargi Balasubramaniam and Arka Dhar and Yuguo Liao and Adam Bloniarz and Divyansh Shukla and Elizabeth Cole and Jong Lee and Sheng Zhang and Sushant Kafle and Siddharth Vashishtha and Parsa Mahmoudieh and Grace Chen and Raphael Hoffmann and Pranesh Srinivasan and Agustin Dal Lago and Yoav Ben Shalom and Zi Wang and Michael Elabd and Anuj Sharma and Junhyuk Oh and Suraj Kothawade and Maigo Le and Marianne Monteiro and Shentao Yang and Kaiz Alarakyia and Robert Geirhos and Diana Mincu and Håvard Garnes and Hayato Kobayashi and Soroosh Mariooryad and Kacper Krasowiak and Zhixin and Lai and Shibl Mourad and Mingqiu Wang and Fan Bu and Ophir Aharoni and Guanjie Chen and Abhimanyu Goyal and Vadim Zubov and Ankur Bapna and Elahe Dabir and Nisarg Kothari and Kay Lamerigts and Nicola De Cao and Jeremy Shar and Christopher Yew and Nitish Kulkarni and Dre Mahaarachchi and Mandar Joshi and Zhenhai Zhu and Jared Lichtarge and Yichao Zhou and Hannah Muckenhirn and Vittorio Selo and Oriol Vinyals and Peter Chen and Anthony Brohan and Vaibhav Mehta and Sarah Cogan and Ruth Wang and Ty Geri and Wei-Jen Ko and Wei Chen and Fabio Viola and Keshav Shivam and Lisa Wang and Madeleine Clare Elish and Raluca Ada Popa and Sébastien Pereira and Jianqiao Liu and Raphael Koster and Donnie Kim and Gufeng Zhang and Sayna Ebrahimi and Partha Talukdar and Yanyan Zheng and Petra Poklukar and Ales Mikhalap and Dale Johnson and Anitha Vijayakumar and Mark Omernick and Matt Dibb and Ayush Dubey and Qiong Hu and Apurv Suman and Vaibhav Aggarwal and Ilya Kornakov and Fei Xia and Wing Lowe and Alexey Kolganov and Ted Xiao and Vitaly Nikolaev and Steven Hemingray and Bonnie Li and Joana Iljazi and Mikołaj Rybiński and Ballie Sandhu and Peggy Lu and Thang Luong and Rodolphe Jenatton and Vineetha Govindaraj and Hui and Li and Gabriel Dulac-Arnold and Wonpyo Park and Henry Wang and Abhinit Modi and Jean Pouget-Abadie and Kristina Greller and Rahul Gupta and Robert Berry and Prajit Ramachandran and Jinyu Xie and Liam McCafferty and Jianling Wang and Kilol Gupta and Hyeontaek Lim and Blaž Bratanič and Andy Brock and Ilia Akolzin and Jim Sproch and Dan Karliner and Duhyeon Kim and Adrian Goedeckemeyer and Noam Shazeer and Cordelia Schmid and Daniele Calandriello and Parul Bhatia and Krzysztof Choromanski and Ceslee Montgomery and Dheeru Dua and Ana Ramalho and Helen King and Yue Gao and Lynn Nguyen and David Lindner and Divya Pitta and Oleaser Johnson and Khalid Salama and Diego Ardila and Michael Han and Erin Farnese and Seth Odoom and Ziyue Wang and Xiangzhuo Ding and Norman Rink and Ray Smith and Harshal Tushar Lehri and Eden Cohen and Neera Vats and Tong He and Parthasarathy Gopavarapu and Adam Paszke and Miteyan Patel and Wouter Van Gansbeke and Lucia Loher and Luis Castro and Maria Voitovich and Tamara von Glehn and Nelson George and Simon Niklaus and Zach Eaton-Rosen and Nemanja Rakićević and Erik Jue and Sagi Perel and Carrie Zhang and Yuval Bahat and Angéline Pouget and Zhi Xing and Fantine Huot and Ashish Shenoy and Taylor Bos and Vincent Coriou and Bryan Richter and Natasha Noy and Yaqing Wang and Santiago Ontanon and Siyang Qin and Gleb Makarchuk and Demis Hassabis and Zhuowan Li and Mandar Sharma and Kumaran Venkatesan and Iurii Kemaev and Roxanne Daniel and Shiyu Huang and Saloni Shah and Octavio Ponce and Warren and Chen and Manaal Faruqui and Jialin Wu and Slavica Andačić and Szabolcs Payrits and Daniel McDuff and Tom Hume and Yuan Cao and MH Tessler and Qingze Wang and Yinan Wang and Ivor Rendulic and Eirikur Agustsson and Matthew Johnson and Tanya Lando and Andrew Howard and Sri Gayatri Sundara Padmanabhan and Mayank Daswani and Andrea Banino and Michael Kilgore and Jonathan Heek and Ziwei Ji and Alvaro Caceres and Conglong Li and Nora Kassner and Alexey Vlaskin and Zeyu Liu and Alex Grills and Yanhan Hou and Roykrong Sukkerd and Gowoon Cheon and Nishita Shetty and Larisa Markeeva and Piotr Stanczyk and Tejas Iyer and Yuan Gong and Shawn Gao and Keerthana Gopalakrishnan and Tim Blyth and Malcolm Reynolds and Avishkar Bhoopchand and Misha Bilenko and Dero Gharibian and Vicky Zayats and Aleksandra Faust and Abhinav Singh and Min Ma and Hongyang Jiao and Sudheendra Vijayanarasimhan and Lora Aroyo and Vikas Yadav and Sarah Chakera and Ashwin Kakarla and Vilobh Meshram and Karol Gregor and Gabriela Botea and Evan Senter and Dawei Jia and Geza Kovacs and Neha Sharma and Sebastien Baur and Kai Kang and Yifan He and Lin Zhuo and Marija Kostelac and Itay Laish and Songyou Peng and Louis O'Bryan and Daniel Kasenberg and Girish Ramchandra Rao and Edouard Leurent and Biao Zhang and Sage Stevens and Ana Salazar and Ye Zhang and Ivan Lobov and Jake Walker and Allen Porter and Morgan Redshaw and Han Ke and Abhishek Rao and Alex Lee and Hoi Lam and Michael Moffitt and Jaeyoun Kim and Siyuan Qiao and Terry Koo and Robert Dadashi and Xinying Song and Mukund Sundararajan and Peng Xu and Chizu Kawamoto and Yan Zhong and Clara Barbu and Apoorv Reddy and Mauro Verzetti and Leon Li and George Papamakarios and Hanna Klimczak-Plucińska and Mary Cassin and Koray Kavukcuoglu and Rigel Swavely and Alain Vaucher and Jeffrey Zhao and Ross Hemsley and Michael Tschannen and Heming Ge and Gaurav Menghani and Yang Yu and Natalie Ha and Wei He and Xiao Wu and Maggie Song and Rachel Sterneck and Stefan Zinke and Dan A. Calian and Annie Marsden and Alejandro Cruzado Ruiz and Matteo Hessel and Almog Gueta and Benjamin Lee and Brian Farris and Manish Gupta and Yunjie Li and Mohammad Saleh and Vedant Misra and Kefan Xiao and Piermaria Mendolicchio and Gavin Buttimore and Varvara Krayvanova and Nigamaa Nayakanti and Matthew Wiethoff and Yash Pande and Azalia Mirhoseini and Ni Lao and Jasmine Liu and Yiqing Hua and Angie Chen and Yury Malkov and Dmitry Kalashnikov and Shubham Gupta and Kartik Audhkhasi and Yuexiang Zhai and Sudhindra Kopalle and Prateek Jain and Eran Ofek and Clemens Meyer and Khuslen Baatarsukh and Hana Strejček and Jun Qian and James Freedman and Ricardo Figueira and Michal Sokolik and Olivier Bachem and Raymond Lin and Dia Kharrat and Chris Hidey and Pingmei Xu and Dennis Duan and Yin Li and Muge Ersoy and Richard Everett and Kevin Cen and Rebeca Santamaria-Fernandez and Amir Taubenfeld and Ian Mackinnon and Linda Deng and Polina Zablotskaia and Shashank Viswanadha and Shivanker Goel and Damion Yates and Yunxiao Deng and Peter Choy and Mingqing Chen and Abhishek Sinha and Alex Mossin and Yiming Wang and Arthur Szlam and Susan Hao and Paul Kishan Rubenstein and Metin Toksoz-Exley and Miranda Aperghis and Yin Zhong and Junwhan Ahn and Michael Isard and Olivier Lacombe and Florian Luisier and Chrysovalantis Anastasiou and Yogesh Kalley and Utsav Prabhu and Emma Dunleavy and Shaan Bijwadia and Justin Mao-Jones and Kelly Chen and Rama Pasumarthi and Emily Wood and Adil Dostmohamed and Nate Hurley and Jiri Simsa and Alicia Parrish and Mantas Pajarskas and Matt Harvey and Ondrej Skopek and Yony Kochinski and Javier Rey and Verena Rieser and Denny Zhou and Sun Jae Lee and Trilok Acharya and Guowang Li and Joe Jiang and Xiaofan Zhang and Bryant Gipson and Ethan Mahintorabi and Marco Gelmi and Nima Khajehnouri and Angel Yeh and Kayi Lee and Loic Matthey and Leslie Baker and Trang Pham and Han Fu and Alex Pak and Prakhar Gupta and Cristina Vasconcelos and Adam Sadovsky and Brian Walker and Sissie Hsiao and Patrik Zochbauer and Andreea Marzoca and Noam Velan and Junhao Zeng and Gilles Baechler and Danny Driess and Divya Jain and Yanping Huang and Lizzie Tao and John Maggs and Nir Levine and Jon Schneider and Erika Gemzer and Samuel Petit and Shan Han and Zach Fisher and Dustin Zelle and Courtney Biles and Eugene Ie and Asya Fadeeva and Casper Liu and Juliana Vicente Franco and Adrian Collister and Hao Zhang and Renshen Wang and Ruizhe Zhao and Leandro Kieliger and Kurt Shuster and Rui Zhu and Boqing Gong and Lawrence Chan and Ruoxi Sun and Sujoy Basu and Roland Zimmermann and Jamie Hayes and Abhishek Bapna and Jasper Snoek and Weel Yang and Puranjay Datta and Jad Al Abdallah and Kevin Kilgour and Lu Li and SQ Mah and Yennie Jun and Morgane Rivière and Abhijit Karmarkar and Tammo Spalink and Tao Huang and Lucas Gonzalez and Duc-Hieu Tran and Averi Nowak and John Palowitch and Martin Chadwick and Ellie Talius and Harsh Mehta and Thibault Sellam and Philipp Fränken and Massimo Nicosia and Kyle He and Aditya Kini and David Amos and Sugato Basu and Harrison Jobe and Eleni Shaw and Qiantong Xu and Colin Evans and Daisuke Ikeda and Chaochao Yan and Larry Jin and Lun Wang and Sachin Yadav and Ilia Labzovsky and Ramesh Sampath and Ada Ma and Candice Schumann and Aditya Siddhant and Rohin Shah and John Youssef and Rishabh Agarwal and Natalie Dabney and Alessio Tonioni and Moran Ambar and Jing Li and Isabelle Guyon and Benny Li and David Soergel and Boya Fang and Georgi Karadzhov and Cristian Udrescu and Trieu Trinh and Vikas Raunak and Seb Noury and Dee Guo and Sonal Gupta and Mara Finkelstein and Denis Petek and Lihao Liang and Greg Billock and Pei Sun and David Wood and Yiwen Song and Xiaobin Yu and Tatiana Matejovicova and Regev Cohen and Kalyan Andra and David D'Ambrosio and Zhiwei Deng and Vincent Nallatamby and Ebrahim Songhori and Rumen Dangovski and Andrew Lampinen and Pankil Botadra and Adam Hillier and Jiawei Cao and Nagabhushan Baddi and Adhi Kuncoro and Toshihiro Yoshino and Ankit Bhagatwala and Marcáurelio Ranzato and Rylan Schaeffer and Tianlin Liu and Shuai Ye and Obaid Sarvana and John Nham and Chenkai Kuang and Isabel Gao and Jinoo Baek and Shubham Mittal and Ayzaan Wahid and Anita Gergely and Bin Ni and Josh Feldman and Carrie Muir and Pascal Lamblin and Wolfgang Macherey and Ethan Dyer and Logan Kilpatrick and Víctor Campos and Mukul Bhutani and Stanislav Fort and Yanif Ahmad and Aliaksei Severyn and Kleopatra Chatziprimou and Oleksandr Ferludin and Mason Dimarco and Aditya Kusupati and Joe Heyward and Dan Bahir and Kevin Villela and Katie Millican and Dror Marcus and Sanaz Bahargam and Caglar Unlu and Nicholas Roth and Zichuan Wei and Siddharth Gopal and Deepanway Ghoshal and Edward Lee and Sharon Lin and Jennie Lees and Dayeong Lee and Anahita Hosseini and Connie Fan and Seth Neel and Marcus Wu and Yasemin Altun and Honglong Cai and Enrique Piqueras and Josh Woodward and Alessandro Bissacco and Salem Haykal and Mahyar Bordbar and Prasha Sundaram and Sarah Hodkinson and Daniel Toyama and George Polovets and Austin Myers and Anu Sinha and Tomer Levinboim and Kashyap Krishnakumar and Rachita Chhaparia and Tatiana Sholokhova and Nitesh Bharadwaj Gundavarapu and Ganesh Jawahar and Haroon Qureshi and Jieru Hu and Nikola Momchev and Matthew Rahtz and Renjie Wu and Aishwarya P S and Kedar Dhamdhere and Meiqi Guo and Umang Gupta and Ali Eslami and Mariano Schain and Michiel Blokzijl and David Welling and Dave Orr and Levent Bolelli and Nicolas Perez-Nieves and Mikhail Sirotenko and Aman Prasad and Arjun Kar and Borja De Balle Pigem and Tayfun Terzi and Gellért Weisz and Dipankar Ghosh and Aditi Mavalankar and Dhruv Madeka and Kaspar Daugaard and Hartwig Adam and Viraj Shah and Dana Berman and Maggie Tran and Steven Baker and Ewa Andrejczuk and Grishma Chole and Ganna Raboshchuk and Mahdi Mirzazadeh and Thais Kagohara and Shimu Wu and Christian Schallhart and Bernett Orlando and Chen Wang and Alban Rrustemi and Hao Xiong and Hao Liu and Arpi Vezer and Nolan Ramsden and Shuo-yiin Chang and Sidharth Mudgal and Yan Li and Nino Vieillard and Yedid Hoshen and Farooq Ahmad and Ambrose Slone and Amy Hua and Natan Potikha and Mirko Rossini and Jon Stritar and Sushant Prakash and Zifeng Wang and Xuanyi Dong and Alireza Nazari and Efrat Nehoran and Kaan Tekelioglu and Yinxiao Li and Kartikeya Badola and Tom Funkhouser and Yuanzhen Li and Varun Yerram and Ramya Ganeshan and Daniel Formoso and Karol Langner and Tian Shi and Huijian Li and Yumeya Yamamori and Amayika Panda and Alaa Saade and Angelo Scorza Scarpati and Chris Breaux and CJ Carey and Zongwei Zhou and Cho-Jui Hsieh and Sophie Bridgers and Alena Butryna and Nishesh Gupta and Vaibhav Tulsyan and Sanghyun Woo and Evgenii Eltyshev and Will Grathwohl and Chanel Parks and Seth Benjamin and Rina Panigrahy and Shenil Dodhia and Daniel De Freitas and Chris Sauer and Will Song and Ferran Alet and Jackson Tolins and Cosmin Paduraru and Xingyi Zhou and Brian Albert and Zizhao Zhang and Lei Shu and Mudit Bansal and Sarah Nguyen and Amir Globerson and Owen Xiao and James Manyika and Tom Hennigan and Rong Rong and Josip Matak and Anton Bakalov and Ankur Sharma and Danila Sinopalnikov and Andrew Pierson and Stephen Roller and Geoff Brown and Mingcen Gao and Toshiyuki Fukuzawa and Amin Ghafouri and Kenny Vassigh and Iain Barr and Zhicheng Wang and Anna Korsun and Rajesh Jayaram and Lijie Ren and Tim Zaman and Samira Khan and Yana Lunts and Dan Deutsch and Dave Uthus and Nitzan Katz and Masha Samsikova and Amr Khalifa and Nikhil Sethi and Jiao Sun and Luming Tang and Uri Alon and Xianghong Luo and Dian Yu and Abhishek Nayyar and Bryce Petrini and Will Truong and Vincent Hellendoorn and Nikolai Chinaev and Chris Alberti and Wei Wang and Jingcao Hu and Vahab Mirrokni and Ananth Balashankar and Avia Aharon and Aahil Mehta and Ahmet Iscen and Joseph Kready and Lucas Manning and Anhad Mohananey and Yuankai Chen and Anshuman Tripathi and Allen Wu and Igor Petrovski and Dawsen Hwang and Martin Baeuml and Shreyas Chandrakaladharan and Yuan Liu and Rey Coaguila and Maxwell Chen and Sally Ma and Pouya Tafti and Susheel Tatineni and Terry Spitz and Jiayu Ye and Paul Vicol and Mihaela Rosca and Adrià Puigdomènech and Zohar Yahav and Sanjay Ghemawat and Hanzhao Lin and Phoebe Kirk and Zaid Nabulsi and Sergey Brin and Bernd Bohnet and Ken Caluwaerts and Aditya Srikanth Veerubhotla and Dan Zheng and Zihang Dai and Petre Petrov and Yichong Xu and Ramin Mehran and Zhuo Xu and Luisa Zintgraf and Jiho Choi and Spurthi Amba Hombaiah and Romal Thoppilan and Sashank Reddi and Lukasz Lew and Li Li and Kellie Webster and KP Sawhney and Lampros Lamprou and Siamak Shakeri and Mayank Lunayach and Jianmin Chen and Sumit Bagri and Alex Salcianu and Ying Chen and Yani Donchev and Charlotte Magister and Signe Nørly and Vitor Rodrigues and Tomas Izo and Hila Noga and Joe Zou and Thomas Köppe and Wenxuan Zhou and Kenton Lee and Xiangzhu Long and Danielle Eisenbud and Anthony Chen and Connor Schenck and Chi Ming To and Peilin Zhong and Emanuel Taropa and Minh Truong and Omer Levy and Danilo Martins and Zhiyuan Zhang and Christopher Semturs and Kelvin Zhang and Alex Yakubovich and Pol Moreno and Lara McConnaughey and Di Lu and Sam Redmond and Lotte Weerts and Yonatan Bitton and Tiziana Refice and Nicolas Lacasse and Arthur Conmy and Corentin Tallec and Julian Odell and Hannah Forbes-Pollard and Arkadiusz Socala and Jonathan Hoech and Pushmeet Kohli and Alanna Walton and Rui Wang and Mikita Sazanovich and Kexin Zhu and Andrei Kapishnikov and Rich Galt and Matthew Denton and Ben Murdoch and Caitlin Sikora and Kareem Mohamed and Wei Wei and Uri First and Tim McConnell and Luis C. Cobo and James Qin and Thi Avrahami and Daniel Balle and Yu Watanabe and Annie Louis and Adam Kraft and Setareh Ariafar and Yiming Gu and Eugénie Rives and Charles Yoon and Andrei Rusu and James Cobon-Kerr and Chris Hahn and Jiaming Luo and Yuvein and Zhu and Niharika Ahuja and Rodrigo Benenson and Raphaël Lopez Kaufman and Honglin Yu and Lloyd Hightower and Junlin Zhang and Darren Ni and Lisa Anne Hendricks and Gabby Wang and Gal Yona and Lalit Jain and Pablo Barrio and Surya Bhupatiraju and Siva Velusamy and Allan Dafoe and Sebastian Riedel and Tara Thomas and Zhe Yuan and Mathias Bellaiche and Sheena Panthaplackel and Klemen Kloboves and Sarthak Jauhari and Canfer Akbulut and Todor Davchev and Evgeny Gladchenko and David Madras and Aleksandr Chuklin and Tyrone Hill and Quan Yuan and Mukundan Madhavan and Luke Leonhard and Dylan Scandinaro and Qihang Chen and Ning Niu and Arthur Douillard and Bogdan Damoc and Yasumasa Onoe and Fabian Pedregosa and Fred Bertsch and Chas Leichner and Joseph Pagadora and Jonathan Malmaud and Sameera Ponda and Andy Twigg and Oleksii Duzhyi and Jingwei Shen and Miaosen Wang and Roopal Garg and Jing Chen and Utku Evci and Jonathan Lee and Leon Liu and Koji Kojima and Masa Yamaguchi and Arunkumar Rajendran and AJ Piergiovanni and Vinodh Kumar Rajendran and Marco Fornoni and Gabriel Ibagon and Harry Ragan and Sadh MNM Khan and John Blitzer and Andrew Bunner and Guan Sun and Takahiro Kosakai and Scott Lundberg and Ndidi Elue and Kelvin Guu and SK Park and Jane Park and Arunachalam Narayanaswamy and Chengda Wu and Jayaram Mudigonda and Trevor Cohn and Hairong Mu and Ravi Kumar and Laura Graesser and Yichi Zhang and Richard Killam and Vincent Zhuang and Mai Giménez and Wael Al Jishi and Ruy Ley-Wild and Alex Zhai and Kazuki Osawa and Diego Cedillo and Jialu Liu and Mayank Upadhyay and Marcin Sieniek and Roshan Sharma and Tom Paine and Anelia Angelova and Sravanti Addepalli and Carolina Parada and Kingshuk Majumder and Avery Lamp and Sanjiv Kumar and Xiang Deng and Artiom Myaskovsky and Tea Sabolić and Jeffrey Dudek and Sarah York and Félix de Chaumont Quitry and Jiazhong Nie and Dee Cattle and Alok Gunjan and Bilal Piot and Waleed Khawaja and Seojin Bang and Simon Wang and Siavash Khodadadeh and Raghavender R and Praynaa Rawlani and Richard Powell and Kevin Lee and Johannes Griesser and GS Oh and Cesar Magalhaes and Yujia Li and Simon Tokumine and Hadas Natalie Vogel and Dennis Hsu and Arturo BC and Disha Jindal and Matan Cohen and Zi Yang and Junwei Yuan and Dario de Cesare and Tony Bruguier and Jun Xu and Monica Roy and Alon Jacovi and Dan Belov and Rahul Arya and Phoenix Meadowlark and Shlomi Cohen-Ganor and Wenting Ye and Patrick Morris-Suzuki and Praseem Banzal and Gan Song and Pranavaraj Ponnuramu and Fred Zhang and George Scrivener and Salah Zaiem and Alif Raditya Rochman and Kehang Han and Badih Ghazi and Kate Lee and Shahar Drath and Daniel Suo and Antonious Girgis and Pradeep Shenoy and Duy Nguyen and Douglas Eck and Somit Gupta and Le Yan and Joao Carreira and Anmol Gulati and Ruoxin Sang and Daniil Mirylenka and Emma Cooney and Edward Chou and Mingyang Ling and Cindy Fan and Ben Coleman and Guilherme Tubone and Ravin Kumar and Jason Baldridge and Felix Hernandez-Campos and Angeliki Lazaridou and James Besley and Itay Yona and Neslihan Bulut and Quentin Wellens and AJ Pierigiovanni and Jasmine George and Richard Green and Pu Han and Connie Tao and Geoff Clark and Chong You and Abbas Abdolmaleki and Justin Fu and Tongzhou Chen and Ashwin Chaugule and Angad Chandorkar and Altaf Rahman and Will Thompson and Penporn Koanantakool and Mike Bernico and Jie Ren and Andrey Vlasov and Sergei Vassilvitskii and Maciej Kula and Yizhong Liang and Dahun Kim and Yangsibo Huang and Chengxi Ye and Dmitry Lepikhin and Wesley Helmholz},
      year={2025},
      eprint={2507.06261},
      archivePrefix={arXiv},
      primaryClass={cs.CL},
      url={https://arxiv.org/abs/2507.06261}, 
}

@article{liu2025smartgrid,
title = {Time Synchronization Techniques in the Modern Smart Grid: A Comprehensive Survey},
author = {Liu, Yu and Sun, Biao and Wu, Yuru and Zhang, Yongxin and Yang, Jiahui and Wang, Wen and Thotakura, Naga Lakshmi and Liu, Qian and Liu, Yilu},
abstractNote = {In modern smart grids, accurate and synchronized time signals are essential for effective monitoring, protection, and control. Various time synchronization methods exist, each tailored to specific application needs. Widely adopted solutions, such as GPS, however, are vulnerable to challenges such as signal loss and cyber-attacks, underscoring the need for reliable backup or supplementary solutions. This paper examines the timing requirements across different power grid applications and provides a comprehensive review of available time synchronization mechanisms. Through a comparative analysis of timing methods based on accuracy, flexibility, reliability, and security, this study offers insights to guide the selection of optimal solutions for seamless grid integration.},
doi = {10.3390/en18051163},
journal = {Energies},
number = 5,
volume = 18,
place = {United States},
year = {2025},
month = {2}
}

@article{guo2017gps1588,
  author={Guo, Hao and Crossley, Peter},
  journal={IEEE Transactions on Power Delivery}, 
  title={Design of a Time Synchronization System Based on GPS and IEEE 1588 for Transmission Substations}, 
  year={2017},
  volume={32},
  number={4},
  pages={2091-2100},
  doi={10.1109/TPWRD.2016.2600759}}

@article{balakrishnan2023iiotptp,
title = {Clock synchronization in industrial Internet of Things and potential works in precision time protocol: Review, challenges and future directions},
journal = {International Journal of Cognitive Computing in Engineering},
volume = {4},
pages = {205-219},
year = {2023},
issn = {2666-3074},
doi = {https://doi.org/10.1016/j.ijcce.2023.06.001},
url = {https://www.sciencedirect.com/science/article/pii/S2666307423000219},
author = {Balakrishnan K and Dhanalakshmi R and Bam Bahadur Sinha and Gopalakrishnan R}
}

@article{lombardi2016financialtime,
  author    = {Michael A. Lombardi and Andrew N. Novick and George Neville-Neil and Ben Cooke},
  title     = {Accurate, Traceable, and Verifiable Time Synchronization for World Financial Markets},
  journal   = {Journal of Research of the National Institute of Standards and Technology},
  year      = {2016},
  volume    = {121},
  pages     = {436--463},
  doi       = {10.6028/jres.121.023},
  publisher = {National Institute of Standards and Technology}
}

@article{haber1991timestamp,
  author    = {Stuart Haber and W. Scott Stornetta},
  title     = {How to time-stamp a digital document},
  journal   = {Journal of Cryptology},
  year      = {1991},
  volume    = {3},
  pages     = {99--111},
  doi       = {10.1007/BF00196791},
  publisher = {Springer}
}

@article{vanini2024clockcorrect,
title = {Was the clock correct? Exploring timestamp interpretation through time anchors for digital forensic event reconstruction},
journal = {Forensic Science International: Digital Investigation},
volume = {49},
pages = {301759},
year = {2024},
note = {DFRWS USA 2024 - Selected Papers from the 24th Annual Digital Forensics Research Conference USA},
issn = {2666-2817},
doi = {https://doi.org/10.1016/j.fsidi.2024.301759},
url = {https://www.sciencedirect.com/science/article/pii/S2666281724000787},
author = {Céline Vanini and Christopher J. Hargreaves and Harm {van Beek} and Frank Breitinger}
}

@misc{grattafiori2024llama3herdmodels,
      title={The Llama 3 Herd of Models}, 
      author={Aaron Grattafiori and Abhimanyu Dubey and Abhinav Jauhri and Abhinav Pandey and Abhishek Kadian and Ahmad Al-Dahle and Aiesha Letman and Akhil Mathur and Alan Schelten and Alex Vaughan and Amy Yang and Angela Fan and Anirudh Goyal and Anthony Hartshorn and Aobo Yang and Archi Mitra and Archie Sravankumar and Artem Korenev and Arthur Hinsvark and Arun Rao and Aston Zhang and Aurelien Rodriguez and Austen Gregerson and Ava Spataru and Baptiste Roziere and Bethany Biron and Binh Tang and Bobbie Chern and Charlotte Caucheteux and Chaya Nayak and Chloe Bi and Chris Marra and Chris McConnell and Christian Keller and Christophe Touret and Chunyang Wu and Corinne Wong and Cristian Canton Ferrer and Cyrus Nikolaidis and Damien Allonsius and Daniel Song and Danielle Pintz and Danny Livshits and Danny Wyatt and David Esiobu and Dhruv Choudhary and Dhruv Mahajan and Diego Garcia-Olano and Diego Perino and Dieuwke Hupkes and Egor Lakomkin and Ehab AlBadawy and Elina Lobanova and Emily Dinan and Eric Michael Smith and Filip Radenovic and Francisco Guzmán and Frank Zhang and Gabriel Synnaeve and Gabrielle Lee and Georgia Lewis Anderson and Govind Thattai and Graeme Nail and Gregoire Mialon and Guan Pang and Guillem Cucurell and Hailey Nguyen and Hannah Korevaar and Hu Xu and Hugo Touvron and Iliyan Zarov and Imanol Arrieta Ibarra and Isabel Kloumann and Ishan Misra and Ivan Evtimov and Jack Zhang and Jade Copet and Jaewon Lee and Jan Geffert and Jana Vranes and Jason Park and Jay Mahadeokar and Jeet Shah and Jelmer van der Linde and Jennifer Billock and Jenny Hong and Jenya Lee and Jeremy Fu and Jianfeng Chi and Jianyu Huang and Jiawen Liu and Jie Wang and Jiecao Yu and Joanna Bitton and Joe Spisak and Jongsoo Park and Joseph Rocca and Joshua Johnstun and Joshua Saxe and Junteng Jia and Kalyan Vasuden Alwala and Karthik Prasad and Kartikeya Upasani and Kate Plawiak and Ke Li and Kenneth Heafield and Kevin Stone and Khalid El-Arini and Krithika Iyer and Kshitiz Malik and Kuenley Chiu and Kunal Bhalla and Kushal Lakhotia and Lauren Rantala-Yeary and Laurens van der Maaten and Lawrence Chen and Liang Tan and Liz Jenkins and Louis Martin and Lovish Madaan and Lubo Malo and Lukas Blecher and Lukas Landzaat and Luke de Oliveira and Madeline Muzzi and Mahesh Pasupuleti and Mannat Singh and Manohar Paluri and Marcin Kardas and Maria Tsimpoukelli and Mathew Oldham and Mathieu Rita and Maya Pavlova and Melanie Kambadur and Mike Lewis and Min Si and Mitesh Kumar Singh and Mona Hassan and Naman Goyal and Narjes Torabi and Nikolay Bashlykov and Nikolay Bogoychev and Niladri Chatterji and Ning Zhang and Olivier Duchenne and Onur Çelebi and Patrick Alrassy and Pengchuan Zhang and Pengwei Li and Petar Vasic and Peter Weng and Prajjwal Bhargava and Pratik Dubal and Praveen Krishnan and Punit Singh Koura and Puxin Xu and Qing He and Qingxiao Dong and Ragavan Srinivasan and Raj Ganapathy and Ramon Calderer and Ricardo Silveira Cabral and Robert Stojnic and Roberta Raileanu and Rohan Maheswari and Rohit Girdhar and Rohit Patel and Romain Sauvestre and Ronnie Polidoro and Roshan Sumbaly and Ross Taylor and Ruan Silva and Rui Hou and Rui Wang and Saghar Hosseini and Sahana Chennabasappa and Sanjay Singh and Sean Bell and Seohyun Sonia Kim and Sergey Edunov and Shaoliang Nie and Sharan Narang and Sharath Raparthy and Sheng Shen and Shengye Wan and Shruti Bhosale and Shun Zhang and Simon Vandenhende and Soumya Batra and Spencer Whitman and Sten Sootla and Stephane Collot and Suchin Gururangan and Sydney Borodinsky and Tamar Herman and Tara Fowler and Tarek Sheasha and Thomas Georgiou and Thomas Scialom and Tobias Speckbacher and Todor Mihaylov and Tong Xiao and Ujjwal Karn and Vedanuj Goswami and Vibhor Gupta and Vignesh Ramanathan and Viktor Kerkez and Vincent Gonguet and Virginie Do and Vish Vogeti and Vítor Albiero and Vladan Petrovic and Weiwei Chu and Wenhan Xiong and Wenyin Fu and Whitney Meers and Xavier Martinet and Xiaodong Wang and Xiaofang Wang and Xiaoqing Ellen Tan and Xide Xia and Xinfeng Xie and Xuchao Jia and Xuewei Wang and Yaelle Goldschlag and Yashesh Gaur and Yasmine Babaei and Yi Wen and Yiwen Song and Yuchen Zhang and Yue Li and Yuning Mao and Zacharie Delpierre Coudert and Zheng Yan and Zhengxing Chen and Zoe Papakipos and Aaditya Singh and Aayushi Srivastava and Abha Jain and Adam Kelsey and Adam Shajnfeld and Adithya Gangidi and Adolfo Victoria and Ahuva Goldstand and Ajay Menon and Ajay Sharma and Alex Boesenberg and Alexei Baevski and Allie Feinstein and Amanda Kallet and Amit Sangani and Amos Teo and Anam Yunus and Andrei Lupu and Andres Alvarado and Andrew Caples and Andrew Gu and Andrew Ho and Andrew Poulton and Andrew Ryan and Ankit Ramchandani and Annie Dong and Annie Franco and Anuj Goyal and Aparajita Saraf and Arkabandhu Chowdhury and Ashley Gabriel and Ashwin Bharambe and Assaf Eisenman and Azadeh Yazdan and Beau James and Ben Maurer and Benjamin Leonhardi and Bernie Huang and Beth Loyd and Beto De Paola and Bhargavi Paranjape and Bing Liu and Bo Wu and Boyu Ni and Braden Hancock and Bram Wasti and Brandon Spence and Brani Stojkovic and Brian Gamido and Britt Montalvo and Carl Parker and Carly Burton and Catalina Mejia and Ce Liu and Changhan Wang and Changkyu Kim and Chao Zhou and Chester Hu and Ching-Hsiang Chu and Chris Cai and Chris Tindal and Christoph Feichtenhofer and Cynthia Gao and Damon Civin and Dana Beaty and Daniel Kreymer and Daniel Li and David Adkins and David Xu and Davide Testuggine and Delia David and Devi Parikh and Diana Liskovich and Didem Foss and Dingkang Wang and Duc Le and Dustin Holland and Edward Dowling and Eissa Jamil and Elaine Montgomery and Eleonora Presani and Emily Hahn and Emily Wood and Eric-Tuan Le and Erik Brinkman and Esteban Arcaute and Evan Dunbar and Evan Smothers and Fei Sun and Felix Kreuk and Feng Tian and Filippos Kokkinos and Firat Ozgenel and Francesco Caggioni and Frank Kanayet and Frank Seide and Gabriela Medina Florez and Gabriella Schwarz and Gada Badeer and Georgia Swee and Gil Halpern and Grant Herman and Grigory Sizov and Guangyi and Zhang and Guna Lakshminarayanan and Hakan Inan and Hamid Shojanazeri and Han Zou and Hannah Wang and Hanwen Zha and Haroun Habeeb and Harrison Rudolph and Helen Suk and Henry Aspegren and Hunter Goldman and Hongyuan Zhan and Ibrahim Damlaj and Igor Molybog and Igor Tufanov and Ilias Leontiadis and Irina-Elena Veliche and Itai Gat and Jake Weissman and James Geboski and James Kohli and Janice Lam and Japhet Asher and Jean-Baptiste Gaya and Jeff Marcus and Jeff Tang and Jennifer Chan and Jenny Zhen and Jeremy Reizenstein and Jeremy Teboul and Jessica Zhong and Jian Jin and Jingyi Yang and Joe Cummings and Jon Carvill and Jon Shepard and Jonathan McPhie and Jonathan Torres and Josh Ginsburg and Junjie Wang and Kai Wu and Kam Hou U and Karan Saxena and Kartikay Khandelwal and Katayoun Zand and Kathy Matosich and Kaushik Veeraraghavan and Kelly Michelena and Keqian Li and Kiran Jagadeesh and Kun Huang and Kunal Chawla and Kyle Huang and Lailin Chen and Lakshya Garg and Lavender A and Leandro Silva and Lee Bell and Lei Zhang and Liangpeng Guo and Licheng Yu and Liron Moshkovich and Luca Wehrstedt and Madian Khabsa and Manav Avalani and Manish Bhatt and Martynas Mankus and Matan Hasson and Matthew Lennie and Matthias Reso and Maxim Groshev and Maxim Naumov and Maya Lathi and Meghan Keneally and Miao Liu and Michael L. Seltzer and Michal Valko and Michelle Restrepo and Mihir Patel and Mik Vyatskov and Mikayel Samvelyan and Mike Clark and Mike Macey and Mike Wang and Miquel Jubert Hermoso and Mo Metanat and Mohammad Rastegari and Munish Bansal and Nandhini Santhanam and Natascha Parks and Natasha White and Navyata Bawa and Nayan Singhal and Nick Egebo and Nicolas Usunier and Nikhil Mehta and Nikolay Pavlovich Laptev and Ning Dong and Norman Cheng and Oleg Chernoguz and Olivia Hart and Omkar Salpekar and Ozlem Kalinli and Parkin Kent and Parth Parekh and Paul Saab and Pavan Balaji and Pedro Rittner and Philip Bontrager and Pierre Roux and Piotr Dollar and Polina Zvyagina and Prashant Ratanchandani and Pritish Yuvraj and Qian Liang and Rachad Alao and Rachel Rodriguez and Rafi Ayub and Raghotham Murthy and Raghu Nayani and Rahul Mitra and Rangaprabhu Parthasarathy and Raymond Li and Rebekkah Hogan and Robin Battey and Rocky Wang and Russ Howes and Ruty Rinott and Sachin Mehta and Sachin Siby and Sai Jayesh Bondu and Samyak Datta and Sara Chugh and Sara Hunt and Sargun Dhillon and Sasha Sidorov and Satadru Pan and Saurabh Mahajan and Saurabh Verma and Seiji Yamamoto and Sharadh Ramaswamy and Shaun Lindsay and Shaun Lindsay and Sheng Feng and Shenghao Lin and Shengxin Cindy Zha and Shishir Patil and Shiva Shankar and Shuqiang Zhang and Shuqiang Zhang and Sinong Wang and Sneha Agarwal and Soji Sajuyigbe and Soumith Chintala and Stephanie Max and Stephen Chen and Steve Kehoe and Steve Satterfield and Sudarshan Govindaprasad and Sumit Gupta and Summer Deng and Sungmin Cho and Sunny Virk and Suraj Subramanian and Sy Choudhury and Sydney Goldman and Tal Remez and Tamar Glaser and Tamara Best and Thilo Koehler and Thomas Robinson and Tianhe Li and Tianjun Zhang and Tim Matthews and Timothy Chou and Tzook Shaked and Varun Vontimitta and Victoria Ajayi and Victoria Montanez and Vijai Mohan and Vinay Satish Kumar and Vishal Mangla and Vlad Ionescu and Vlad Poenaru and Vlad Tiberiu Mihailescu and Vladimir Ivanov and Wei Li and Wenchen Wang and Wenwen Jiang and Wes Bouaziz and Will Constable and Xiaocheng Tang and Xiaojian Wu and Xiaolan Wang and Xilun Wu and Xinbo Gao and Yaniv Kleinman and Yanjun Chen and Ye Hu and Ye Jia and Ye Qi and Yenda Li and Yilin Zhang and Ying Zhang and Yossi Adi and Youngjin Nam and Yu and Wang and Yu Zhao and Yuchen Hao and Yundi Qian and Yunlu Li and Yuzi He and Zach Rait and Zachary DeVito and Zef Rosnbrick and Zhaoduo Wen and Zhenyu Yang and Zhiwei Zhao and Zhiyu Ma},
      year={2024},
      eprint={2407.21783},
      archivePrefix={arXiv},
      primaryClass={cs.AI},
      url={https://arxiv.org/abs/2407.21783}, 
}

@misc{openai_gpt5_nano_2026,
  author       = {{OpenAI}},
  title        = {{GPT-5-nano}},
  year         = {2026},
  howpublished = {\url{https://openai.com}},
  note         = {Proprietary large language model},
}


\newpage

\appendix
\renewcommand{\thetable}{A\arabic{table}}
\setcounter{table}{0}

\section{Dataset Details}
\label{app:dataset}

\subsection{Artifact License and Usage Terms}

The CivilTime-Bench dataset and evaluation code are publicly available at
\url{https://github.com/CivilTime-Bench/CivilTime-Bench}
under the Apache License~2.0.
The dataset is synthetically generated and does not contain real user logs
or personally identifiable information.
Use of third-party model checkpoints remains subject to their respective
licenses and terms of use.

\subsection{Realistic Scenario Dataset: HumanTime-500}
\label{sec:humantime500}

To complement our programmatically generated dataset, which provides controlled and systematic coverage, we introduce HumanTime-500, a naturalistic evaluation set containing civil-time problems resembling those encountered in everyday software and operational workflows. Examples include scheduling, calendar coordination, log normalization, billing cutoffs, compliance deadlines, and reporting.

These settings require calendrical operations such as time-zone conversion, International Date Line transitions, daylight saving time (DST) discontinuities, and business-day calculations.

Rather than using real requests, tickets, or user logs, we generate fictionalized scenarios with \textbf{GPT-5 Pro}. The model serves as a privacy-preserving authoring tool, producing application-grounded examples without reproducing individual users' data. HumanTime-500 is therefore \textbf{LLM-authored and human-validated}, rather than human-authored.

\paragraph{Composition.}

HumanTime-500 contains \textbf{500} English evaluation items across \textbf{11} task categories:

\begin{itemize}[nosep, leftmargin=1.3em]
\item \texttt{BUSINESS\_DAY}: 60
\item \texttt{OFFSET}: 60
\item \texttt{DATE\_LINE}: 50
\item \texttt{RELATIVE\_DAY}: 50
\item \texttt{DST\_SKIP}: 50
\item \texttt{DST\_REPEAT}: 50
\item \texttt{DST\_AMBIG}: 50
\item \texttt{INTERVAL}: 40
\item \texttt{CONSTRAINTS}: 40
\item \texttt{ISO\_WEEK}: 25
\item \texttt{WEEKSTART}: 25
\end{itemize}

Scenarios cover a range of IANA time zones, including DST-observing regions and fractional UTC offsets. The dataset is evaluation-only and is not used for fine-tuning or selecting in-context exemplars.

HumanTime-500 provides a complementary test of transfer beyond the synthetic templates, but its limited size and LLM-authored nature constrain claims about generalization to naturally occurring user requests.

\paragraph{Format and Answer Contract.}

Each item is represented as a JSONL record containing fields such as \texttt{id}, \texttt{type}, \texttt{tz}, \texttt{input\_text}, and \texttt{gold}. Cross-time-zone tasks may additionally specify source and destination time zones.

Gold answers follow the answer forms defined in Section~\ref{sec:output-contract}: \texttt{ABSTAIN}, dates and ISO week labels, timestamps with explicit UTC offsets, ambiguity pairs (\texttt{A||B}), and intervals (\texttt{A..B}).

In HumanTime-500, all 40 \texttt{INTERVAL} gold answers use \texttt{A..B}. The 25 \texttt{ISO\_WEEK} gold answers are represented as dates (\texttt{YYYY-MM-DD}), whereas the synthetic evaluation splits use ISO week labels (\texttt{YYYY-Www}).

For ambiguity pairs, the intended representation identifies the valid instants under the stated policy. When no valid answer exists under that policy, the expected answer is \texttt{ABSTAIN}.

The evaluator separately measures raw-output contract compliance, extracted-answer format validity, and exact-match correctness. The v2 raw-output contract also accepts a JSON object containing a canonical answer value, without extracting answers from surrounding prose.

\paragraph{Human Validation.}

Three human evaluators review the generated scenarios for plausibility and correctness, provide independent answers, and cross-check and adjudicate disagreements during curation.

The resulting benchmark combines two complementary resources: (i) SyntheticTimeBench, which provides controlled coverage of calendrical edge cases, and (ii) HumanTime-500, which evaluates transfer to more naturalistic, LLM-authored and human-validated scenarios.

\subsection{Category-Specific Generators}
\label{app:dataset-generators}

SyntheticTimeBench contains \textbf{60,011 training examples} and \textbf{10,000 synthetic evaluation examples}. Both splits are generated using the same category-specific generators, with disjoint examples. This supports controlled evaluation but also limits the extent to which synthetic performance demonstrates generalization beyond the generator distribution.

Generation uses seeded randomness, IANA time-zone identifiers, and \texttt{pytz}-based localization. The primary generation period is 2024--2026, with explicitly selected dates outside this range for certain ISO week-boundary tasks.

The principal synthetic generators cover the following categories.

\begin{enumerate}[leftmargin=1.3em, itemsep=3pt]

\item \textbf{\texttt{OFFSET}}
\begin{itemize}[nosep]
\item \emph{Inputs:} Time zone sampled from regional pools; selected months include March, April, and October; minute increments are sampled from $\{15,30,45,60,90,120,180\}$.
\item \emph{Construction:} Localize the starting civil time and perform the specified time arithmetic.
\item \emph{Output:} Canonical timestamp with explicit UTC offset.
\end{itemize}

\item \textbf{\texttt{DATE\_LINE}}
\begin{itemize}[nosep]
\item \emph{Inputs:} Source time zones include \texttt{Pacific/Honolulu}, \texttt{Pacific/Apia}, and \texttt{Pacific/Tongatapu}; destination time zones include \texttt{Pacific/Kiritimati}, \texttt{Pacific/Tarawa}, and \texttt{Pacific/Fakaofo}.
\item \emph{Construction:} Localize a late-December 2025 starting time, convert through UTC, add a duration selected from $\{60,70,90,120,200\}$ minutes, and convert to the destination zone.
\item \emph{Output:} Canonical timestamp in the destination time zone.
\end{itemize}

\item \textbf{\texttt{DST\_SKIP}}
\begin{itemize}[nosep]
\item \emph{Inputs:} Time zones include \texttt{Europe/London}, \texttt{America/New\_York}, \texttt{Australia/Sydney}, and \texttt{Australia/Lord\_Howe}; duration increments are sampled from $\{30,60,90\}$ minutes.
\item \emph{Construction:} Apply the specified handling policy for nonexistent local times during spring-forward transitions, then perform the requested time arithmetic.
\item \emph{Output:} Canonical timestamp with explicit UTC offset.
\end{itemize}

\item \textbf{\texttt{DST\_REPEAT}}
\begin{itemize}[nosep]
\item \emph{Inputs:} DST-observing time zones, including \texttt{Europe/London}, \texttt{America/New\_York}, and \texttt{Australia/Lord\_Howe}; occurrence policy specifies the first or second occurrence of a repeated local time.
\item \emph{Construction:} Resolve the local time according to the specified occurrence and perform the requested time arithmetic.
\item \emph{Output:} Canonical timestamp corresponding to the selected occurrence.
\end{itemize}

\item \textbf{\texttt{DST\_AMBIG}}
\begin{itemize}[nosep]
\item \emph{Inputs:} Ambiguous local timestamps in DST-observing regions; policy specifies returning an ambiguity pair or abstaining.
\item \emph{Construction:} Apply the stated ambiguity policy to the localized candidate instants.
\item \emph{Output:} \texttt{A||B} or \texttt{ABSTAIN}. The stored synthetic gold does not always contain two distinct endpoints: 964 of the 1,134 \texttt{DST\_AMBIG} items in the 10k split have the form \texttt{A||A}.
\end{itemize}

\item \textbf{\texttt{BUSINESS\_DAY}}
\begin{itemize}[nosep]
\item \emph{Inputs:} Time zones include \texttt{America/New\_York}, \texttt{Europe/London}, and \texttt{Asia/Tokyo}; business-day increment $k\in[5,15]$.
\item \emph{Construction:} Advance to the next weekday, then add $k-1$ business days, excluding weekends. Set the resulting local time to 09:00.
\item \emph{Output:} Canonical timestamp on the target business day.
\end{itemize}

\item \textbf{\texttt{CONSTRAINTS}}
\begin{itemize}[nosep]
\item \emph{Inputs:} Time zones include \texttt{Europe/London}, \texttt{America/New\_York}, \texttt{Asia/Tokyo}, and \texttt{Australia/Sydney}.
\item \emph{Construction:} Identify the first weekday after the 15th of the following month and localize it at 09:00.
\item \emph{Output:} Canonical timestamp satisfying the date constraint.
\end{itemize}

\item \textbf{\texttt{RELATIVE\_DAY}}
\begin{itemize}[nosep]
\item \emph{Inputs:} Anchor dates and expressions such as \emph{next Monday}, \emph{this Monday}, \emph{the Monday after next}, and \emph{two Fridays from now}.
\item \emph{Construction:} Resolve the relative weekday expression against the provided anchor and set the local time to 09:00.
\item \emph{Output:} Canonical timestamp on the resolved date.
\end{itemize}

\item \textbf{\texttt{INTERVAL}}
\begin{itemize}[nosep]
\item \emph{Inputs:} Time zones include \texttt{America/New\_York} and \texttt{Europe/London}; durations are sampled from $\{90,135,165,195,240\}$ minutes.
\item \emph{Construction:} Localize the starting instant, calculate the ending instant, and represent the half-open interval $[A,B)$.
\item \emph{Output:} Synthetic JSONL gold stores the interval as \texttt{A||B}, rather than the documented contract form \texttt{A..B}. This applies to all 334 interval items in the 10k split and all 34 in the API 1k subset. Before EM scoring, the evaluator converts \texttt{..} to \texttt{||}, making the two serializations equivalent for exact-match comparison.
\end{itemize}

\item \textbf{\texttt{ISO\_WEEK}}
\begin{itemize}[nosep]
\item \emph{Inputs:} Selected boundary dates, including \texttt{2018-12-31}, \texttt{2020-12-31}, \texttt{2025-01-01}, and \texttt{2015-01-01}.
\item \emph{Construction:} Calculate the ISO week-year and week number using the calendar's ISO week convention.
\item \emph{Output:} \texttt{YYYY-Www} in the synthetic evaluation splits. HumanTime-500 instead stores date-form answers for this task type.
\end{itemize}

\end{enumerate}

\paragraph{Additional Week-Start Task.}

\texttt{WEEKSTART} evaluates the first date of a week under a specified convention, such as a Sunday-based week. Its answer is represented as \texttt{YYYY-MM-DD}. This task is present in HumanTime-500 (25 items) and the API synthetic 1k subset (24 items), but absent from the full synthetic 10k evaluation split. Consequently, not all evaluation splits contain the same set of task types.

\noindent\textit{Utilities.}
The generation pipeline uses time-zone localization, ISO-formatted timestamps, weekday arithmetic, and ISO week calculations. A seeded orchestrator invokes category-specific generators and writes JSONL records with task identifiers, input text, and reference answers.

\subsection{Type $\times$ Year Coverage}
\label{app:ds:heatmaps}

\noindent\textbf{Purpose.}
These heatmaps illustrate the distribution of generated examples across civil-time task types and calendar years. The principal generation period is 2024--2026, while selected ISO week-boundary examples use dates outside that range.

\begin{figure}[t]
  \centering
  \includegraphics[width=\linewidth]{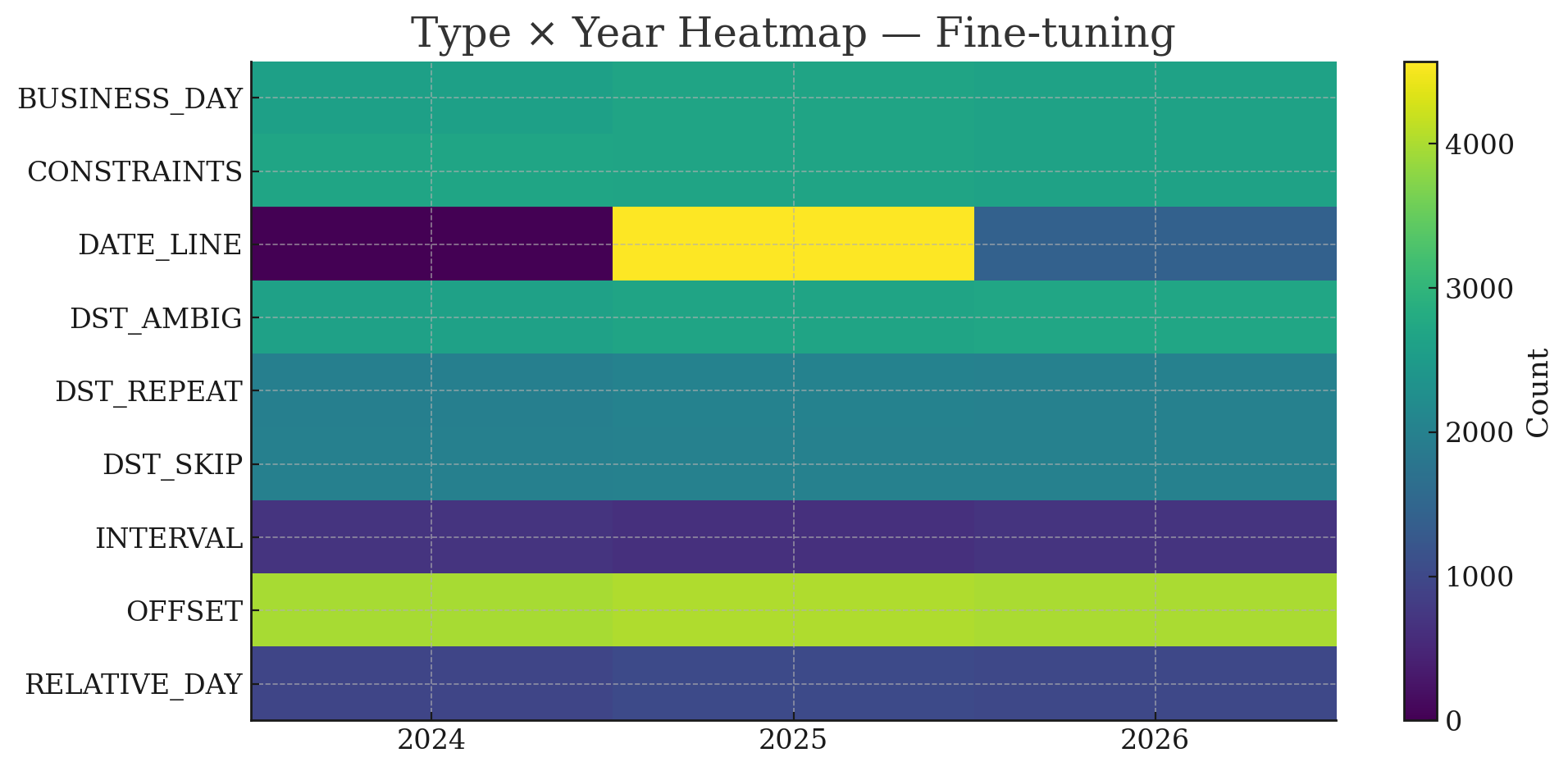}
  \caption{\textbf{Type $\times$ year coverage in the training split.}
  Distribution of generated examples across civil-time task types and calendar years in the 60,011-item training set. The visualization illustrates coverage of DST transitions, time-zone arithmetic, and International Date Line tasks.}
  \label{fig:heatmap-train}
\end{figure}

\begin{figure}[t]
  \centering
  \includegraphics[width=\linewidth]{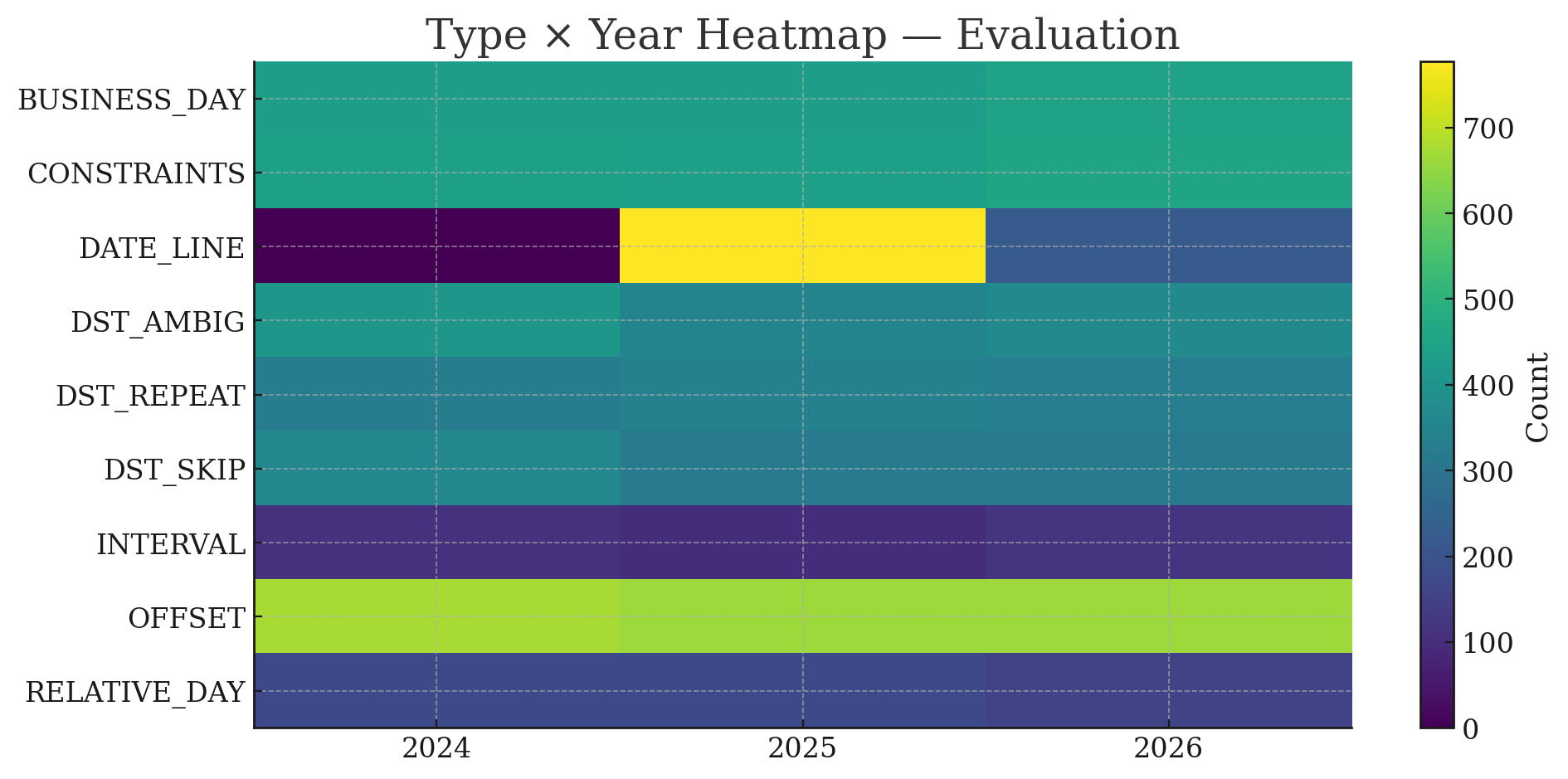}
  \caption{\textbf{Type $\times$ year coverage in the synthetic evaluation split.}
  Distribution of examples across task types and calendar years in the 10,000-item evaluation set. Training and evaluation use the same generators but contain disjoint examples.}
  \label{fig:heatmap-eval}
\end{figure}

\paragraph{Reproducibility.}
The underlying task-type and year-count distributions can be recovered from the released JSONL splits. The shared generation process supports controlled evaluation but should not be interpreted as evidence of generalization to independently collected naturalistic data.

\subsection{Task Type Distribution: Training vs.\ Evaluation}
\label{app:ds:type-counts}

\noindent\textbf{Purpose.}
Absolute task counts characterize the distribution of examples available for training and evaluation, including differences in representation across civil-time categories.

\begin{figure}[t]
  \centering
  \includegraphics[width=\linewidth]{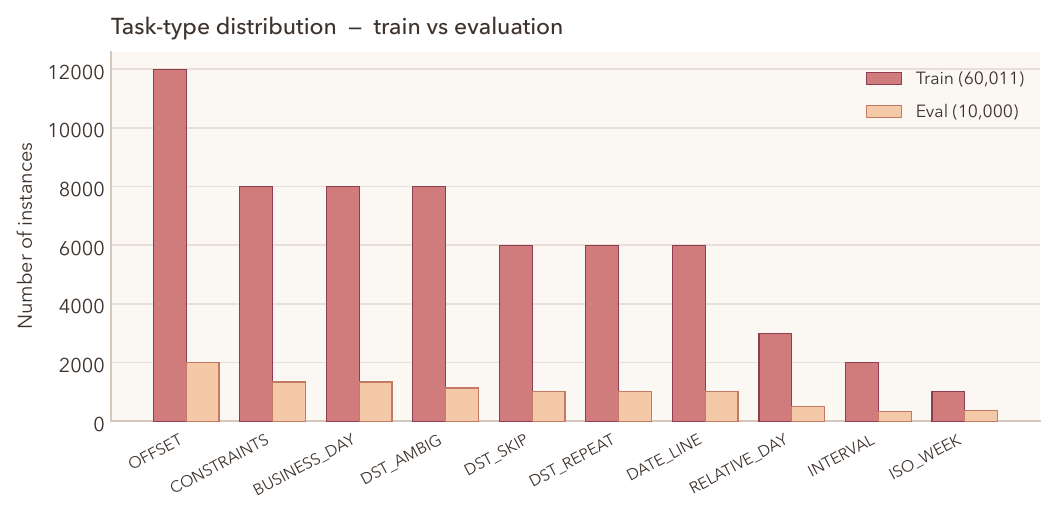}
  \caption{\textbf{Task-type distribution in training and synthetic evaluation.}
  Absolute instance counts per type for the 60,011-item training split and 10,000-item synthetic evaluation split. The figure illustrates the relative availability of examples across calendrical stressors.}
  \label{fig:type-counts}
\end{figure}

\subsection{Normalized Type Composition}
\label{app:ds:type-normalized}

\noindent\textbf{Why normalize?}
Comparing per-type proportions rather than absolute counts makes the composition of the training and evaluation splits easier to assess despite their different sizes.

\begin{figure}[t]
  \centering
  \includegraphics[width=\linewidth]{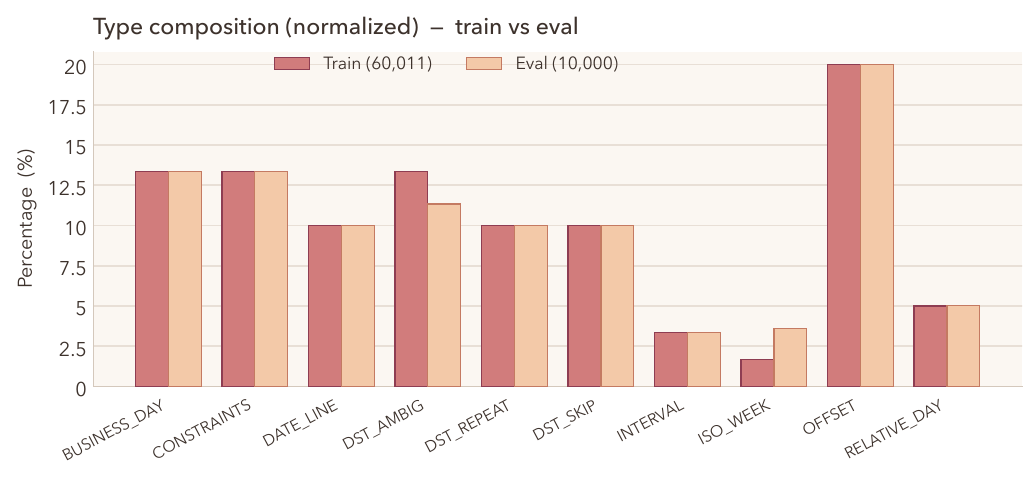}
  \caption{\textbf{Normalized task-type composition.}
  Percentage of examples assigned to each task type in the training and synthetic evaluation splits. The comparison illustrates the similarity of their category distributions, consistent with their shared generation process.}
  \label{fig:type-normalized}
\end{figure}

\subsection{Duration Distribution in \texttt{OFFSET}}
\label{app:ds:offset-duration}

\noindent\textbf{Design.}
The \texttt{OFFSET} generator uses duration increments drawn from $\{15,30,45,60,90,120,180\}$ minutes. These increments test time arithmetic involving both whole-hour and fractional-hour durations.

\begin{figure}[t]
  \centering
  \includegraphics[width=\linewidth]{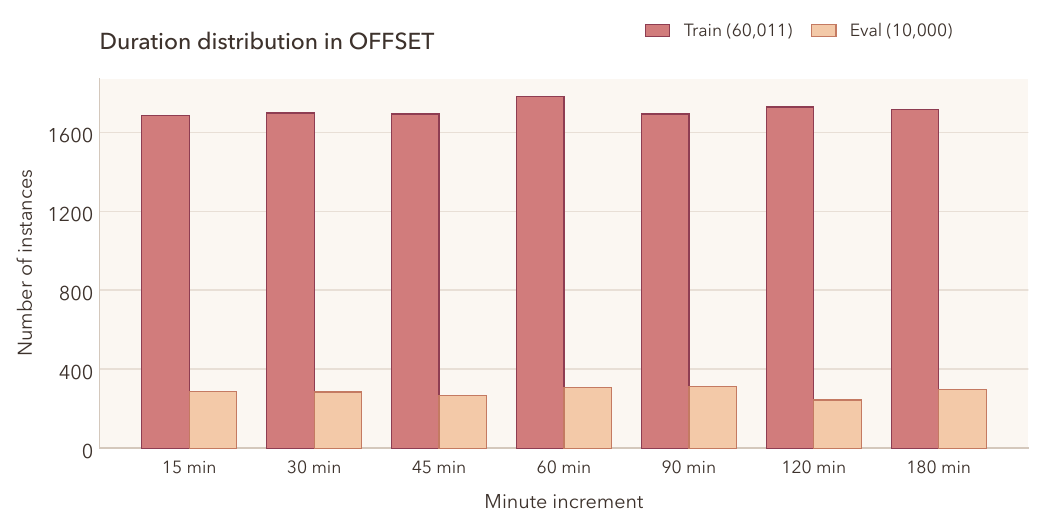}
  \caption{\textbf{Duration distribution in \texttt{OFFSET}.}
  Distribution of selected minute increments in the training and synthetic evaluation splits. The duration buckets include 30- and 45-minute increments, complementing the benchmark's coverage of time zones with non-hour UTC offsets.}
  \label{fig:offset-duration}
\end{figure}

\subsection{Randomness Diagnostics}
\label{app:ds:randomness}

\noindent\textbf{Goal.}
We inspect the generated dataset order and split composition for potential sequencing artifacts. Such checks complement the task-type and year-distribution analyses but do not establish statistical independence between training and evaluation distributions.

\begin{figure}[t]
  \centering
  \includegraphics[width=0.92\linewidth]{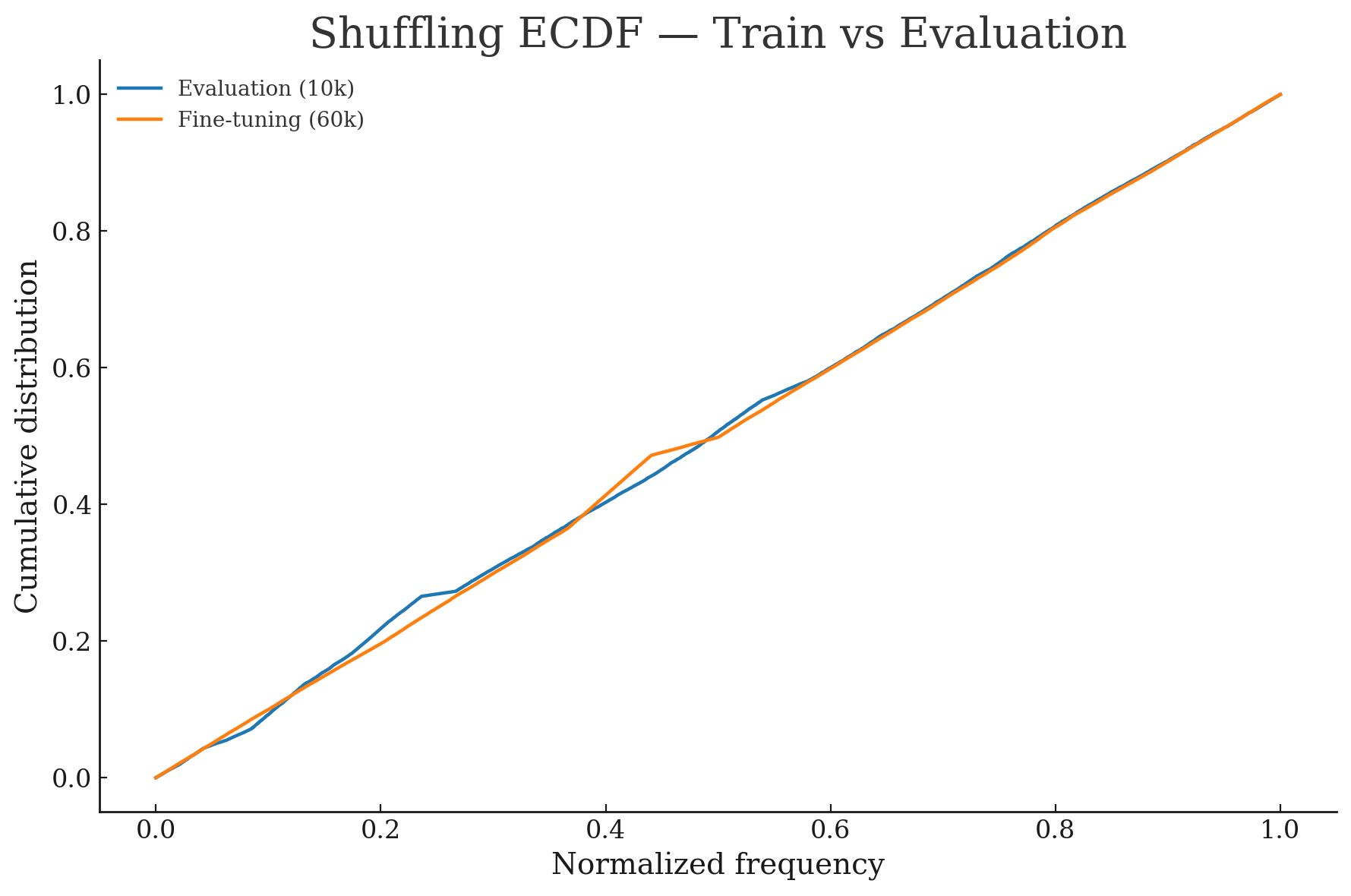}
  \caption{\textbf{Shuffling diagnostic: training vs.\ evaluation.}
  Empirical cumulative distributions of shuffled-index surrogates for the two splits. This diagnostic characterizes the generated ordering but does not independently establish the absence of all sampling biases or template overlap.}
  \label{fig:ecdf-combined}
\end{figure}

\paragraph{Additional checks.}
Adjacent-type agreement and year-distribution checks provide supplementary diagnostics of the generation process. Since synthetic training and evaluation examples share generators, these diagnostics should be interpreted as checks of the controlled sampling setup rather than independent evidence of naturalistic generalization.

\subsection{Evaluation and Labeling Protocol}
\label{app:ds:protocol}

\noindent\textbf{Anchor time.}
Prompts requiring an anchor specify a date, time, and relevant time zone or offset, rather than relying on an unspecified current time. For example, a prompt may state: \emph{``Assume the current instant is 2025-09-22 12:00:00 in Europe/London (BST/UTC+1).''}

\noindent\textbf{Canonical output formats.}
The five canonical answer forms are defined in Section~\ref{sec:output-contract}. They cover abstention, dates and ISO week labels, timestamps with offsets, ambiguity pairs, and half-open intervals. Some stored gold labels differ in serialization between splits, as described above; these differences are handled by the corrected evaluator.

\noindent\textbf{Raw-output and extracted-answer evaluation.}
The corrected evaluation distinguishes three quantities:

\begin{itemize}[nosep, leftmargin=1.3em]
\item \textbf{ContractOK:} Compliance of the raw completion with the output contract, without extracting an answer from surrounding prose. The v2 evaluator accepts either a canonical one-line answer or a JSON object containing a canonical answer value.
\item \textbf{FormatOK:} Validity of the extracted and normalized prediction against the accepted answer forms.
\item \textbf{Exact Match (EM):} Strict equality between the normalized prediction and the normalized gold representation.
\end{itemize}

Answer extraction and normalization are deterministic and label-agnostic. Normalization resolves permitted surface-form and serialization differences without performing calendrical reasoning. ContractOK is evaluated separately in the v2 protocol; the corrected results in Tables~\ref{tab:zeroshot_results}, \ref{tab:icl_results}, and~\ref{tab:finetune_results} report EM and FormatOK.

\noindent\textbf{Interval serialization.}
The documented interval contract uses \texttt{A..B}. However, synthetic interval gold labels are stored as \texttt{A||B}, whereas HumanTime-500 interval labels use \texttt{A..B}. Before exact-match comparison, the corrected evaluator converts occurrences of \texttt{..} to \texttt{||}. Therefore, these two serializations receive the same EM outcome when their endpoints match. This scoring equivalence does not change the documented interval contract or the separate raw-output compliance check.

\noindent\textbf{Deterministic decoding.}
Models are evaluated with temperature $T=0$ and deterministic decoding settings where supported. We report a single run per model and condition, rather than averaging across multiple sampling seeds. LLaMA and Phi-4 few-shot conditions were rerun with distinct 2-, 4-, and 8-shot prompts following the correction of a prompt-construction error.

\noindent\textbf{Evaluation coverage.}
Open-source models are evaluated on the full 10,000-item synthetic evaluation split, whereas the API models use a 1,000-item synthetic subset. HumanTime-500 contains 500 evaluation items.

For every condition, EM is calculated against the full corresponding gold split. Missing predictions contribute zero to the exact-match numerator and are therefore counted as incorrect. Incomplete prediction coverage is reported for affected runs.

\noindent\textbf{Abstention.}
Abstention is correct only when the reference answer is \texttt{ABSTAIN}. The evaluator distinguishes whether an abstention is formatted correctly from whether it is the correct response under the specified task policy.

\noindent\textbf{Ground truth and time-zone rules.}
Synthetic reference answers are generated using the specified time-zone and calendrical libraries, including \texttt{pytz} and IANA time-zone rules. The evaluation depends on these rules and the reference-generation implementation; it does not assume that different external libraries necessarily produce identical outputs under all historical or policy-dependent conditions.

\noindent\textbf{Temporal-error diagnostics.}
Median Temporal Error (MedTE) is reported in seconds only for predictions for which temporal error is defined, together with temporal-error coverage (TEcov). It is diagnostic and does not modify EM. In particular, a zero median does not imply that every evaluation item was answered correctly.

\noindent\textbf{Statistical testing and uncertainty.}
We use paired McNemar tests to compare corrected per-example EM outcomes. Tests use the continuity-corrected chi-squared statistic, with a two-sided exact binomial test when the number of discordant pairs is at most 50. Reported significance thresholds are unadjusted.

We compute 95\% Wilson score confidence intervals for overall and per-category EM. These intervals quantify evaluation-item uncertainty under the single-run setup and should not be interpreted as variation across repeated model runs.

For $k$ correct predictions among $n$ evaluation items, let $\widehat{p}=k/n$ and $z=1.96$. The Wilson interval is

\[
\frac{
\widehat{p}+\frac{z^2}{2n}
\pm
z\sqrt{
\frac{\widehat{p}(1-\widehat{p})}{n}
+\frac{z^2}{4n^2}
}
}{
1+\frac{z^2}{n}
}.
\]

For incomplete runs, the denominator remains the full number of gold items.

\paragraph{Per-Category Confidence Intervals.}

\begin{figure*}[t]
\centering
\includegraphics[width=0.96\textwidth]{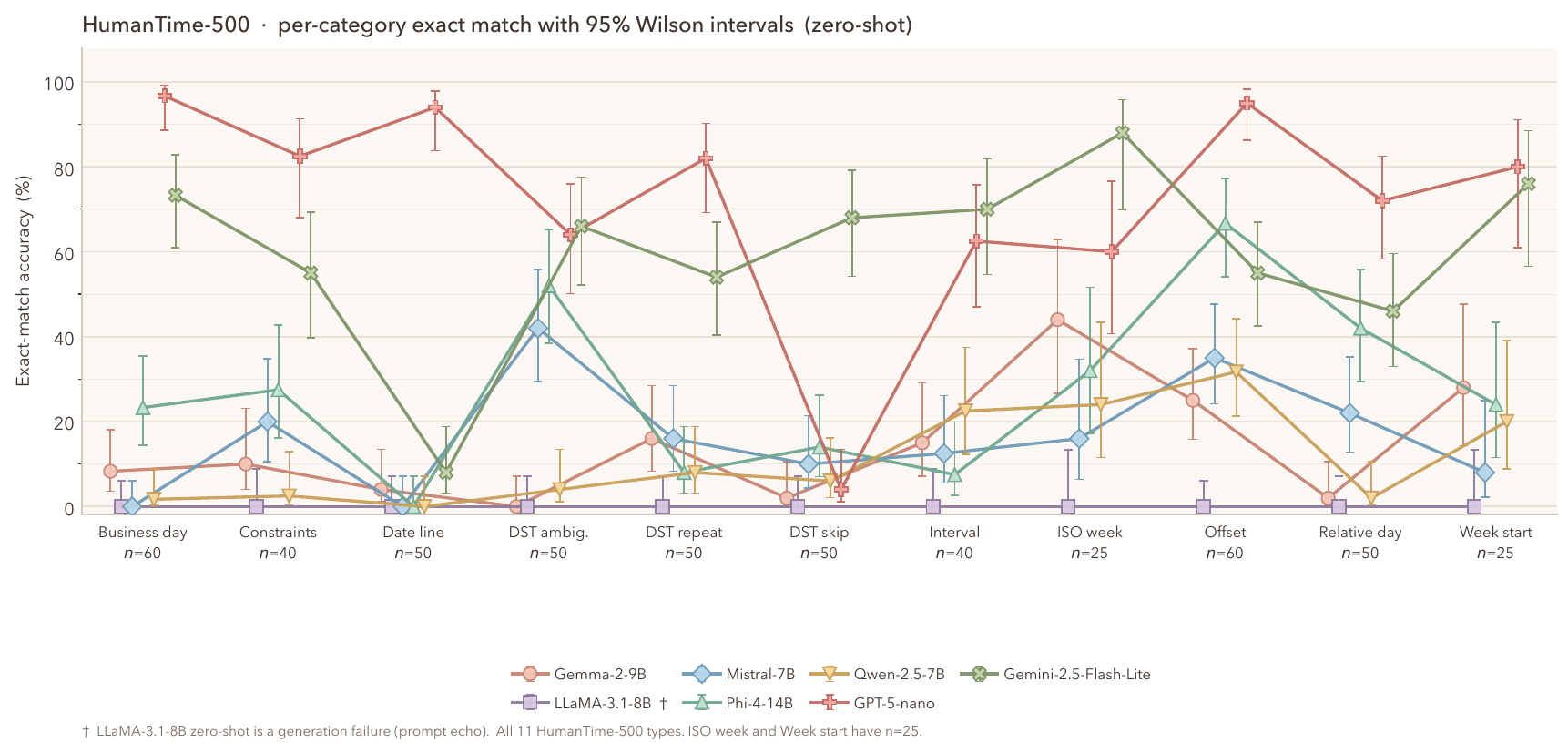}
\caption{\textbf{Per-category zero-shot exact-match accuracy with 95\% Wilson confidence intervals on HumanTime-500.}
Results for all seven evaluated models across the 11 civil-time task types. Points show per-category EM, and error bars show Wilson score intervals computed from the number of correct predictions and the number of evaluation items in each category ($n=25$--$60$). The smaller \textsc{ISO\_WEEK} and \textsc{WEEKSTART} categories each contain 25 items. LLaMA-3.1-8B's zero-shot results reflect a prompt-echo generation failure and should not be interpreted as an isolated measure of calendrical reasoning ability.}
\label{fig:humantime_wilson_zeroshot}
\end{figure*}

Figure~\ref{fig:humantime_wilson_zeroshot} presents per-category zero-shot EM with 95\% Wilson confidence intervals for all seven evaluated models on HumanTime-500.

Because HumanTime-500 contains relatively few examples per category, particularly for \textsc{ISO\_WEEK} and \textsc{WEEKSTART}, uncertainty should be considered when comparing task-specific performance.

\subsection{Output Contract Reminder}
\label{app:ds:contract}

The benchmark defines five canonical answer forms:

\begin{enumerate}[nosep, leftmargin=1.3em]
\item \texttt{ABSTAIN} for cases requiring abstention under the stated policy.
\item \texttt{YYYY-MM-DD} or \texttt{YYYY-Www} for dates and ISO week labels.
\item ISO-8601 timestamps with explicit UTC offsets or \texttt{Z}.
\item \texttt{A||B} for ambiguity pairs.
\item \texttt{A..B} for half-open intervals $[A,B)$.
\end{enumerate}

These forms describe the intended answer contract. The released evaluation splits contain some differing gold serializations, notably for interval and ISO-week tasks. The corrected EM evaluator accounts for the interval delimiter difference, while raw-output compliance and extracted-answer validity remain distinct measurements.

\subsection{Pseudocode: Evaluation and Scoring}
\label{app:a9:pseudocode}

\noindent\textbf{Setup.}
The corrected evaluation pipeline separates raw-output compliance, deterministic answer extraction and normalization, exact-match correctness, and temporal-error diagnostics. The pseudocode below summarizes this evaluation protocol. Unlike the earlier tolerance-based scoring description, EM is binary and does not award partial credit.

\FloatBarrier

\begin{algorithm}[H]
\caption{Civil-Time Evaluation Protocol}
\label{alg:tz-eval}
\small
\begin{algorithmic}[1]
\Require Evaluation items $\mathcal{D}$ with gold labels; predictions $P$
\State $N \gets |\mathcal{D}|$
\State Initialize per-item records $R \gets \emptyset$
\ForAll{$x \in \mathcal{D}$}
  \State $r \gets P[x.\texttt{id}]$ if present, else $\bot$
  \If{$r=\bot$}
    \State $c \gets 0$; $f \gets 0$; $e \gets 0$; $d \gets \bot$
  \Else
    \State $(c,f,e,d) \gets \textsc{ScoreItem}(r,x.\texttt{gold},x.\texttt{type})$
  \EndIf
  \State Append $(x.\texttt{id},x.\texttt{type},c,f,e,d)$ to $R$
\EndFor
\State $\mathrm{EM} \gets \frac{1}{N}\sum_{i=1}^{N}R_i.e$
\State $\mathrm{FormatOK} \gets \frac{1}{N}\sum_{i=1}^{N}R_i.f$
\State $\mathrm{ContractOK} \gets \frac{1}{N}\sum_{i=1}^{N}R_i.c$
\ForAll{task types $t$}
  \State $R_t \gets \{q\in R:q.\texttt{type}=t\}$
  \State $n_t \gets |R_t|$; $k_t \gets \sum_{q\in R_t}q.e$
  \State $\mathrm{EM}_t \gets k_t/n_t$
  \State $\mathrm{CI}_t \gets \textsc{Wilson95}(k_t,n_t)$
\EndFor
\State $\mathcal{E} \gets \{q.d:q\in R,\ q.d\neq\bot\}$
\State $\mathrm{TEcov} \gets |\mathcal{E}|/N$
\State $\mathrm{MedTE} \gets \textsc{Median}(\mathcal{E})$
       \Comment{If defined}
\State \Return overall metrics, per-type metrics, CIs, coverage
\end{algorithmic}
\end{algorithm}

\begin{algorithm}[H]
\caption{Answer Normalization and Item Scoring}
\label{alg:tz-grade}
\small
\begin{algorithmic}[1]

\Function{ExtractNorm}{$r$}
  \State Extract answer deterministically from raw completion $r$
  \State Normalize permitted surface-form differences
  \State Preserve the represented calendrical answer
  \State \Return normalized answer string, or $\bot$
\EndFunction

\Function{IsAcceptedFormat}{$s$}
  \State \Return whether $s$ matches any accepted answer form
  \Statex \hspace{\algorithmicindent} (independent of task type)
\EndFunction

\Function{CanonicalizeValue}{$s$}
  \State Normalize supported timestamp, seconds, and offset forms
  \State Replace \texttt{..} with \texttt{||} for EM comparison
  \State Collapse identical endpoints \texttt{A||A} to \texttt{A}
  \State \Return canonicalized value
\EndFunction

\Function{RawContractOK}{$r$}
  \If{$r$ is exactly a canonical one-line answer}
    \State \Return $1$
  \ElsIf{$r$ is a JSON object containing a canonical answer value}
    \State \Return $1$
  \Else
    \State \Return $0$
  \EndIf
\EndFunction

\Function{ScoreItem}{$r,g,t$}
  \State $c \gets \textsc{RawContractOK}(r)$
  \State $\hat{y} \gets \textsc{ExtractNorm}(r)$
  \State $g' \gets \textsc{NormalizeGold}(g)$
  \State $f \gets \mathbb{1}[\hat{y}\neq\bot \land
              \textsc{IsAcceptedFormat}(\hat{y})]$
  \If{$\hat{y}=\bot$}
    \State $e \gets 0$; $d \gets \bot$
  \Else
    \State $u \gets \textsc{CanonicalizeValue}(\hat{y})$
    \State $v \gets \textsc{CanonicalizeValue}(g')$
    \State $e \gets \mathbb{1}[u=v]$
         \Comment{Strict EM; no partial credit}
    \State $d \gets \textsc{TemporalErrorIfDefined}(\hat{y},g',t)$
         \Comment{Diagnostic only}
  \EndIf
  \State \Return $(c,f,e,d)$
\EndFunction

\end{algorithmic}
\end{algorithm}

\noindent\textbf{Interpretation.}
Exact Match requires equality after deterministic, label-agnostic extraction and full value canonicalization, including supported timestamp normalization, interval delimiter conversion (\texttt{A..B} to \texttt{A||B}), and identical-endpoint collapse (\texttt{A||A} to \texttt{A}). FormatOK checks the extracted answer against the accepted forms independently of task type. Temporal-error calculations remain separate diagnostics and introduce no tolerance or partial credit into EM. Overall and per-type statistics use the full evaluation denominator, including missing predictions.

\section{Evaluation Details}
\label{app:norm}

\paragraph{Accepted output contract.}

The benchmark defines five canonical answer forms:

\begin{enumerate}[leftmargin=1.3em, itemsep=2pt]
  \item \texttt{ABSTAIN}.
  \item \texttt{YYYY-MM-DD} or \texttt{YYYY-Www} for date and ISO week labels.
  \item A single ISO-8601 instant with seconds and an explicit UTC offset (or \texttt{Z}):\\
  \texttt{YYYY-MM-DDThh:mm:ss}\allowbreak$\pm$\allowbreak\texttt{hh:mm}
  \ \ or \ \
  \texttt{YYYY-MM-DDThh:mm:ssZ}.
  \item \texttt{A||B} for civil-time ambiguity, representing an ambiguity pair.
  \item \texttt{A..B} for intervals, interpreted as half-open $[A,B)$ with the end excluded.
\end{enumerate}

The documented contract reserves \texttt{||} for ambiguity and \texttt{..} for intervals. However, the released evaluation splits contain some serialization differences. All 334 \texttt{INTERVAL} gold labels in the synthetic 10k split and all 34 in the API 1k subset use \texttt{A||B}, whereas all 40 HumanTime-500 interval labels use \texttt{A..B}. The corrected EM evaluator treats these delimiters as equivalent by converting \texttt{..} to \texttt{||} before comparison.

Similarly, synthetic \texttt{ISO\_WEEK} labels use \texttt{YYYY-Www}, whereas HumanTime-500 uses \texttt{YYYY-MM-DD}. Some synthetic \texttt{DST\_AMBIG} gold labels also contain identical endpoints (\texttt{A||A}); consequently, the stored representation should not always be interpreted as containing two distinct instants. The evaluator canonicalizes repeated endpoints to a single value for EM comparison.

These distinctions concern stored gold representations and EM normalization; they do not change the documented output contract.

\paragraph{Two-stage protocol and normalizer design.}

We evaluate predictions using the following pipeline:

\[
\begin{gathered}
\text{Raw completion}
\longrightarrow
\text{Extraction/normalization}\\
\longrightarrow
\text{Validation and scoring}
\longrightarrow
\text{Aggregation}.
\end{gathered}
\]

\textbf{Stage~1} records the raw model completion $r_i=f(x_i)$, generated using deterministic decoding settings where supported (temperature $T=0$).

\textbf{Stage~2} applies a deterministic normalizer $\hat{y}_i=g(r_i,t_i)$ that extracts and standardizes the intended answer according to task type $t_i$.

The normalizer is \textbf{label-agnostic}: it does not consult the gold answer $y_i$ when selecting or normalizing a prediction. It is also \textbf{non-reasoning}: it does not perform time-zone conversion, calendar arithmetic, or other operations intended to solve the task.

Normalization addresses surface-form differences, including:

\begin{enumerate}[leftmargin=1.3em, itemsep=2pt]
\item Completing omitted seconds where supported (\texttt{HH:MM} $\rightarrow$ \texttt{HH:MM:00}).
\item Standardizing UTC offsets (e.g., \texttt{+0330} $\rightarrow$ \texttt{+03:30}).
\item Normalizing equivalent UTC designators, such as \texttt{Z} and \texttt{+00:00}.
\item Removing surrounding whitespace, code fences, or prose during answer extraction.
\item Selecting an answer candidate using deterministic extraction rules when the completion contains multiple candidate expressions.
\end{enumerate}

These operations standardize the representation of an extracted answer without establishing its calendrical correctness.

If no valid answer candidate can be extracted, the normalized prediction is marked invalid. Such failures are not automatically converted to \texttt{ABSTAIN}.

\paragraph{Raw-output compliance and extracted-answer validity.}

The evaluation distinguishes \textbf{ContractOK}, \textbf{FormatOK}, and \textbf{Exact Match (EM)}.

\textbf{ContractOK} evaluates the raw completion without extracting an answer from surrounding text. Under the v2 protocol, a raw response is compliant if it is either a canonical one-line answer or a JSON object containing a canonical answer value.

\textbf{FormatOK} evaluates whether the extracted and normalized prediction satisfies one of the accepted answer forms. This validation is based on the prediction's syntax, independently of the task type. Task type may guide answer extraction, but it is not used to impose a type-specific FormatOK requirement.

\textbf{EM} evaluates whether the normalized prediction exactly matches the normalized reference answer. No temporal tolerance or partial credit is used.

ContractOK is measured separately using the v2 evaluator. The corrected results in Tables~\ref{tab:zeroshot_results}, \ref{tab:icl_results}, and~\ref{tab:finetune_results} report EM and FormatOK; they should not be interpreted as raw-output ContractOK measurements.

\paragraph{Contract compliance and exact match.}

Let $N$ be the number of gold evaluation items, $\hat{y}_i$ the extracted and normalized prediction, and $y_i$ the reference answer.

Let $\mathrm{OK}(\hat{y}_i)$ indicate whether the normalized prediction satisfies one of the five accepted output forms, without conditioning on task type.

We define:

\[
\begin{aligned}
\mathrm{FormatOK}
&=
\frac{1}{N}
\sum_{i=1}^{N}
\mathbf{1}\{\mathrm{OK}(\hat{y}_i)\},\\
\mathrm{EM}
&=
\frac{1}{N}
\sum_{i=1}^{N}
\mathbf{1}\{
\hat{y}_i\neq\emptyset
\land h(\hat{y}_i)=h(y_i)
\},
\end{aligned}
\]

where $h(\cdot)$ denotes the deterministic \texttt{canonicalize\_value} operation used for exact-match comparison.

This canonicalization includes:

\begin{enumerate}[leftmargin=1.3em, itemsep=2pt]
\item Standardizing supported timestamp representations, including omitted seconds and UTC offset formats.
\item Normalizing equivalent UTC representations, including \texttt{Z} and \texttt{+00:00}.
\item Converting interval delimiters from \texttt{..} to \texttt{||}.
\item Collapsing identical paired values, \texttt{A||A} $\rightarrow$ \texttt{A}.
\end{enumerate}

The same canonicalization is applied to predictions and gold labels. In particular, an identical-endpoint ambiguity label can match its corresponding single-instant representation after canonicalization.

The EM comparison remains strict after normalization; it does not perform calendrical computation or award credit for temporally close but unequal answers.

Let $\mathcal{T}$ denote the set of task types and $N_t$ the number of evaluation items belonging to type $t$. Per-type and macro-averaged EM are:

\[
\begin{aligned}
\mathrm{EM}_t
&=
\frac{1}{N_t}
\sum_{i:t_i=t}
\mathbf{1}\{
\hat{y}_i\neq\emptyset
\land h(\hat{y}_i)=h(y_i)
\},\\
\mathrm{MacroEM}
&=
\frac{1}{|\mathcal{T}|}
\sum_{t\in\mathcal{T}}
\mathrm{EM}_t.
\end{aligned}
\]

Overall EM is instance-weighted, whereas MacroEM gives equal weight to each task type. This distinction is relevant to Figure~\ref{fig:strategy_macro_em}, which reports macro-averaged results.

All metrics use the full gold evaluation denominator. Missing predictions are counted as incorrect rather than being excluded from the calculation.

\paragraph{Robust abstention metrics.}

A single micro-averaged abstention metric can be misleading when abstention-required items are rare. We therefore distinguish correct abstention on items requiring it from unnecessary abstention on items requiring a substantive answer.





Let
\[
\begin{aligned}
A &= \{i : h(y_i)=\mathrm{ABSTAIN}\},\\
A^c &= \{i : h(y_i)\neq\mathrm{ABSTAIN}\}.
\end{aligned}
\]

For brevity, define
$\hat{a}_i=h(\hat{y}_i)$.

We define Abstention-Required Accuracy
(ARA) and Over-Abstention Rate (OAR) as

\[
\begin{aligned}
\mathrm{ARA}
&= \frac{1}{|A|}
\sum_{i\in A}
\mathbb{1}[\hat{a}_i=\mathrm{ABSTAIN}],\\[4pt]
\mathrm{OAR}
&= \frac{1}{|A^c|}
\sum_{i\in A^c}
\mathbb{1}[\hat{a}_i=\mathrm{ABSTAIN}].
\end{aligned}
\]

Higher ARA indicates better identification of abstention-required cases, while lower OAR indicates fewer unnecessary abstentions.

We also report Abstain-F1, treating an explicit abstention as the positive prediction. With $TP$, $FP$, and $FN$ denoting correct abstentions, unnecessary abstentions, and missed required abstentions, respectively,

\[
\mathrm{Abstain\mbox{-}F1}
=
\frac{2TP}{2TP+FP+FN},
\]

when the denominator is nonzero.

These quantities are defined only when their respective denominators are nonzero. An invalid prediction is not treated as an abstention unless the model explicitly produces the reserved \texttt{ABSTAIN} answer.

\paragraph{Temporal error (seconds) and coverage.}
\label{app:temporal_error}

To characterize temporal near-misses without replacing EM, we compute temporal error for predictions and reference answers whose temporal representations can be parsed and compared.

Let $\tau(\cdot)$ map an ISO-8601 instant with an explicit UTC offset (or \texttt{Z}) to Unix seconds.

For a single-instant reference answer $y_i$, define

\[
\mathcal{Y}_i=\{\tau(y_i)\}.
\]

For an ambiguity reference $y_i=\texttt{A||B}$, define

\[
\mathcal{Y}_i=
\{\tau(\texttt{A}),\tau(\texttt{B})\}.
\]

If the prediction can be parsed as an eligible single instant $\hat{\tau}_i$, its temporal error is the distance to the nearest valid reference instant:

\[
\mathrm{TE}_i
=
\min_{\tau^\star\in\mathcal{Y}_i}
\left|
\hat{\tau}_i-\tau^\star
\right|.
\]

When canonicalization collapses an identical-endpoint reference \texttt{A||A} to \texttt{A}, the reference is treated as a single instant for temporal-error calculation.

For an interval reference $y_i=\texttt{A..B}$ and prediction $\hat{y}_i=\widehat{\texttt{A}}\texttt{..}\widehat{\texttt{B}}$, or their equivalent \texttt{||}-delimited representations, temporal error is defined when both endpoints of each interval can be parsed.

Let $g_s,g_e$ denote the reference start and end instants in Unix seconds, and $p_s,p_e$ the corresponding predicted endpoints. We define

\[
\mathrm{TE}_i
=
\max
\left(
|p_s-g_s|,
|p_e-g_e|
\right).
\]

Thus, interval temporal error measures the larger absolute endpoint deviation, rather than excluding interval predictions from temporal-error analysis.

Temporal error remains undefined when the required temporal representations cannot be parsed or compared. This includes invalid predictions, abstentions, and non-instant labels for which no supported temporal-error calculation applies.

We report \textbf{Temporal Error Coverage (TEcov)} as

\[
\mathrm{TEcov}
=
\frac{1}{N}
\sum_{i=1}^{N}
\mathbf{1}\{
\mathrm{TE}_i
\text{ is defined}
\}.
\]

Median Temporal Error (MedTE) is computed over the subset with defined temporal errors:

\[
\mathrm{MedTE}
=
\operatorname{median}
\left(
\{\mathrm{TE}_i:
\mathrm{TE}_i
\text{ is defined}\}
\right).
\]

A reported $\mathrm{MedTE}=0$ seconds therefore does not imply error-free performance over the full evaluation set. It must be interpreted alongside EM and TEcov.

\paragraph{Paraphrase robustness (FlipRate; optional).}

When semantically equivalent paraphrase variants are available, items may be grouped by a shared \texttt{parent\_id}.

Let $\mathcal{G}$ denote the set of paraphrase groups and

\[
c_i=
\mathbf{1}\{
\hat{y}_i\neq\emptyset
\land h(\hat{y}_i)=h(y_i)
\}
\]

indicate strict EM correctness.

A group exhibits a correctness flip when its members do not share the same correctness outcome:

\[
\resizebox{\columnwidth}{!}{$\displaystyle
\mathrm{FlipRate}
=
\frac{1}{|\mathcal{G}|}
\sum_{g\in\mathcal{G}}
\mathbf{1}
\left\{
\exists\, i,j\in g:
c_i\neq c_j
\right\}.
$}
\]

Lower FlipRate indicates greater consistency across semantically equivalent formulations. This is an optional diagnostic and is not included among the primary corrected evaluation metrics.

\paragraph{Statistical uncertainty.}

We compute 95\% Wilson score confidence intervals for overall and per-type EM. These intervals describe uncertainty across evaluation items under the single-run setup and do not quantify variation across independently sampled model generations.

Paired comparisons use McNemar tests computed from corrected per-example EM outcomes, with continuity-corrected chi-squared tests and two-sided exact binomial tests when the number of discordant pairs is at most 50. Reported significance thresholds are unadjusted.

Per-category Wilson intervals for all seven models on HumanTime-500 are visualized in Figure~\ref{fig:humantime_wilson_zeroshot}.

\paragraph{Reproducibility and fairness.}

All models are evaluated using the same scoring definitions and normalization protocol. Open-source models use the full 10,000-item synthetic evaluation split, while API models use a 1,000-item subset. All models are evaluated on HumanTime-500.

Decoding uses temperature $T=0$ and deterministic settings where supported, with a single run per model and condition. Incomplete prediction coverage is explicitly reported, and missing predictions count as incorrect.

The use of a shared evaluator helps distinguish output formatting from answer correctness. However, scoring consistency alone does not establish that all observed performance differences arise exclusively from calendrical reasoning.


\begin{algorithm*}[t]
\small
\caption{Deterministic answer extraction and normalization}
\label{alg:norm}
\begin{algorithmic}[1]

\Require Raw model completion $r$ and task type $t$
\Ensure Normalized prediction $\hat{y}$, or $\emptyset$ if extraction fails

\State $r \gets \textsc{Trim}(r)$
\State $r \gets \textsc{StripFences}(r)$
\State $r \gets \textsc{StandardizeWhitespace}(r)$

\State $C \gets \textsc{ExtractCandidatesInOrder}(r,t)$
\Comment{No gold-label access}

\ForAll{$c\in C$}
    \State $c' \gets \textsc{CanonicalizeSurfaceForm}(c)$
    \Comment{No calendrical computation}
    \If{$\textsc{AcceptedFormat}(c')$}
        \State \Return $c'$
        \Comment{First accepted candidate}
    \EndIf
\EndFor

\State \Return $\emptyset$

\end{algorithmic}
\end{algorithm*}

\paragraph{Exact-match canonicalization.}

After answer extraction, the corrected evaluator applies \texttt{canonicalize\_value} to both the normalized prediction and the reference answer before EM comparison.

For clarity, the essential transformations are summarized as

\[
\begin{aligned}
\texttt{A..B}
&\longrightarrow
\texttt{A||B},\\
\texttt{A||A}
&\longrightarrow
\texttt{A},\\
\text{timestamp variants}
&\longrightarrow
\text{standardized representation}.
\end{aligned}
\]

Timestamp standardization includes supported seconds, UTC offset, and \texttt{Z}/\texttt{+00:00} normalization. These transformations are applied without consulting the correct answer during prediction extraction.

This canonicalization reconciles the different stored interval delimiters and identical-endpoint ambiguity representations without introducing temporal tolerance or partial credit.

Raw-output ContractOK is evaluated separately under the v2 protocol, without answer extraction. FormatOK is determined from the extracted answer's accepted syntax, independently of its task type or agreement with the gold label.

\section{Systems and Protocol Details}
\label{app:systems}

\subsection{Models (Expanded)}
\label{app:models}

We evaluate seven language models: six instruction-tuned models and one base model (Llama-3.1-8B), comprising five open-weight models and two API-based models. All models use the same evaluation and scoring protocol, although the baseline zero-shot condition does not explicitly restate the output contract in its prompt.

\paragraph{Gemma-2-9B-IT~\citep{gemma2}.}
An instruction-tuned 9B-parameter model released by Google. We use the official instruction-tuned checkpoint and tokenizer. Low-precision loading is employed where required by available GPU memory. Decoding uses temperature $T=0$ under the experimental settings described in Section~\ref{sec:experimental-conditions}.

\paragraph{Mistral-7B-Instruct-v0.3~\citep{jiang2023mistral7b}.}
An instruction-tuned 7B-parameter model from Mistral AI. We use the official instruction-tuned weights and tokenizer. Fine-tuning employs QLoRA with 4-bit NF4 quantization and LoRA rank 32, consistent with the other open-weight models.

\paragraph{Qwen-2.5-7B-Instruct~\citep{qwen25}.}
An instruction-tuned 7B-parameter model from the Qwen family. We use the official instruction-tuned checkpoint and tokenizer. The model identity is Qwen-2.5-7B-Instruct throughout the paper, regardless of abbreviated labels used in intermediate evaluation artifacts.

\paragraph{Llama-3.1-8B (Base)~\citep{grattafiori2024llama3herdmodels}.}
An 8B-parameter base language model from Meta. We use the \texttt{meta-llama/Llama-3.1-8B} checkpoint, not the instruction-tuned variant. Prompts are manually wrapped with role-header tokens, including \texttt{<|start\_header\_id|>}, rather than formatted using \texttt{apply\_chat\_template}. In baseline zero-shot evaluation, the model frequently continues or echoes the prompt instead of producing the requested answer, resulting in 0.0\% EM. This behavior is consistent with the use of a base model and should not be interpreted as an isolated measure of calendrical reasoning ability. The corrected 2-, 4-, and 8-shot evaluations use distinct prompts.

\paragraph{Phi-4-14B.}
An instruction-tuned 14B-parameter model from Microsoft. We use the official checkpoint and tokenizer, with low-precision loading where necessary. Its corrected in-context learning evaluations also use distinct 2-, 4-, and 8-shot prompts following the prompt-construction correction.

\paragraph{Gemini-2.5-Flash-Lite~\citep{gemini25}.}
An API-served model from Google. We evaluate it using the same benchmark scoring protocol as the open-weight models, with deterministic generation settings where supported. Requests are subject to provider-specific rate limits and operational constraints.

\paragraph{GPT-5-nano~\citep{openai_gpt5_nano_2026}.}
An API-served model from OpenAI. We use the same task prompts and evaluation protocol as applicable to the other models, with deterministic decoding settings where supported. Incomplete prediction coverage is handled consistently by counting missing predictions as incorrect against the full gold evaluation split.

\subsection{Compute and Budgets}
\label{app:compute}

Open-weight experiments, including zero-shot evaluation, in-context learning, and fine-tuning, are conducted using \textbf{2$\times$ NVIDIA T4 GPUs} in the Kaggle environment, with 8-bit or 4-bit loading where applicable.

Parameter-efficient fine-tuning uses QLoRA with \textbf{4-bit NF4 quantization and LoRA rank 32}. Training uses the 60,011-example SyntheticTimeBench training split with completion-only loss, masking user-prompt tokens. The fine-tuning configuration is held constant across the five open-weight models; additional training details are provided in Appendix~\ref{app:sft-details}.

Prompting evaluations use temperature $T=0$ and a maximum generation length of 128 new tokens where supported. We report a single run per model and condition rather than averaging across multiple random seeds.

API evaluations are subject to provider-specific request limits and operational budgets. Consequently, GPT-5-nano and Gemini-2.5-Flash-Lite are evaluated on a 1,000-item synthetic subset, rather than the full 10,000-item evaluation split used for open-weight models. Both API models are also evaluated on HumanTime-500.

Some evaluation runs have incomplete prediction coverage. The corrected scoring protocol computes EM against the full gold split, treating missing predictions as incorrect rather than restricting the denominator to completed predictions.

Although five open-weight models were fine-tuned, only four have usable matched synthetic evaluation results: Gemma, LLaMA, Mistral, and Phi-4. Gemma's HumanTime-500 fine-tuning dump is invalid and excluded from the reported results. Qwen's synthetic fine-tuning predictions could not be matched to the evaluation gold identifiers; its HumanTime-500 fine-tuning result is reported separately (41.6\% EM; 436/500 predictions).

\subsection{Reporting}
\label{app:reporting}

We distinguish \textbf{Exact Match (EM)}, \textbf{FormatOK}, and \textbf{ContractOK} as defined in Section~\ref{sec:eval-metrics} and Appendix~\ref{app:norm}.

\textbf{EM} is the primary correctness metric, requiring equality between the canonicalized prediction and gold answer. \textbf{FormatOK} measures whether an extracted and normalized prediction satisfies an accepted output form, independently of task type. \textbf{ContractOK}, evaluated separately under the v2 protocol, measures raw-output compliance without answer extraction and accepts either a canonical one-line answer or a JSON object containing a canonical answer value.

The corrected main results in Tables~\ref{tab:zeroshot_results}, \ref{tab:icl_results}, and~\ref{tab:finetune_results} report EM and FormatOK; these should not be interpreted as ContractOK measurements.

We additionally report macro-averaged EM and per-category accuracy where applicable. Macro-EM gives equal weight to each task type, whereas overall EM is weighted by the number of evaluation items. The two aggregations are therefore not interchangeable.

For temporal-error diagnostics, we report Median Temporal Error (MedTE) over cases with defined temporal error, together with temporal-error coverage (TEcov). A zero MedTE does not imply error-free performance over the full evaluation set. Abstention behavior is characterized using Abstention-Required Accuracy (ARA), Over-Abstention Rate (OAR), and Abstain-F1 where applicable. Paraphrase FlipRate is treated as an optional diagnostic.

We report \textbf{95\% Wilson confidence intervals} for overall and per-category EM and use paired McNemar tests for comparisons based on corrected per-example outcomes. McNemar tests use the continuity-corrected chi-squared statistic, with a two-sided exact binomial test when the number of discordant pairs is at most 50. Significance thresholds are unadjusted.

Per-category Wilson intervals for HumanTime-500 are presented in Figure~\ref{fig:humantime_wilson_zeroshot}. These intervals reflect uncertainty across evaluation items under the single-run setup, not variability across repeated model generations.

All reported accuracy metrics use the full corresponding gold denominator, including missing predictions as incorrect. Invalid evaluation artifacts are excluded rather than interpreted as model failures.

Finally, while all models share the same scoring protocol, prompting conditions differ by design: baseline zero-shot uses minimal instructions, contract-aware zero-shot explicitly restates the output contract, ICL adds type-matched exemplars, and fine-tuning adapts model parameters. The use of a base rather than instruction-tuned LLaMA checkpoint is an additional model-specific consideration when interpreting its zero-shot results. Differences in evaluation-set size and prediction coverage are reported to support appropriate interpretation of cross-model comparisons.

\section{Prompt Enhancement}
\label{app:prompt_enhancement}

We evaluate a \textbf{contract-aware zero-shot} prompt (\emph{ZS-contract}) that explicitly restates the civil-time conventions and output policy used in evaluation. The prompt specifies the five canonical answer forms: \texttt{ABSTAIN}, date or ISO week labels, ISO-8601 timestamps with explicit UTC offsets, ambiguity pairs (\texttt{A||B}), and intervals (\texttt{A..B}).

No exemplars are provided, distinguishing this condition from in-context learning. Decoding uses temperature $T=0$ and a maximum of 128 new tokens where supported. As described in Appendix~\ref{app:norm}, raw-output ContractOK is evaluated separately from extracted-answer FormatOK and exact-match correctness.

We examine per-category \textbf{Exact Match (EM)} with \textbf{95\% Wilson confidence intervals} for three representative open-weight models: Gemma-2-9B, Mistral-7B, and Qwen-2.5-7B-Instruct. Overall EM is computed over the full synthetic evaluation split of 10,000 items, with missing predictions counted as incorrect.

\begin{figure*}[t]
\centering
\includegraphics[width=0.96\textwidth]{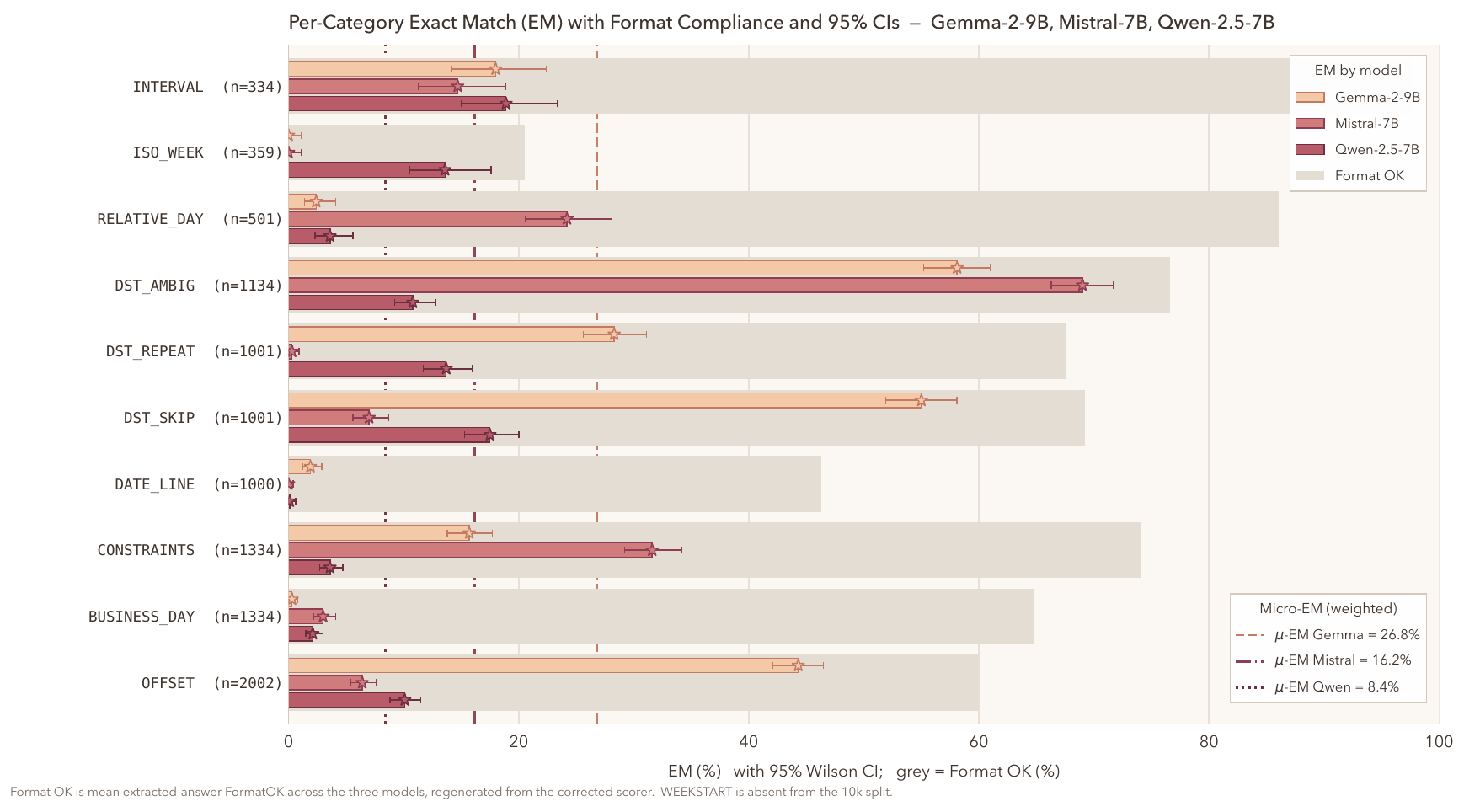}
\caption{\textbf{Contract-aware zero-shot performance by task type on SyntheticTimeBench.}
Per-category Exact Match (EM) for Gemma-2-9B, Mistral-7B, and Qwen-2.5-7B-Instruct on the 10,000-item synthetic evaluation split. Colored bars show EM with 95\% Wilson confidence intervals; gray backgrounds indicate the mean extracted-answer FormatOK across the three models. Vertical reference lines indicate overall instance-weighted EM (26.8\%, 16.2\%, and 8.4\%, respectively). Category sample sizes are shown alongside task labels. All 10 task types present in the full synthetic split are included; \textsc{WEEKSTART} is absent. Results use the corrected scoring protocol and the contract-aware zero-shot condition.}
\label{fig:appendix-prompt-enhancement-corrected}
\end{figure*}

\paragraph{Aggregate.}
Under \emph{ZS-contract}, overall EM is \textbf{26.8\%} for Gemma-2-9B, \textbf{16.2\%} for Mistral-7B, and \textbf{8.4\%} for Qwen-2.5-7B-Instruct. Their corresponding FormatOK values are 96.4\%, 50.0\%, and 51.3\%, respectively. The gap between FormatOK and EM demonstrates that producing an accepted answer form does not guarantee calendrical correctness. FormatOK measures extracted-answer validity and should not be conflated with raw-output ContractOK.

\paragraph{Per-category patterns.}
\begin{itemize}
  \item \textbf{\texttt{DST\_SKIP} and \texttt{OFFSET}:} Gemma achieves 55.0\% and 44.3\% EM, respectively, outperforming the other two models on these categories.

  \item \textbf{\texttt{CONSTRAINTS} and \texttt{RELATIVE\_DAY}:} Mistral performs best among the three models, reaching 31.6\% and 24.2\% EM, respectively.

  \item \textbf{\texttt{DATE\_LINE} and \texttt{BUSINESS\_DAY}:} All three models perform poorly. On \texttt{DATE\_LINE}, EM is at most 1.9\%, while \texttt{BUSINESS\_DAY} remains at or below 3.0\%.

  \item \textbf{\texttt{ISO\_WEEK}:} Gemma and Mistral achieve 0.0\% EM, whereas Qwen reaches 13.6\%, indicating that the relative ranking of models varies across task types.

  \item \textbf{\texttt{DST\_AMBIG}:} Mistral achieves 69.0\% EM, followed by Gemma at 58.1\% and Qwen at 10.8\%. These results should be interpreted in light of the identical-endpoint ambiguity labels and their canonicalization, as described in Appendix~\ref{app:norm}.
\end{itemize}

\paragraph{Takeaways.}
Contract-aware prompting has heterogeneous effects across models and civil-time tasks. Gemma performs relatively well on offset arithmetic and DST skips, while Mistral performs better on ambiguity, constraints, and relative-day tasks. However, date-line conversion and business-day composition remain difficult for all three models. As illustrated in Figure~\ref{fig:appendix-prompt-enhancement-corrected}, the differences between per-category EM and mean FormatOK further reinforce the distinction between producing a valid answer representation and correctly resolving civil-time operations.

\section{Fine-Tuning Details}
\label{app:sft-details}

\subsection{QLoRA Configuration}
\label{app:sft-config}

We apply parameter-efficient QLoRA fine-tuning to five open-weight models: Gemma-2-9B-IT, Mistral-7B-Instruct-v0.3, Qwen-2.5-7B-Instruct, Llama-3.1-8B (base), and Phi-4-14B. Training uses the 60,011-example SyntheticTimeBench training split with completion-only loss, masking user-prompt tokens.

The following configuration documents the shared settings for Gemma, Mistral, and Qwen. These settings should not be assumed to describe every model-specific training detail for LLaMA and Phi-4.

\begin{itemize}\itemsep3pt
  \item \textbf{Tokenizer / length}: Fast tokenizer; left-padding and left-truncation; maximum sequence length $=256$.

  \item \textbf{Quantization}: 4-bit NF4 via \texttt{bitsandbytes}; compute dtype \texttt{fp16}.

  \item \textbf{LoRA adapters}: Rank $r=32$, scaling parameter $\alpha=32$, dropout $p=0.05$, and no bias adaptation. Target modules include attention projections $\{q,k,v,o\}$ and MLP projections $\{\text{gate},\text{up},\text{down}\}$. Base-model weights remain frozen.

  \item \textbf{Optimization}: Learning rate $2\times10^{-4}$; cosine learning-rate schedule with 3\% warm-up; one training epoch; gradient accumulation $=32$; per-device batch size $=2$ for training and evaluation; mixed-precision \texttt{fp16}; sequence packing disabled.

  \item \textbf{Validation / checkpoints}: Evaluation every 100 steps; checkpoint saving every 500 steps; \texttt{load\_best\_model\_at\_end} enabled using evaluation loss; early-stopping patience $=3$.

  \item \textbf{Monitoring}: Domain metrics, including EM, FormatOK, and abstention-related metrics, monitored on a fixed validation slice and logged to \texttt{domain\_metrics.csv}, alongside training and evaluation curves. The wall-clock guard is $42{,}900$ seconds.

  \item \textbf{Fine-tuned inference}: Quantized base model with LoRA adapters loaded separately (without merging). The documented fine-tuned inference configuration uses \texttt{max\_new\_tokens}$=50$, distinct from the 128-token limit used in the prompting experiments.
\end{itemize}

\subsection{Per-Model Notes}
\label{app:sft-models}

\begin{itemize}\itemsep3pt

  \item \textbf{Gemma-2-9B-IT}: Uses its native tokenizer and chat template with the configuration above. Its synthetic fine-tuning evaluation produces 8,532 predictions against 10,000 gold items, achieving 74.6\% EM. The HumanTime-500 fine-tuning prediction dump is invalid and is excluded from the corrected results.

  \item \textbf{Mistral-7B-Instruct-v0.3}: Uses its native tokenizer and instruction template with the configuration above. Fine-tuning achieves 81.4\% EM on SyntheticTimeBench and 45.2\% EM on HumanTime-500, with prediction coverage of 8,532/10,000 and 432/500, respectively.

  \item \textbf{Qwen-2.5-7B-Instruct}: Uses its native tokenizer and chat template with the configuration above. Its synthetic fine-tuning prediction identifiers do not match the available gold identifiers, preventing valid scoring on the 10,000-item evaluation split. Its HumanTime-500 fine-tuning results are reported separately: 41.6\% EM with 436/500 predictions.

  \item \textbf{Llama-3.1-8B (base)}: Uses the \texttt{meta-llama/Llama-3.1-8B} base checkpoint, not the instruction-tuned variant. Prompts use manually constructed role-header tokens rather than \texttt{apply\_chat\_template}. The corrected synthetic fine-tuning evaluation achieves 86.6\% EM with complete coverage (10,000/10,000); HumanTime-500 EM is 39.6\% with 500/500 predictions.

  \item \textbf{Phi-4-14B}: Fine-tuned using parameter-efficient adaptation. Its corrected synthetic evaluation achieves 74.6\% EM with complete prediction coverage (10,000/10,000). On HumanTime-500, it achieves 36.6\% EM with 500/500 predictions.

\end{itemize}

\subsection{Evaluation Coverage and Reporting}
\label{app:sft-coverage}

Although all five open-weight models underwent fine-tuning, only four have usable matched synthetic evaluation results: Gemma, LLaMA, Mistral, and Phi-4. Their SyntheticTimeBench EM ranges from 74.6\% to 86.6\%.

For HumanTime-500, the three models with valid results on both evaluation splits achieve 36.6--45.2\% EM. Qwen's HumanTime-500 result is reported separately because its synthetic fine-tuning predictions cannot be matched to the corresponding evaluation gold identifiers. Gemma's invalid HumanTime-500 dump is excluded.

All EM values are computed against the full gold evaluation split, with missing predictions counted as incorrect. Consequently, prediction coverage and EM are reported separately; incomplete runs are not evaluated using only their successfully generated predictions.

Fine-tuning results are evaluated using the corrected normalization and scoring protocol described in Appendix~\ref{app:norm}. FormatOK measures extracted-answer validity, whereas raw-output ContractOK is evaluated separately under the v2 protocol. Median Temporal Error (MedTE) is calculated only over cases with defined temporal error and should be interpreted alongside temporal-error coverage rather than as a substitute for EM.

\section{Statement on the Use of AI Tools}
\label{app:ai-tools}

During the preparation of this manuscript, the authors used artificial intelligence (AI) tools to assist with research-related and editorial tasks. Specifically, AI tools were employed for:

\begin{itemize}
  \item \textbf{Scholarly writing and editing:} Improving grammar, clarity, organization, and terminology; refining abstracts, section text, and figure/table captions while preserving the authors' intended scientific content.

  \item \textbf{Code and evaluation assistance:} Identifying potential implementation inconsistencies, reviewing evaluation procedures, and assisting with checks of reported metrics, statistical calculations, and experimental results. Final implementations, scoring decisions, and reported findings were reviewed by the authors.

  \item \textbf{Figures and presentation:} Assisting with the preparation and refinement of plots, tables, and visualizations based on experimental data and evaluation outputs provided by the authors.

  \item \textbf{Dataset authoring:} GPT-5 Pro was used to generate fictionalized, application-grounded scenarios for HumanTime-500. These LLM-authored examples were subsequently reviewed and validated by three human evaluators, including checks of plausibility and reference-answer correctness, as described in Appendix~\ref{sec:humantime500}.
\end{itemize}

AI-generated suggestions and outputs were subject to human review and, where necessary, correction or rejection. The authors retained responsibility for the research questions, experimental design, dataset curation, methodological decisions, statistical interpretation, and scientific conclusions. AI tools were not treated as independent sources of experimental evidence.

The authors take full responsibility for the accuracy of the manuscript, the correctness of the code and reported results, the validity of the released benchmark, and the reproducibility of the experiments. No confidential, proprietary, or personally identifiable information was provided to AI services.

\section{Potential Risks and Ethical Considerations}
\label{app:Potential Risks and Ethical Considerations}

This appendix expands the discussion of dataset construction and intended use in Appendix~\ref{app:dataset}. We outline plausible misuse scenarios, potential downstream harms, bias and representativeness concerns, and recommended safeguards for research and deployment.

\subsection{Scope and Threat Model}
\label{sec:task}

Our work targets \emph{research} on civil-time reasoning in NLP systems. We primarily consider non-adversarial scientific users, while also accounting for (i) well-intentioned practitioners who may overestimate model capabilities and (ii) misuse in settings where temporal correctness is safety- or compliance-critical, such as healthcare scheduling, financial settlement, legal deadlines, and audit trails.

We do not explicitly evaluate sophisticated adversarial actors. Nevertheless, the benchmark exposes failure modes that could be exploited or amplified in downstream applications. Strong benchmark performance should not be interpreted as evidence of readiness for safety-critical deployment.

\subsection{Misuse Scenarios}
\label{app:risks:misuse}

\paragraph{Automation without oversight.}
Deploying models evaluated or adapted using our resources to automatically compute time-sensitive decisions, such as court filing deadlines, cross-border tax timelines, or medication schedules, could cause harm when temporal assumptions are incorrect. Errors involving DST transitions, time-zone offsets, or jurisdiction-specific rules may propagate into operational decisions.

\paragraph{Regulatory or compliance claims.}
Using model outputs to establish compliance with regulatory requirements, such as retention periods or financial reporting deadlines, may create legal and operational risks if historical time-zone changes, jurisdiction-specific holidays, or other calendar conventions are mishandled. Our benchmark does not establish regulatory compliance.

\paragraph{Security and forensics.}
Incorrect conversion or interpretation of timestamps across time zones may compromise incident reconstruction, log correlation, or audit-trail analysis. Such errors could obscure event ordering or undermine the reliability of forensic conclusions.

\paragraph{Misinformation.}
Generating convincing but temporally inaccurate summaries of events may contribute to timeline distortion and misinformation, particularly when readers rely on precise dates, durations, or chronological relationships.

\paragraph{Proxy discrimination.}
Temporal features, including local holidays, regional workweeks, and scheduling conventions, may correlate with demographic or geographic characteristics. Their careless use in downstream decision-making systems could introduce disparate impacts across populations.

\subsection{Downstream Harm Analysis}
\label{app:risks:harm}

Table~\ref{tab:risk_matrix} summarizes plausible downstream risks and recommended mitigations. Likelihood and impact assessments are qualitative and illustrative rather than empirically estimated. Actual risks depend on the deployment domain, system architecture, and available oversight.

\begin{table*}[t]
  \centering
  \small
  \setlength{\tabcolsep}{5pt}
  \renewcommand{\arraystretch}{1.12}

  \begin{tabular}{
    @{}
    p{3.2cm}
    p{2.2cm}
    p{2.2cm}
    p{5.4cm}
    @{}
  }
    \hline
    \textbf{Risk}
    & \textbf{Likelihood}
    & \textbf{Impact}
    & \textbf{Primary Mitigations} \\
    \hline

    Automation without oversight
    & Medium
    & High
    & Human review, domain-specific validation, fail-closed defaults,
      and tests covering jurisdictional edge cases. \\

    Compliance claims
    & Low--Medium
    & High
    & Independent legal review, authoritative calendrical sources,
      and traceable provenance for temporal rules. \\

    Security or forensic misinterpretation
    & Low
    & High
    & Independent timestamp verification, synchronized clock sources,
      and immutable audit logs. \\

    Misinformation
    & Medium
    & Medium
    & Timestamped source verification, uncertainty communication,
      and temporal factual-consistency checks. \\

    Proxy discrimination
    & Low--Medium
    & Medium
    & Fairness audits, documentation of regional coverage,
      and avoidance of sensitive-attribute inference. \\
    \hline
  \end{tabular}

  \caption{\textbf{Qualitative risk matrix for downstream use.}
  Illustrative likelihood and impact assessments for potential harms arising from incorrect civil-time reasoning, together with recommended mitigations. Ratings are not derived from measured deployment frequencies and may differ substantially across application domains.}
  \label{tab:risk_matrix}
\end{table*}

\subsection{Bias, Coverage, and Representativeness}
\label{app:risks:bias}

Temporal conventions are not uniform across regions or historical periods. DST rules change over time, jurisdictions differ in weekend and holiday conventions, and historical calendar transitions create additional edge cases.

Our synthetic training and evaluation sets share the same generators, while HumanTime-500 contains only 500 LLM-authored, human-validated English examples. Although the benchmark covers several important civil-time phenomena, these design choices limit its representativeness.

Potential sources of bias include:

\begin{itemize}
    \item \textbf{Locale bias:} Over-representation of selected IANA time zones, calendar conventions, and English-language expressions.

    \item \textbf{Rule-version bias:} Dependence on particular time-zone database and library versions, whose rules may subsequently change.

    \item \textbf{Phenomenon bias:} Incomplete coverage of rare but consequential events, including historical time-zone transitions, missing-day changes, leap seconds, and jurisdiction-dependent holidays.

    \item \textbf{Generation bias:} Shared synthetic generators may introduce recurring templates or patterns that make held-out synthetic evaluation easier than independently authored naturalistic problems.
\end{itemize}

We therefore encourage per-category and per-locale evaluation, explicit documentation of temporal rule versions, and testing on independently collected data.

The corrected evaluation reports per-category EM with 95\% Wilson confidence intervals. These intervals help characterize uncertainty, particularly for the smaller HumanTime-500 categories, but do not eliminate representativeness limitations.

\subsection{Privacy and Safety}
\label{app:risks:privacy}

The benchmark is constructed using synthetic examples rather than real user logs. HumanTime-500 scenarios were authored using GPT-5 Pro and subsequently validated by human evaluators. This approach reduces the need to collect sensitive operational records or personally identifiable information.

Nevertheless, synthetic generation does not automatically guarantee the absence of sensitive or inappropriate content. We recommend the following precautions when extending or releasing such resources:

\begin{itemize}
    \item Prefer synthetic or appropriately licensed, non-personal sources over real user records where feasible.

    \item Screen generated examples for personally identifiable information, including names, exact addresses, telephone numbers, and identifying records.

    \item Supplement automated screening with human review, especially for naturalistic scenarios.

    \item Avoid unnecessarily incorporating sensitive personal attributes or realistic confidential information into prompts and labels.
\end{itemize}

If released artifacts are found to contain sensitive information or harmful content, affected items should be reviewed and corrected or withdrawn, with corresponding changes documented.

\subsection{Environmental Considerations}
\label{app:risks:env}

Evaluating and fine-tuning language models requires computational resources, particularly when comparing multiple models and prompting conditions.

Our experiments use parameter-efficient QLoRA adaptation for five open-weight models and deterministic, single-run evaluation per model and condition. These choices limit some computational costs relative to full-parameter fine-tuning or extensive repeated sampling, although they do not eliminate the environmental impact of the experiments.

To support more resource-efficient follow-up research, we recommend:

\begin{itemize}
    \item Reusing model checkpoints and evaluation outputs when methodologically appropriate.

    \item Preferring parameter-efficient adaptation when it provides adequate performance.

    \item Reporting GPU-hours, energy consumption, or carbon estimates when reliably measured.

    \item Releasing reproducible evaluation code and results to reduce unnecessary duplication of expensive experiments.
\end{itemize}

\subsection{Safeguards and Mitigations}
\label{app:risks:safeguards}

We recommend the following safeguards for practitioners extending or deploying systems evaluated using our benchmark:

\begin{enumerate}
    \item \textbf{Intended-use restrictions.}
    Treat benchmark performance as research evidence, not deployment certification. Safety- and compliance-critical applications require independent validation and appropriate domain approval.

    \item \textbf{Human oversight.}
    Require qualified human review for decisions with material consequences. Systems should avoid automatically treating a syntactically valid timestamp as evidence of correctness.

    \item \textbf{Authoritative temporal computation.}
    Prefer validated calendrical libraries and authoritative jurisdiction-specific rules for operational calculations. Model-generated answers should be independently checked before consequential use.

    \item \textbf{Version pinning and provenance.}
    Record versions of time-zone databases and relevant temporal libraries. Revalidate affected calculations and regression tests when dependencies or jurisdictional rules change.

    \item \textbf{Evaluation beyond averages.}
    Report performance by task type, time zone, and temporal phenomenon where data permit, together with uncertainty estimates and prediction coverage.

    \item \textbf{Boundary testing.}
    Include regression tests for DST transitions, date-line crossings, fractional UTC offsets, year boundaries, and relevant historical rule changes.

    \item \textbf{Documentation.}
    Communicate known failure modes, dataset limitations, and the distinction between formatting compliance and calendrical correctness.
\end{enumerate}

\subsection{Red-Teaming and Adversarial Testing}
\label{app:risks:redteam}

Our benchmark includes controlled stress tests targeting difficult civil-time phenomena, including ambiguous or nonexistent local times, fractional offsets, date-line crossings, and multi-step calendar operations. These tests reveal important failure modes but do not constitute a comprehensive adversarial security evaluation.

Future work should extend this coverage through targeted adversarial testing. We recommend:

\begin{itemize}
    \item Combining multiple temporal edge cases within individual problems, such as cross-zone conversions near DST transitions.

    \item Introducing conflicting or underspecified temporal information to test whether models correctly identify the need for abstention.

    \item Varying prompt formulations and decoding strategies to identify prompt-dependent failures.

    \item Recording reproducible error cases and distinguishing reasoning failures from output-format and generation failures.
\end{itemize}

Such evaluations would help characterize robustness beyond the controlled distributions examined in this study.

\subsection{Responsible Release and Governance}
\label{app:risks:release}

Responsible release requires clear documentation of artifact provenance, permitted uses, limitations, and mechanisms for addressing identified errors.

We recommend the following governance practices for the benchmark and any downstream extensions:

\begin{itemize}
    \item \textbf{License and scope.}
    Specify the applicable licenses and intended-use conditions for each released artifact, as discussed in Appendix~\ref{app:dataset}. The benchmark is intended for research and evaluation, not as certification for safety- or compliance-critical applications.

    \item \textbf{Issue reporting.}
    Provide a mechanism for reporting incorrect labels, evaluation inconsistencies, potential harms, and documentation errors.

    \item \textbf{Sensitive reports.}
    Establish an appropriate contact mechanism for concerns involving privacy, security, or potentially harmful content that should not initially be disclosed publicly.

    \item \textbf{Versioning and corrections.}
    Document dataset and evaluator revisions, including changes to canonicalization, time-zone rules, gold labels, and scoring procedures. Preserve sufficient provenance to distinguish results obtained under different evaluation versions.

    \item \textbf{Transparency.}
    Clearly distinguish validated evaluation results from incomplete or unusable experimental artifacts. Report prediction coverage and avoid presenting missing or invalid runs as successful evaluations.
\end{itemize}

These measures are recommended governance practices; their implementation and availability should be documented alongside the released artifacts.

\subsection{Limitations of This Risk Assessment}
\label{app:risks:limitations}

This assessment focuses on foreseeable risks associated with text-based civil-time reasoning and downstream use of the benchmark. It is qualitative and does not constitute a formal security audit, fairness evaluation, or domain-specific safety certification.

Real-world applications introduce additional hazards involving institutional procedures, legal requirements, operational dependencies, and human decision-making. These factors require domain expertise, independent verification, and continued monitoring.

The safeguards described above can reduce risks but cannot eliminate incorrect temporal predictions or guarantee safe deployment. In particular, the substantial performance gap between synthetic evaluation and HumanTime-500 underscores the importance of testing systems beyond the benchmark distribution before relying on them for consequential decisions.

\section{Artifacts}
\label{app:Artifacts}

This appendix documents the research artifacts associated with CivilTime-Bench, including benchmark data, evaluation code, output formats, intended use, privacy considerations, and reproducibility. The resources are intended to support independent evaluation of civil-time reasoning in language models. Dataset construction and evaluation procedures are described in Appendices~\ref{app:dataset} and~\ref{app:norm}, respectively.

\subsection{What We Release}
\label{app:artifacts:what}

The research artifacts comprise:

\begin{itemize}[nosep, leftmargin=1.3em]
    \item \textbf{Benchmark data:} SyntheticTimeBench, containing 60,011 training examples and 10,000 synthetic evaluation examples, and HumanTime-500, containing 500 LLM-authored, human-validated evaluation items.
    \item \textbf{Code:} Benchmark and evaluation repository:
    \url{https://github.com/CivilTime-Bench/CivilTime-Bench}.
    \item \textbf{Evaluation tools:} Deterministic answer extraction, canonicalization, exact-match scoring, format validation, and temporal-error diagnostics.
    \item \textbf{Statistical analysis:} Per-example comparisons supporting McNemar tests and overall/per-category 95\% Wilson confidence intervals.
\end{itemize}

Fine-tuned model checkpoints are not required to reproduce the scoring of saved predictions and are not claimed as released artifacts here.

\subsection{Availability and Licensing}
\label{app:artifacts:license}

The dataset and evaluation code are publicly available at
\url{https://github.com/CivilTime-Bench/CivilTime-Bench}
under the Apache License~2.0.
Users should consult the repository \texttt{LICENSE} file for redistribution, modification, and commercial-use terms.
Use of third-party model checkpoints remains subject to their respective licenses.
The intended-use notes below do not add restrictions beyond that license.

\subsection{Documentation and Schema}
\label{app:artifacts:docs}

\paragraph{Task.}
CivilTime-Bench evaluates textual civil-time reasoning involving time-zone conversions, daylight saving time transitions, fractional UTC offsets, International Date Line crossings, ISO week calculations, business-day arithmetic, relative dates, and intervals.

\paragraph{Data format.}
Benchmark items are represented as JSONL records containing fields such as:

\begin{verbatim}
{
  "id": "...",
  "type": "...",
  "tz": "...",
  "input_text": "...",
  "gold": "..."
}
\end{verbatim}

Cross-time-zone tasks may additionally include \texttt{tz\_src} and \texttt{tz\_dst}. The \texttt{gold} field contains the reference answer used for evaluation.

\paragraph{Dataset composition.}
SyntheticTimeBench provides controlled examples generated using seeded category-specific generators. Its training and evaluation splits contain disjoint examples but share the same generation procedures.

HumanTime-500 provides complementary naturalistic scenarios across 11 task types. These examples were authored using GPT-5 Pro and subsequently validated by three human evaluators. HumanTime-500 is evaluation-only.

The full synthetic 10k evaluation split covers 10 task types. The API synthetic 1k subset and HumanTime-500 additionally contain \texttt{WEEKSTART} examples. Task distributions and generation details are provided in Appendix~\ref{app:dataset}.

\subsection{Intended Use and Misuse}
\label{app:artifacts:intended}

\textbf{Intended use.}
The artifacts are designed for research and education in civil-time reasoning, evaluation methodology, temporal robustness, output-format compliance, and parameter-efficient model adaptation.

\textbf{Out-of-scope use.}
Benchmark performance should not be interpreted as certification for safety-critical or compliance-sensitive applications, including medical scheduling, financial settlement, legal deadlines, or forensic timeline reconstruction.

Such applications require independent validation against authoritative temporal rules and appropriate domain-specific oversight. Additional risks and mitigations are discussed in Appendix~\ref{app:Potential Risks and Ethical Considerations}.

\subsection{Privacy and Personally Identifiable Information}
\label{app:artifacts:privacy}

The benchmark is constructed from synthetic examples rather than real user requests or operational logs. HumanTime-500 uses fictionalized scenarios designed to resemble practical civil-time problems without reproducing identifiable user records.

This design reduces privacy risks associated with collecting real scheduling, financial, or administrative information. Nevertheless, synthetic generation does not eliminate the need for content review.

Potential privacy concerns or incorrectly included sensitive material should be reported through the release repository so that affected examples can be reviewed and corrected.

\subsection{Evaluation Contract and Metrics}
\label{app:artifacts:eval}

\paragraph{Output contract.}

The benchmark defines five canonical answer forms:

\begin{enumerate}[leftmargin=1.3em, itemsep=2pt]
    \item \texttt{ABSTAIN} for cases requiring abstention under the stated policy.
    \item \texttt{YYYY-MM-DD} or \texttt{YYYY-Www} for dates and ISO week labels.
    \item An ISO-8601 timestamp with seconds and an explicit UTC offset (or \texttt{Z}).
    \item \texttt{A||B} for civil-time ambiguity.
    \item \texttt{A..B} for half-open intervals $[A,B)$.
\end{enumerate}

The documented contract uses \texttt{..} for intervals and \texttt{||} for ambiguity. However, the synthetic evaluation splits store interval reference answers using \texttt{A||B}, whereas HumanTime-500 uses \texttt{A..B}. The scorer canonicalizes these representations before exact-match comparison.

Similarly, synthetic \texttt{ISO\_WEEK} reference answers use \texttt{YYYY-Www}, whereas HumanTime-500 uses date-form labels. Identical-endpoint ambiguity representations (\texttt{A||A}) are also canonicalized to a single value.

These serialization differences are documented in Appendix~\ref{app:norm}.

\paragraph{Scoring pipeline.}

Evaluation separates raw-output compliance, answer extraction, normalization, and correctness scoring.

\textbf{ContractOK} measures whether the raw completion satisfies the output contract without extracting an answer from surrounding prose. The v2 evaluator accepts either a canonical one-line answer or a JSON object containing a canonical answer value.

\textbf{FormatOK} measures whether the extracted and normalized prediction matches an accepted answer form, independently of task type.

\textbf{Exact Match (EM)} requires equality between the canonicalized prediction and reference answer. Canonicalization includes supported timestamp normalization, interval delimiter normalization, and identical-endpoint collapse. No temporal tolerance or partial credit is used for EM.

Tables~\ref{tab:zeroshot_results}, \ref{tab:icl_results}, and~\ref{tab:finetune_results} report EM and FormatOK. Raw-output ContractOK is evaluated separately under the v2 protocol.

\paragraph{Additional metrics.}

We report overall EM and, where applicable, macro-averaged and per-category EM. Overall EM is weighted by the number of evaluation items, whereas MacroEM averages equally across task types.

Temporal-error diagnostics include Median Temporal Error (MedTE) and temporal-error coverage (TEcov). For eligible single-instant predictions, error is measured in seconds. Ambiguity cases use the nearest valid endpoint, while parseable intervals use the maximum absolute deviation of their two endpoints.

Abstention diagnostics include Abstention-Required Accuracy (ARA), Over-Abstention Rate (OAR), and Abstain-F1. Paraphrase FlipRate is an optional diagnostic.

\paragraph{Statistical uncertainty.}

We compute 95\% Wilson score confidence intervals for overall and per-category EM. Paired McNemar tests compare per-example EM outcomes, using a continuity-corrected chi-squared statistic or a two-sided exact binomial test when the number of discordant pairs is at most 50.

Statistical significance thresholds are unadjusted. The intervals reflect uncertainty across evaluation items under the single-run setup, rather than variability across repeated model generations.

\subsection{Reproducibility}
\label{app:artifacts:repro}

\paragraph{Deterministic evaluation.}
Models are evaluated using temperature $T=0$ and deterministic decoding settings where supported. Each model-condition combination is evaluated in a single run.

\paragraph{Split definitions.}
Open-weight models are evaluated on the 10,000-item synthetic evaluation split, whereas API models use a 1,000-item synthetic subset. All seven models are evaluated on HumanTime-500.

\paragraph{Prediction coverage.}
The scorer computes EM against the full corresponding gold split. Missing predictions count as incorrect, and incomplete prediction coverage is reported separately.

This convention is particularly important for the API few-shot conditions and some fine-tuning runs.

\paragraph{Invalid artifacts.}
Unusable evaluation outputs are excluded from reported comparisons rather than treated as valid model predictions.

Specifically, Gemma's HumanTime-500 fine-tuning dump is invalid, and Qwen's synthetic fine-tuning predictions cannot be reliably matched to the corresponding gold identifiers. Qwen's HumanTime-500 fine-tuning result is reported separately.

\paragraph{Environment and implementation.}
Dataset generation relies on specified calendrical and time-zone libraries, including \texttt{pytz} and IANA time-zone rules. Reproduction should use the dependency versions and execution instructions associated with the released repository.

Fine-tuning and model-specific implementation details are documented in Appendices~\ref{app:sft-details} and~\ref{app:systems}.

\subsection{Known Limitations}
\label{app:artifacts:limits}

CivilTime-Bench concentrates on selected calendrical phenomena under modern Gregorian assumptions and a curated set of time zones and years.

The synthetic training and evaluation splits share generators, limiting conclusions about generalization beyond their template distribution. HumanTime-500 offers complementary naturalistic evaluation but contains only 500 LLM-authored, human-validated English examples.

The benchmark does not comprehensively cover historical calendar transitions, non-Gregorian calendars, region-specific floating holidays, or all jurisdiction-dependent scheduling conventions.

Consequently, strong performance on the released benchmark should not be extrapolated to unrestricted real-world temporal reasoning without further evaluation.

\subsection{How to Use}
\label{app:artifacts:howto}

To reproduce evaluation results, users should obtain the benchmark data and scoring code from the released repository, prepare model predictions associated with the corresponding evaluation item identifiers, and run the documented scorer against the matching gold split.

The evaluation procedure consists of:

\begin{enumerate}[leftmargin=1.3em, itemsep=2pt]
    \item Selecting the appropriate synthetic or HumanTime-500 evaluation split.
    \item Providing predictions matched to the gold item identifiers.
    \item Applying deterministic answer extraction and canonicalization.
    \item Computing EM, FormatOK, temporal-error diagnostics, and prediction coverage.
    \item Computing per-category statistics and confidence intervals when required.
\end{enumerate}

The repository should be consulted for the exact executable commands, required dependencies, and supported prediction-file formats. The complete scoring definitions are provided in Appendix~\ref{app:norm}.

\subsection{Versioning and Checksums}
\label{app:artifacts:versioning}

Reproducibility depends on identifying the benchmark data, prediction files, and scoring implementation used for each reported result.

EM and FormatOK from the released scorer should be distinguished from any earlier implementations. ContractOK values produced by the v2 raw-output evaluator should likewise be identified separately from the main results tables.

The repository documentation identifies the applicable dataset and scorer versions. File checksums, manifests, and migration notes can provide additional verification when included in the released artifacts.

\subsection{Citation and Acknowledgments}
\label{app:artifacts:cite}

Researchers using CivilTime-Bench should cite the accompanying paper and the benchmark repository:

\paragraph{Code and benchmark resources.}
\url{https://github.com/CivilTime-Bench/CivilTime-Bench}

\paragraph{Checklist Mapping (B).}
B: Artifact availability (\S\ref{app:artifacts:what});
B1: Citation (\S\ref{app:artifacts:cite});
B2: Licensing (\S\ref{app:artifacts:license});
B3: Intended use (\S\ref{app:artifacts:intended});
B4: Privacy (\S\ref{app:artifacts:privacy});
B5: Documentation (\S\ref{app:artifacts:docs}).

\end{document}